\documentclass{article}

\usepackage[numbers, sort&compress]{natbib}

\usepackage[final]{neurips_2023}

\usepackage[toc,page,header]{appendix}
\usepackage{minitoc}

\usepackage[utf8]{inputenc} 
\usepackage[T1]{fontenc}    

\usepackage{url}            
\usepackage{booktabs}       
\usepackage{amsfonts}       
\usepackage{nicefrac}       
\usepackage{microtype}      

\usepackage{caption}
\usepackage{subcaption}
\usepackage[dvipsnames]{xcolor}
\usepackage{algorithm}
\usepackage{algorithmicx}
\usepackage{algpseudocode}
\usepackage{xspace}

\usepackage{hyperref}
\hypersetup{
    colorlinks = true,
    linkcolor = blue,
    anchorcolor = blue,
    citecolor = blue,
    filecolor = blue,
    urlcolor = blue!50
    }

\usepackage{hyphenat}
\usepackage{amsfonts,amsmath,amssymb}
\usepackage{hhline}
\usepackage{bm}

\usepackage{pifont}

\usepackage{bbm,enumerate,subcaption}

\usepackage{enumitem}
\usepackage{physics}
\usepackage{siunitx}
\usepackage{caption}
\usepackage{subcaption}
\usepackage{wrapfig}

\usepackage{microtype}
\usepackage{graphicx}
\usepackage{booktabs} 
\usepackage{makecell}
\usepackage{tabularx}

\usepackage{booktabs, multirow} 
\usepackage{soul}
\usepackage{changepage,threeparttable} 

\newcommand{\xmark}{\textcolor{red}{\ding{55}}}%
\newcommand{\cmark}{\textcolor{ForestGreen}{\ding{52}}}%

\usepackage{amsmath,amsfonts,bm}

\def\eqref#1{equation~\ref{#1}}

\def\1{\bm{1}}

\DeclareMathAlphabet{\mathsfit}{\encodingdefault}{\sfdefault}{m}{sl}
\SetMathAlphabet{\mathsfit}{bold}{\encodingdefault}{\sfdefault}{bx}{n}

\DeclareMathOperator*{\argmin}{arg\,min}

\newcommand{\threeS}{\texttt{3S}~\xspace}

\usepackage{pgfplots}
\usepackage[framemethod=TikZ]{mdframed}
\usepackage{comment}
\usepackage{pgfplotstable}
\usepackage{colortbl}

\mdfsetup{%
    middlelinecolor =   none,
    middlelinewidth =   1pt,
    backgroundcolor =   blue!5,
    roundcorner     =   5pt,
}

\newtheorem{example}{Example}

\title{Can You Rely on Your Model Evaluation? Improving Model Evaluation with Synthetic Test Data}

\author{%
  Boris van Breugel\thanks{Equal Contribution} \\
  University of Cambridge\\
  \texttt{bv292@cam.ac.uk} 
  \And
  Nabeel Seedat$^*$ \\
  University of Cambridge\\
  \texttt{ns741@cam.ac.uk} 
  \AND
  Fergus Imrie \\
  UCLA \\
  \texttt{imrie@g.ucla.edu}
  \And
  Mihaela van der Schaar \\
  University of Cambridge\\
  \texttt{mv472@cam.ac.uk}
}

\begin{document}

\doparttoc
\faketableofcontents

\maketitle

\begin{abstract}
Evaluating the performance of machine learning models on diverse and underrepresented subgroups is essential for ensuring fairness and reliability in real-world applications. However, accurately assessing model performance becomes challenging due to two main issues: (1) a scarcity of test data, especially for small subgroups, and (2) possible distributional shifts in the model's deployment setting, which may not align with the available test data.  In this work, we introduce \threeS Testing, a deep generative modeling framework to facilitate model evaluation by generating synthetic test sets for small subgroups and simulating distributional shifts.
Our experiments demonstrate that \threeS Testing outperforms traditional baselines---including real test data alone---in estimating model performance on minority subgroups and under plausible distributional shifts. In addition, \threeS offers intervals around its performance estimates, exhibiting superior coverage of the ground truth compared to existing approaches. 
Overall, these results raise the question of whether we need a paradigm shift away from limited real test data towards synthetic test data.
\end{abstract}

\newcommand{\X}{\mathcal{X}}
\newcommand{\Y}{\mathcal{Y}}
\newcommand{\Z}{\mathcal{Z}}
\renewcommand{\H}{\mathcal{H}}
\newcommand{\G}{\mathcal{G}}
\newcommand{\Dtrain}{\mathcal{D}_{\textrm{train}}}
\newcommand{\Dtest}{\mathcal{D}_{\textrm{test}}}
\newcommand{\D}{\mathcal{D}}
\newcommand{\cD}{\mathcal{D}}

\renewcommand{\P}{\mathbb{P}}
\newcommand{\Ph}{\hat{\P}}
\newcommand{\RR}{\mathbb{R}}
\newcommand{\xh}{\hat{x}}
\newcommand{\yh}{\hat{y}}
\newcommand{\zh}{\hat{z}}
\newcommand{\rxh}{\hat{X}}
\newcommand{\ryh}{\hat{Y}}
\newcommand{\rzh}{\hat{Z}}
\newcommand{\dzh}{\delta \hat{z}}
\newcommand{\dxh}{\delta \hat{x}}
\newcommand{\dyh}{\delta \hat{y}}
\newcommand{\rdzh}{\delta \hat{Z}}
\newcommand{\rdxh}{\delta \hat{X}}
\newcommand{\rdyh}{\delta \hat{Y}}
\newcommand{\proba}[2][\C]{p_{#1}\left(#2 \right)}
\newcommand{\cproba}[3][\C]{\proba[#1]{#2 \mid #3}}
\newcommand{\Supp}{\text{Supp}}
\newcommand{\set}[1]{\left\{ #1 \right\}}

\section{Introduction} \label{sec:introduction}

\textbf{Motivation.} Machine learning (ML) models are increasingly deployed in high-stakes and safety-critical areas, e.g. medicine or finance---settings that demand reliable and measurable performance \cite{shergadwala2022human}. Failure to rigorously test systems could result in models at best failing unpredictably and at worst leading to silent failures. Regrettably, such failures of ML are all too common \citep{oakden2020hidden,suresh2019framework,cabrera2019fairvis,cabreradiscovery,pianykh2020continuous,quinonero2008dataset,koh2021wilds,seedat2022dc}. Many mature industries involve standardized processes to evaluate performance under various testing and operating conditions \citep{gebru2021datasheets}. For instance, automobiles use wind tunnels and crash tests to assess specific components, whilst electronic component data sheets outline conditions where reliable operation is guaranteed. Unfortunately, current evaluation approaches of supervised ML models do not have the same level of detail and rigor. 

The prevailing testing approach in ML is to evaluate only using average prediction performance on a held-out test set. This can hide undesirable performance differences on a more granular level, e.g. for small subgroups \citep{oakden2020hidden,suresh2018learning,goel2020model, cabrera2019fairvis,cabreradiscovery}, 
low-density regions \citep{saria2019tutorial,d2020underspecification,cohen2021problems}, and individuals \citep{deo2015machine,savage2012better,nezhad2017subic}. Standard ML testing also ignores distributional shifts. In an ever-evolving world where ML models are employed across borders, failing to anticipate shifts between train and deployment data can lead to overestimated real-world performance \cite{pianykh2020continuous,quinonero2008dataset,koh2021wilds, patel2008investigating, recht2019imagenet}.

\emph{However, real test data alone does not always suffice for more detailed model evaluation.} Indeed, testing can be done on a granular level by evaluating on individual subgroups, and in theory, shifts could be tested using e.g. rejection or importance sampling of real test data. The main challenge is that \emph{insufficient amounts of test data} cause \emph{inaccurate} performance estimates \cite{miller2021model}. In Sec.~\ref{sec:formulation} we will further explore why this is the case, but for now let us give an example. 

\begin{example}
Consider the real example of estimating model performance on the Adult dataset, looking at race and age variables---see Fig. \ref{fig:example} and experimental details in Appendix \ref{appx:experimental_details}. There are limited samples of the older Black subgroup, leading to significantly erroneous performance estimates compared to an oracle. Similarly, if we tried to engineer a distribution shift towards increased age using rejection sampling, the scarcity of data would yield equally imprecise estimates. Such imprecise performance estimates could mislead us into drawing false conclusions about our model's capabilities. In Sec.~\ref{exp-d1}, we empirically demonstrate how synthetic data can rectify this shortfall.
\end{example}

\begin{wrapfigure}{r}{0.45\textwidth}
\vspace{-5mm}
\centering
\includegraphics[width=0.45\textwidth]{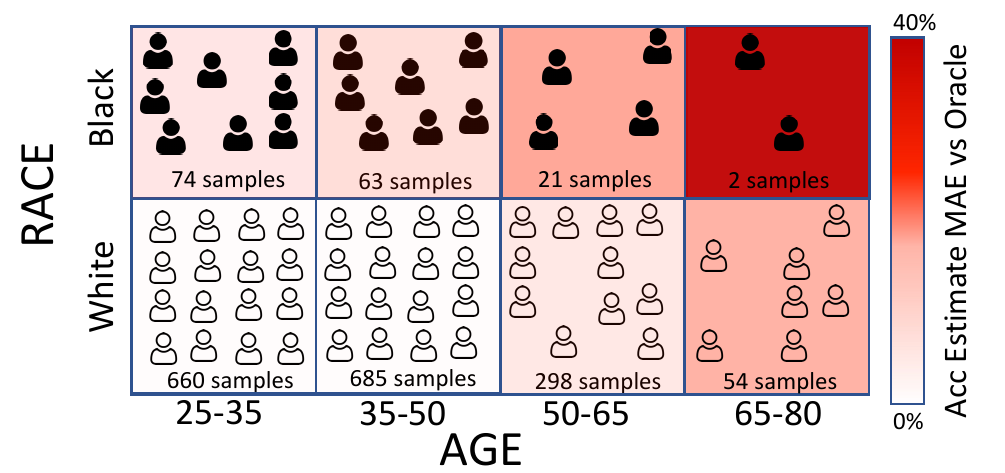}
 \vspace{-4mm}
 \caption{Limitation of testing with real data (Adult dataset):
Insufficient real test samples result in imprecise performance estimates
when (i) testing underrepresented subgroups (e.g. Black) and
(ii) testing effect of a shift (e.g increased age)}
\vspace{-1mm}
\rule{0.45\textwidth}{0.1pt}
\label{fig:example}
\vspace{-7mm}
\end{wrapfigure}

\textbf{Aim.} Our goal is to build a model evaluation framework with synthetic data that allows engineers, auditors, business stakeholders, or policy and compliance teams to understand better when they can rely on the predictions of their trained ML models and where they can improve the model further. We desire the following properties for our evaluation framework:\\
\textbf{(P1) Granular evaluation}: accurately evaluate model performance on a granular level, even for regions with few test samples. \\
\textbf{(P2) Distributional shifts}: accurately assess how distribution shifts affect model performance.

Our primary focus is tabular data. Not only are many high-stakes applications predominately tabular, such as credit scoring and medical forecasting \cite{borisov2021deep,shwartz2022tabular}, but the ubiquity of tabular data in real-world applications also presents opportunities for broad impact. To put it in perspective, nearly 79\% of data scientists work with tabular data on a daily basis, dwarfing the 14\% who work with modalities such as images \cite{kaggle_2017}. 

Moreover, tabular data presents us with interpretable feature identifiers, such as ethnicity or age, instrumental in defining minority groups or shifts.  This contrasts with other modalities, where raw data is not interpretable and external (tabular) metadata is needed.

\textbf{Contributions.} 
\textbf{\textcolor{ForestGreen}{\textcircled{1}}} 
\textbf{\emph{Conceptually}}, we show why real test data may not suffice for \emph{model evaluation} on subgroups and distribution shifts and how synthetic data can help (Sec. \ref{sec:formulation}). 
\textbf{\textcolor{ForestGreen}{\textcircled{2}}} 
\textbf{\emph{Technically}}, we propose the framework  \texttt{3S}-Testing (s.f. \textbf{S}ynthetic data for \textbf{S}ubgroup and \textbf{S}hift Testing) (Sec. \ref{sec:method}). \texttt{3S} uses conditional deep generative models to create synthetic test sets, addressing both (P1) and (P2). \texttt{3S} also accounts for possible errors in the generative process itself, providing uncertainty estimates for its predictions via a deep generative ensemble (DGE) \cite{breugel2023dge}. 
\textbf{\textcolor{ForestGreen}{\textcircled{3}}} 
\textbf{\emph{Empirically}},  we show synthetic test data provides a more accurate estimate of the true model performance on small subgroups compared to baselines, including real test data (Sec.~\ref{exp:subgroup-correct}), with prediction intervals providing good coverage of the real value (Sec.~\ref{exp:coverage}).  We further demonstrate how \texttt{3S} can generate data with shifts which better estimate model performance on shifted data, compared to real data or baselines. \texttt{3S} accommodates both minimal user input (Sec.~\ref{exp-operating}) or some prior knowledge of the target domain (Sec.~\ref{exp:prior_knowledge}).

\section{Related Work} \label{sec:related}
This paper primarily engages with the literature on model testing and benchmarking, synthetic data, and data-centric AI---see Appendix \ref{appx:related_work} for an extended discussion.

\textbf{Model evaluation.} ML models are mostly evaluated on hold-out datasets, providing a measure of aggregate performance \citep{flach2019performance}. 
Such aggregate measures do not account for under-performance on specific subgroups \citep{oakden2020hidden} or assess performance under data shifts \citep{quinonero2008dataset, wiles2021fine}. 

The ML community has tried to remedy these issues by creating better benchmark datasets: either manual corruptions like Imagenet-C \citep{hendrycks2018benchmarking} or by collecting additional real data such as the Wilds benchmark \citep{koh2021wilds}. Benchmark datasets are labor-intensive to collect and evaluation is limited to specific benchmark tasks, hence this approach is not flexible for \emph{any} dataset or task. The second approach is model behavioral testing of specified properties, e.g. see Checklist \citep{ribeiro2020beyond} or HateCheck \citep{rottger2021hatecheck}. Behavioral testing is also labor-intensive, requiring humans to create or validate the tests. In contrast to both paradigms, \texttt{3S} generates synthetic test sets for varying tasks and datasets. 

\textbf{Challenges of model evaluation.} \texttt{3S} aims to mitigate the challenges of model evaluation with limited real test sets, particularly estimating performance for small subgroups or under distributional shifts. We are not the first to address this issue.
$\blacksquare$ \textbf{Subgroups:}  Model-based metrics (MBM \cite{miller2021model}) model the conditional distribution of the predictive model score to enable subgroup performance estimates. \\
$\blacksquare$ \textbf{Distribution shift:} Prior works aim to predict model performance in a shifted target domain using (1) Average samples above a threshold confidence (ATC \cite{gargleveraging}), (2) difference of confidences (DOC \cite{guillory2021predicting}), (3) Importance Re-weighting (IM \cite{chen2021mandoline}). A fundamental difference to \texttt{3S} is that they assume access to \emph{unlabeled data} from the target domain, which is unavailable in many settings, e.g. when studying potential effects of unseen or future shifts. We note that work on robustness to distributional shifts is not directly related, as the goal is to learn a model
robust to the shift, rather than reliably estimating performance of an already-trained model under a shift.

\textbf{Synthetic data.} Improvements in deep generative models have spurred the development of synthetic data for different uses \cite{breugel2023beyond}, including privacy (i.e. to enable data sharing, \citealp{jordon2018pate,assefa2020generating}),  fairness \citep{xu2019achieving,breugel2021decaf}, and improving downstream models \cite{antoniou2017data, dina2022effect, das2022conditional, bing2022conditional}. 
\texttt{3S} provides a completely \emph{different} and unexplored use of synthetic data: improving \emph{testing and evaluation} of ML models. Simulated (CGI-based) and synthetic images have been used previously in computer vision (CV) applications for a variety of purposes --- often to \emph{train} more robust models \cite{wang2019learning,trigueros2021generating}. These 
CV-based methods require additional metadata like lighting, shape, or texture \cite{ruiz2022simulated, khan2019procsy}, which may not be available in practice. Additionally, beyond the practical differences between modalities, the CV methods differ significantly from \texttt{3S} in terms of (i) aim, (ii) approach, and (iii) amount of data---see Table \ref{cv-related_work}, Appendix \ref{appx:related_work}.

\section{Why Synthetic Test Data Can Improve Evaluation} \label{sec:formulation}

\subsection{Why Real Data Fails}\label{sec:real-data-fails}
\textbf{Notation.}
Let $\mathcal{X}$ and $\mathcal{Y}$ be the feature and label space, respectively. The random variable $\tilde{X}=(X,Y)$ is defined on this space, with distribution $p(X,Y)$. We assume access to a trained black-box prediction model $f:\mathcal{X}\rightarrow \mathcal{Y}$ and test dataset $\D_{test,f}= \{x_i, y_i\}_{i=1}^{N_{test,f}}\overset{iid}{\sim}~p(X,Y)$. Importantly, we do \emph{not} assume access to the training data of the predictive models, $\cD_{train,f}$. Lastly, let $M:\mathcal{Y}\times\mathcal{Y}\rightarrow \mathbb{R}$ be a performance metric.

\textbf{Real data does not suffice for estimating granular performance (P1).}
In evaluating performance of $f$ on subgroups, we assume that a subgroup $\mathcal{S}\subset \mathcal{X}$ is given. The usual approach to assess subgroup performance is simply restricting the test set $\cD_{test,f}$ to the subspace $\mathcal{S}$:
\begin{equation}
\label{eq:traditional_evaluation}
    A(f; \cD_{test,f}, \mathcal{S}) = \frac{1}{\cD_{test,f}\cap\mathcal{S}} \sum_{(x,y)\in \cD_{test,f}\cap\mathcal{S}} M(f(x), y).
\end{equation}
This is an unbiased and consistent estimate of the true performance, i.e. for increasing $|\mathcal{D}_{test,f}|$ this converges to the true performance $A^*(f;p,\mathcal{S})=\mathbb{E}[M(f(X),Y)|(X,Y)\in \mathcal{S}]$. 

However, what happens when $|\cD_{test,f}\cap\mathcal{S}|$ is small? The variance $\mathbb{V}_{\cD_{test,f}\sim p} A(f, \cD_{test,f}; \mathcal{S})$ will be large. 
In other words, the expected error of our performance estimates becomes large.

\begin{example} \label{example:toy}
To highlight how this affects Eq. \ref{eq:traditional_evaluation}, let us assume the very simple setting in which $p(Y|X)=p(Y)$. Despite $Y$ being independent of $X$, Eq. \ref{eq:traditional_evaluation} is not---the smaller $\mathcal{S}$, the higher the expected error in our estimate. In particular, for $|\mathcal{S}|\rightarrow 0$ (w.r.t. $p(X)$), almost certainly $|\cD_{test,f}\cap \mathcal{S}|=\emptyset$, making Eq. \ref{eq:traditional_evaluation} meaningless.
\end{example}

As a result, we find that the smaller our subgroup $\mathcal{S}$, the harder it becomes to measure model $f$. At the same time, ML models have been known to perform less consistently on small subgroups \cite{oakden2020hidden,suresh2019framework,cabrera2019fairvis,cabreradiscovery}, hence being able to measure performance on these groups would be most useful. Finally, by definition, minorities are more likely to form these small subgroups, and they are the most vulnerable to historical bias and resulting ML unfairness. In other words, \textbf{model evaluation is most inaccurate on the groups that are most vulnerable and for which the model \emph{itself} is most unreliable.}

\textbf{Real test data fails for distributional shift (P2).}
If we do not take into account shifts between the test and deployment distribution, trivially the test performance will be a poor measure for real-world performance---often leading to overestimated performance \cite{pianykh2020continuous,quinonero2008dataset,koh2021wilds}. Nonetheless, even if we do plan to consider shifts in our evaluation, for example by using importance weighting or rejection sampling based on our shift knowledge, real test data will give poor estimates. The reason is the same as before; in the regions that we oversample or overweight, there may be few data points, leading to high variance and noisy estimates. As expected, problems are most pervasive for large shifts, because these require higher reweighting or oversampling of individual points. 

\subsection{Why Generative Models Can Help}
\begin{wrapfigure}{r}{0.46\textwidth}
 \vspace{-4mm}
    \begin{subfigure}{0.2\textwidth}
    \centering
    \includegraphics[width=\textwidth]{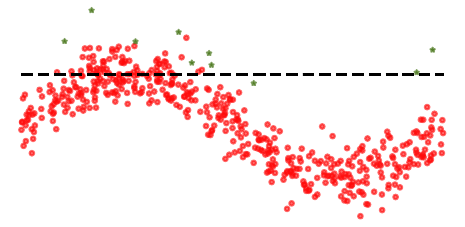}
    \caption{Original data}
    \label{fig:magic_before}
    \end{subfigure}
    \quad\quad
    \begin{subfigure}{0.2\textwidth}
    \centering
    \includegraphics[width=\textwidth]{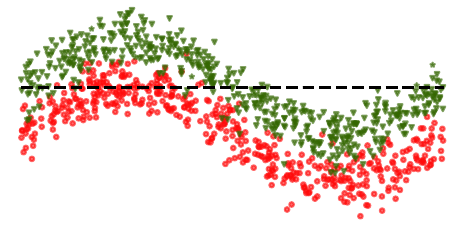}
    \caption{With synthetic data}
    \end{subfigure}
    \caption{\textbf{Illustration why synthetic data can give more accurate estimates using the same real samples.} Assume we want to evaluate $f$ (decision boundary$=$dashed line), which aims to discriminate between $Y=1$ (green stars) and $Y=0$ (red circles). Due to the low number of samples for $Y=1$, evaluating $f$ using the test set alone (Eq. \ref{eq:traditional_evaluation}) has a high variance. On the other hand, a generative model can learn the manifold from $\cD_{test,f}$, and generate additional data for $Y=1$ by only learning the offset (b, green triangles). This can reduce variance of the estimated performance of $f$.}
    \vspace{-1mm}
     \rule{0.45\textwidth}{0.1pt}
    \label{fig:magic}
\vspace{-5mm}\end{wrapfigure}

We have seen there are two problems with real test data. Firstly, the more granular a metric, the higher the noise in Eq. \ref{eq:traditional_evaluation}---even if the distribution is well-behaved like in Example \ref{example:toy}. Secondly, we desire a way to emulate shifts, and simple reweighing or sampling of real data again leads to noisy estimates. 

Generative models can provide a solution to both problems. As we will detail in the next section, instead of using $\cD_{test,f}$, we use a generative model $G$ trained on $\cD_{test,f}$ to create a large synthetic dataset $\cD_{syn}$ for evaluating $f$. We can induce shifts in the learnt distribution, thereby solving problem 2. 
It also solves the first problem. A generative model aims to approximate $p(X,Y)$, which we can regard as effectively interpolating $p(Y|X)$ between real data points to generate more data within $\mathcal{S}$. 

It may seem counterintuitive that $|A^*-A(f;\cD_{syn}, \mathcal{S})|$ would ever be lower than $|A^*-A(f;\cD_{test,f}, \mathcal{S})|$, after all $G$ is trained on $|\cD_{test,f}|$ and there is no new information added to the system. However, even though the generative process may be imperfect, we will see that the noise of the generative process can be significantly lower than the noisy real estimates (Eq. \ref{eq:traditional_evaluation}). Secondly, a generative model can learn implicit data representations \cite{antoniou2017data}, i.e. learn relationships within the data (e.g. low-dimensional manifolds) from the entire dataset and transfer this knowledge to small $\mathcal{S}$. We give a toy example in Fig. \ref{fig:magic}. This motivates modeling the full data distribution $p(X,Y)$, not just $p(Y|X)$.

Of course, synthetic data cannot always help model evaluation, and may in fact induce noise due to an imperfect $G$. Through the inclusion of uncertainty estimates, we promote trustworthiness of results (Sec \ref{sec:method_subset}), and when we combine synthetic data with real data, we observe almost consistent benefits (see Sec. \ref{sec:experiments}). In Section \ref{sec:discussion} we include limitations.

\section{Synthetic Data for Subgroup and Shift Testing} \label{sec:method}

\subsection{Using Deep Generative Models for Synthetic Test Data}
\label{sec:method_subset}
\begin{figure*}[!t]
    \vspace{-5mm}
    \captionsetup[subfigure]{labelformat=empty}
    \centering
    \begin{subfigure}[b]{0.425\textwidth}
    \includegraphics[width=\textwidth]{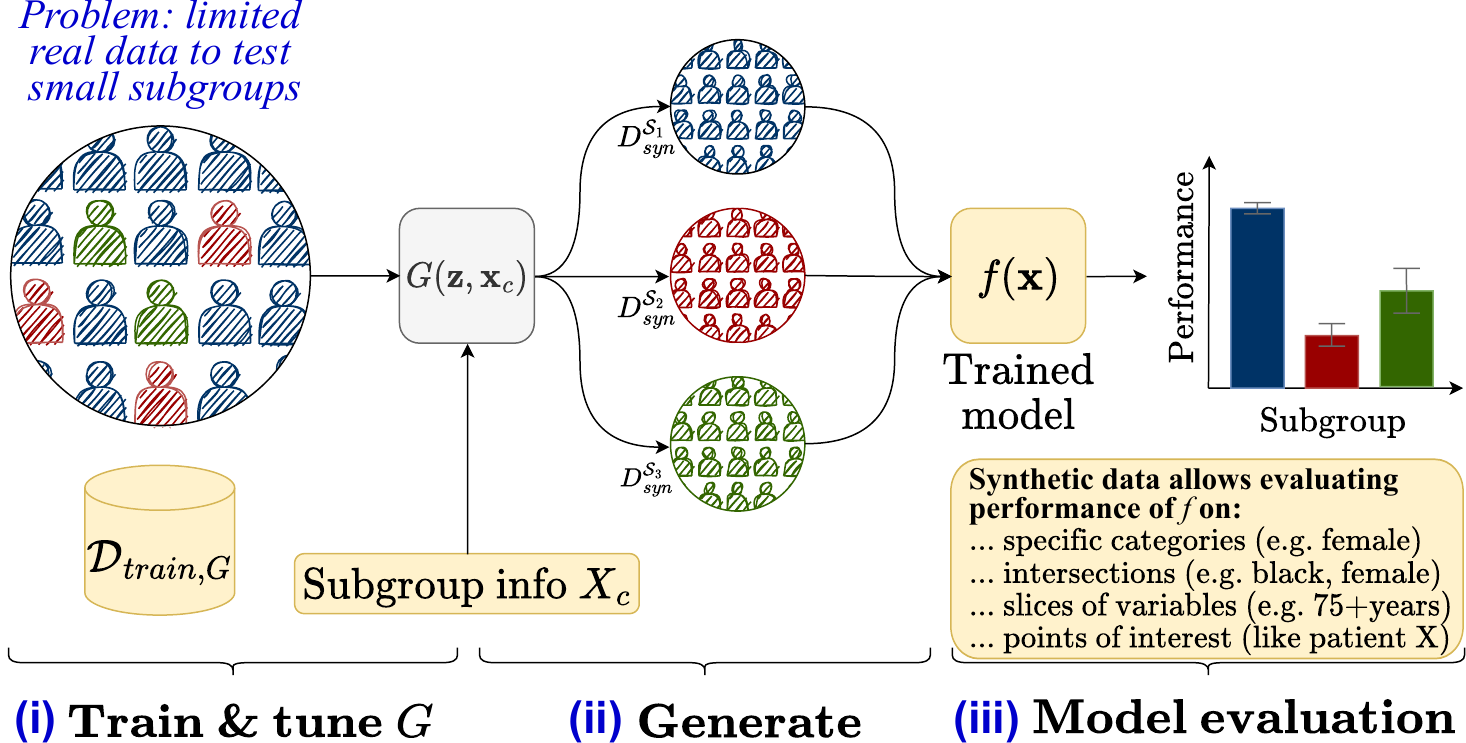}  
    \caption{(P1) Reliable granular evaluation}
    \end{subfigure}
    \quad
    \begin{subfigure}[b]{0.425\textwidth}
    \includegraphics[width=\textwidth]{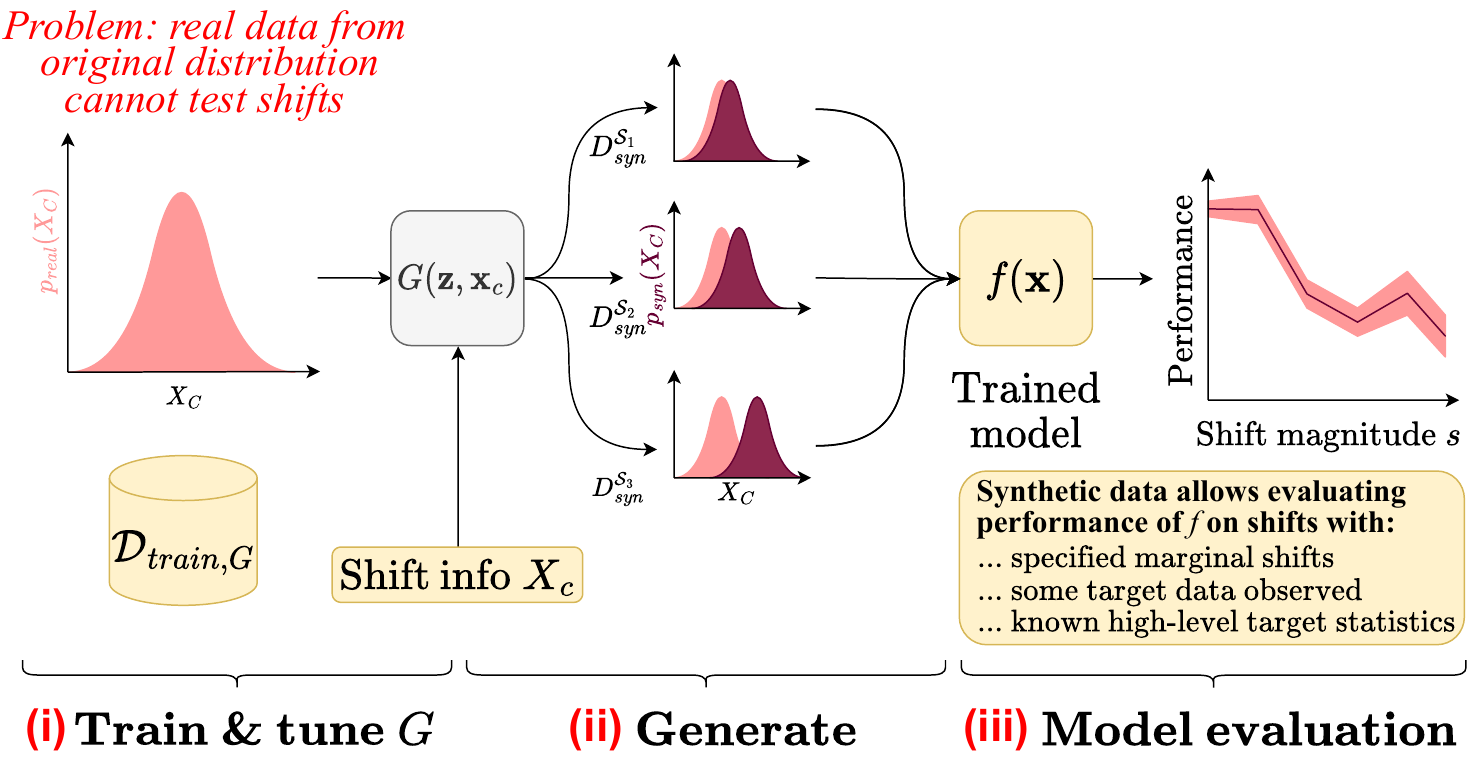}    
    \caption{(P2) Sensitivity to distributional shifts}
    \end{subfigure}
    \caption{
  3S is a framework for evaluating model performance using synthetic data generators. 
  It has three phases: training the generative model, generating synthetic data, and model evaluation. Firstly, 3S enables \textcolor{blue}{(P1) \emph{ granular evaluation}} when there is \textcolor{blue}{(i)} limited real test data in small subgroups, by \textcolor{blue}{(ii)} generating synthetic data conditional on subgroup information $X_c$, thereby \textcolor{blue}{(iii)} permitting more reliable model evaluation even on small subgroups. Secondly, 3S enables assessment of \textcolor{red}{(P2) \emph{distributional shifts}} when  \textcolor{red}{(i)} the real test data does not reflect shifts, by  \textcolor{red}{(ii)} generating synthetic data conditional on marginal shift information of features $X_c$, thereby \textcolor{red}{(iii)} quantifying model sensitivity to distributional shift. Required inputs are denoted in yellow.}
\label{fig:idea}
 \vspace{-4mm}
\end{figure*}

We reiterate that our goal is to generate test datasets that provide insight into model performance on a granular level (P1) and for shifted distributions (P2). We propose using synthetic data for testing purposes, which we refer to as \emph{\texttt{3S}}-testing. This has the following workflow (Fig. \ref{fig:idea}): (1) train a (conditional) generative model on the real test set, (2) generate synthetic data conditionally on the subgroup or shift specification, and (3) evaluate model performance on the generated data, $A(f;\cD_{syn}, \mathcal{S})$. This procedure is flexible w.r.t. the generative model, but a conditional generative model is most suitable since it allows precise generation conditioned on subgroup or shift information out-of-the-box. Throughout this paper we use CTGAN \cite{xu2019modeling} as the generative model---see Appendix \ref{appx:generative_model} for other generative model results and more details on the generative training process.

\textbf{Estimating uncertainty.} Generative models are not perfect, leading to imperfect synthetic datasets and inaccurate \texttt{3S} estimates. To provide insight into the trustworthiness of its estimates, we quantify the uncertainty in the \texttt{3S} generation process through an ensemble of generative models \cite{breugel2023dge}, similar in vain to Deep Ensembles \cite{lakshminarayanan2017}. We (i) initialize and train $K$ generative models independently, (ii) generate synthetic datasets $\{\cD_{syn}^k\}_{k=1}^K$, and (iii) evaluate model $f$ on each. The final estimates are assumed Gaussian, with statistics given by the sample mean and variance:
\begin{equation}
    \hat{\mu}(A) = \frac{1}{K}\sum_{k=1}^K A(f;\cD_{syn}^k, \mathcal{S}) \text{ and } \hat{\sigma}^2(A) = \frac{1}{K-1}\sum_{k=1}^K (A(f;\cD_{syn}^k, \mathcal{S}) - \hat{\mu}(A))^2,
\end{equation}
which can be directly used for constructing a prediction interval. In Sec. \ref{exp:coverage}, we show this provides high empirical coverage of the true value compared to alternatives.

\textbf{Defining subgroups.} The actual definition of subgroups is flexible. Examples include a specific category of one feature (e.g. female), intersectional subgroups \cite{crenshaw1989demarginalizing} (e.g. black, female), slices from continuous variables (e.g. over 75 years old), particular points of interest (e.g. people similar to patient X), and outlier groups. In Appendix \ref{appx:additional_experiments}, we elaborate on some of these further.

\subsection{Generating Synthetic Test Sets with Shifts} \label{sec:method_shifts_general}
Distributional shifts between training and test sets are not unusual in practice \cite{quinonero2008dataset,amodei2016concrete, wu2021online} and have been shown to degrade model performance \cite{pianykh2020continuous,koh2021wilds,cohen2021problems,ji2023large}. Unfortunately, often there may be no or insufficient data available from the shifted target domain. 

\textbf{Defining shifts.} In some cases, there is prior knowledge to define shifts. For example, covariate shift \cite{shimodaira2000improving, moreno2012unifying} focuses on a changing covariate distribution $p(X)$, but a constant label distribution $p(Y|X)$ conditional on the features. Label (prior probability) shift \cite{saerens2002adjusting, moreno2012unifying} is defined vice versa, with fixed $p(X|Y)$ and changing $p(Y)$.\footnote{Concept drifts are beyond the scope of this work.} 

Generalizing this slightly, we assume only the marginal of some variables changes, while the distribution of the other variables conditional on these variables does not. Specifically, let $c\subset \{1,...,|\tilde{X}|\}$ denote the indices of the features or targets in $\Tilde{X}$ of which the marginal distribution may shift. Equivalent to the covariate and label shift literature, we assume the distribution $p(\tilde{X}_{\Bar{c}}|\tilde{X}_{c})$ remains fixed ($\Bar{c}$ denoting the complement of $c$).\footnote{This reduces to label shift---$p(X|Y)$ constant but $p(Y)$ changed)---and covariate shift---$p(Y|X)$ constant but $p(X)$ changed)---for $\tilde{X}_c=Y$ and $\tilde{X}_c = X$, respectively.} Let us denote the marginal's shifted distribution by $p^s(\tilde{X}_{c})$ with $s$ the shift parameterisation, with $p^{0}(\tilde{X}_{c})$ having generated the original data. The full shifted distribution is $p(\tilde{X}_{\Bar{c}}|\tilde{X}_{c})p^s(\tilde{X}_{c})$.

\textbf{Example: single marginal shift.} Without further knowledge, we study the simplest such shifts first: only a single $\tilde{X}_i$'s marginal is shifted. Letting $p^0(\tilde{X}_i)$ denote the original marginal, we define a family of shifts $p^s(\tilde{X}_i)$ with $s\in \mathbb{R}$ the shift magnitude. To illustrate, we choose a mean shift for continuous variables, $p^s(\tilde{X}_i)=p^0(\tilde{X}_i-s)$, and a logistic shift for any binary variable, $\text{logit } p^s(\tilde{X}_i=1) = \text{logit } (p^s(\tilde{X}_i))-s$.\footnote{We consider any categorical variable with $m$ classes using $m$ different shifts of the individual probabilities, scaling the other probabilities appropriately.}
As before, we assume $p(\tilde{X}_{\neg i}|\tilde{X}_i)$ remains constant. This can be repeated for all $i$ and multiple $s$ to characterize the sensitivity of the model performance to distributional shifts. The actual shift can be achieved using any conditional generative model, with the condition given by $\tilde{X}_i$.

\textbf{Incorporating prior knowledge on shift.} 
In many scenarios, we may want to make stronger assumptions about the types of shift to consider. Let us give two use cases. First, we may acquire high-level statistics of some variables in the target domain---e.g. we may know that the age in the target domain approximately follows a normal distribution $\mathcal{N}(50,10)$. In other cases, we may actually acquire data in the target domain for some basic variables (e.g. age and gender), but not all variables. In both cases, we can explicitly use this knowledge for sampling the shifted variables $\tilde{X}_c$, and subsequently generating $\tilde{X}_{\Bar{c}}|\tilde{X}_c$---e.g. sample (case 1) age from $N(50,10)$ or (case 2) (age, gender) from the target dataset. Variables $\tilde{X}_{\Bar{c}}|\tilde{X}_c$ are generated using the original generator $G$, trained on $\cD_{test,f}$. 

\textbf{Characterizing sensitivity to shifts.} This gives the following recipe for testing models under shift. Given some conditional generative model $G$, we (i) train $G$ to approximate $p(X_{\bar{c}}|X_c)$, (2) choose a shifted distribution $p^s(\tilde{X}_c)$---e.g. a marginal mean shift of the original $p^0(X_c)$ (Section \ref{exp-operating}), or drawing $x_c$ samples from a secondary dataset (Section \ref{exp:prior_knowledge}); (3) draw samples $x_c$, and subsequently use $G$ to generate the rest of the variables $X_{\bar{c}}$ conditional on these drawn samples---together giving$\cD_{syn}^s$; and (4) evaluate downstream models; (5) Repeat (2-4) for different shifts (e.g. shift magnitudes $s$) to characterize the sensitivity of the model to distributional shifts. 

\textbf{More general shifts.}
Evidently, marginal shifts can be generalised. We can consider a family of shifts $\mathcal{T}$ and test how a model would behave under different shifts in the family. Let $\mathcal{P}$ be the space of distributions defined on $\mathcal{\tilde{X}}$. We test models on data from $T(p)(\tilde{X})$, for all $T\in \mathcal{T}$, with $T:\mathcal{P}\rightarrow \mathcal{P}$. For example, for single marginal shifts this corresponds to $T^s(p)(\tilde{X}) = p^s(\tilde{X}_i)p(\tilde{X}_{\neg i}|X_i)$. The general recipe for testing models under general shifts then becomes as follows. Let $G$ be some generative model, we (1) Train generator $G$ on $\cD_{train,G}$ to fit $p(X)$; (2) Define family of possible shifts $\mathcal{T}$, either with or without background knowledge; Denote shift with magnitude $s$ by $T^s$; (3) Set $s$ and generate data $\cD_{syn}^s$ from $T^s(p)$; (4) Evaluate model on $\cD_{syn}^s$; (5) Repeat steps 2-4 for different families of shifts and magnitudes $s$.

\section{Use Cases of \texttt{3S} Testing} \label{sec:experiments}
We now demonstrate how \texttt{3S} satisfies (\textbf{P1}) Granular evaluation and (\textbf{P2}) Distributional shifts. We re-iterate that the aim throughout is to estimate the true prediction performance of the model $f$ as closely as possible.  We tune and select the generative model based on Maximum Mean Discrepancy \cite{gretton2012kernel}, see Appendix \ref{appx:generative_model}.  
We describe the experimental details, baselines, and datasets for each experiment further in Appendix \ref{appx:experimental_details} \footnote{Code for use cases found at: 
\url{https://github.com/seedatnabeel/3S-Testing} or \\ \url{https://github.com/vanderschaarlab/3S-Testing}}.

\subsection{(P1) Granular Evaluation}\label{exp-d1}

 \subsubsection{Correctness of Subgroup Performance Estimates}\label{exp:subgroup-correct}
 
\textbf{Goal.} This experiment assesses the value of synthetic data when evaluating model performance on minority subgroups. The challenge with small subgroups is that the conventional paradigm of using a hold-out evaluation set might result in high variance estimates due to the small sample size. 

\textbf{Datasets.} We use the following five real-world medical and finance datasets: Adult \cite{uci}, Covid-19 cases in Brazil \cite{baqui2020ethnic}, Support \cite{knaus1995support}, Bank \cite{jesus2022turning}, and Drug \cite{fehrman2017five}. These datasets have varying characteristics, from sample size to number of features. They also possess representational imbalance and biases, pertinent to \texttt{3S} \cite{cabrera2019fairvis,cabreradiscovery}: \textbf{\textcolor{ForestGreen}{\textcircled{1}}} \emph{Minority subgroups}: we evaluate the following groups which differ in proportional representation - \emph{Adult}: Race; \emph{Covid}: Ethnicity; \emph{Bank}: Employment; \emph{Support}: Race; \emph{Drug}: Ethnicity. 
\textbf{\textcolor{ForestGreen}{\textcircled{2}}} \emph{Intersectional subgroups}: we evaluate intersectional subgroups \cite{crenshaw1989demarginalizing} (e.g. black males or young females)--- see Appendix \ref{appx:modelreport}, intersectional model performance matrix.

\textbf{Set-up.} We evaluate the estimates of subgroup performance for trained model $f$ using different evaluation sets. We consider two baselines: (1) $\cD_{test, f}$: a typical hold-out test dataset and (2) Model-based metrics (MBM) \cite{miller2021model}. MBM uses a bootstrapping approach for obtaining multiple test sets. We compare the baselines to \texttt{3S} testing datasets, which generate data to balance the subgroup samples: (i) \emph{\texttt{3S}} ($\cD_{syn}$): synthetic data generated by $G$, which is trained on $\cD_{test,f}$ and (ii) \emph{\texttt{3S}+} ($\cD_{syn} \cup \cD_{test, f}$): test data \emph{augmented} with the synthetic dataset. 

For some subgroup $\mathcal{S}$, each test set gives an estimated model performance $A(f;\cD_{\cdot}, \mathcal{S})$, which we compare to a pseudo-oracle performance $A(f;\cD_{oracle}, \mathcal{S})$: the oracle is the performance of $f$ evaluated on a large unseen real dataset $\cD_{oracle}\sim p(X,Y)$, where $|\cD_{oracle}|\gg|\cD_{test, f}|$.  As outlined above the subgroups are as follows: (i) Adult: Race, (ii) Drug: Ethnicity, (iii) Covid: Ethnicity (Region), (iv) Support: Race, (v) Bank: Employment status. 

We evaluate the reliability of the different performance estimates based on their Mean Absolute Error (MAE) relative to the Oracle predictive accuracy estimates. We desire low MAE such that our estimates match the oracle.

\begin{figure}
    \centering
    \vspace{-0mm}
    \includegraphics[width=1\textwidth]{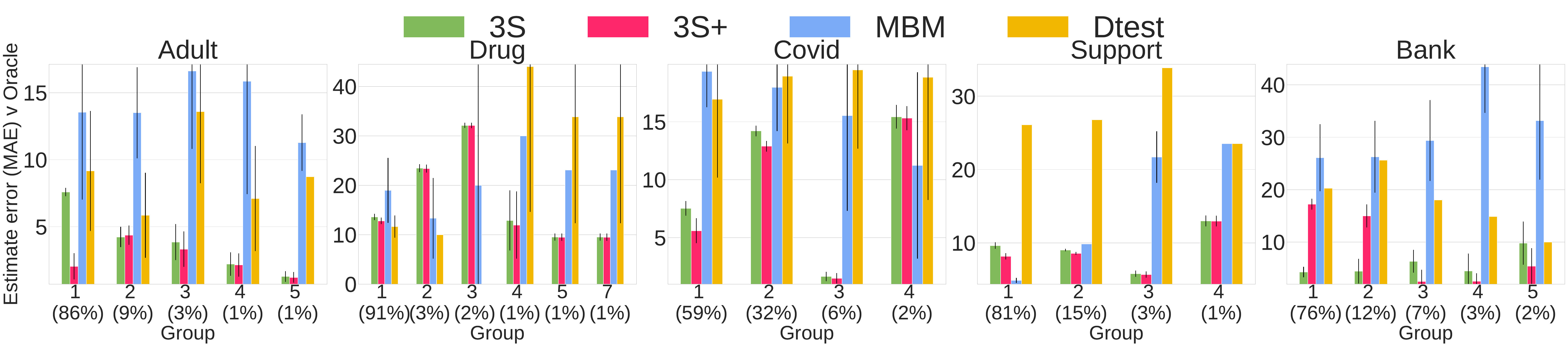}
    \caption{Assessing the reliability of performance estimates based on mean absolute error of the predicted performance estimate to the performance evaluated by oracle ($\downarrow$ better). \texttt{3S} better approximates true estimates compared to both baselines. \texttt{3S}+ enjoys the best of both worlds by combining synthetic and real data.  We evaluate an RF model, with other model classes shown in the Appendix. 
    }
    \label{fig:exp1}
    \vspace{-4mm}
\end{figure}

\textbf{Analysis.} Fig.~\ref{fig:exp1} illustrates across the 5 datasets that the \texttt{3S} synthetic data (red, green) closely matches estimates on the Oracle data. i.e. lower MAE vs baselines. In particular, for small subgroups (e.g. racial minorities), \texttt{3S} provides a more accurate evaluation of model performance (i.e. with estimates closer to the oracle) compared to a conventional hold-out dataset ($\cD_{test, f}$) and MBM. 

In addition, \texttt{3S} estimates have reduced standard deviation. Thus, despite \texttt{3S} using the same (randomly drawn test set) $\cD_{test,f}$ to train its generator, its estimates are more robust to this randomness. The results highlight an evaluation pitfall of the standard hold-out test set paradigm: the estimate's high variance w.r.t. the drawn $\cD_{test,f}$ could lead to potentially misleading conclusions about model performance in the wild, since an end-user only has access to a single draw of $\cD_{test,f}$. e.g., we might incorrectly overestimate the true performance of minorities. The use of synthetic data solves this.

Next, we move beyond single-feature minority subgroups and show that synthetic data can also be used to evaluate performance on \textbf{intersectional groups} --- subgroups with even smaller sample sizes due to the intersection. \texttt{3S} performance estimates on 2-feature intersections are shown in Appendix \ref{appx:modelreport}. Intersectional performance matrices provide model developers more granular insight into where they can improve their model most, as well as inform users how a model may perform on intersections of groups (especially important to evaluate sensitive intersectional subgroups).\footnote{N.B. low-performance estimates by \texttt{3S} only indicate poor model performance; this does not necessarily imply that the data itself is biased for these subgroups. However, it could warrant investigating potential data bias and how to improve the model.} 
Appendix \ref{appx:modelreport} further illustrates how these intersectional performance matrices can be used as part of model reports. 

We evaluate the intersectional performance estimates of \texttt{3S} and baseline $\cD_{test,f}$ using the MAE of the performance matrices w.r.t. the oracle, averaged across 3 models (i.e, RF, GBDT, MLP). The error of \textbf{\texttt{3S} (11.90 $\pm$ 0.19)} is significantly lower than $\cD_{test,f}$ \textbf{(20.29 $\pm$ 0.14)}, hence demonstrating \texttt{3S} provides more reliable intersectional estimates.

\textbf{\textcolor{ForestGreen}{Takeaway.}} Synthetic data provides more accurate performance estimates on small subgroups compared to evaluation on a standard test set. This is especially relevant from a representational bias and fairness perspective---allowing more accurate  performance estimates on minority subgroups.

\subsubsection{Reliability through Confidence Intervals}\label{exp:coverage}

\begin{wrapfigure}{r}{0.33\textwidth}
\vspace{-4mm}
\centering
 \includegraphics[width=0.33\textwidth]{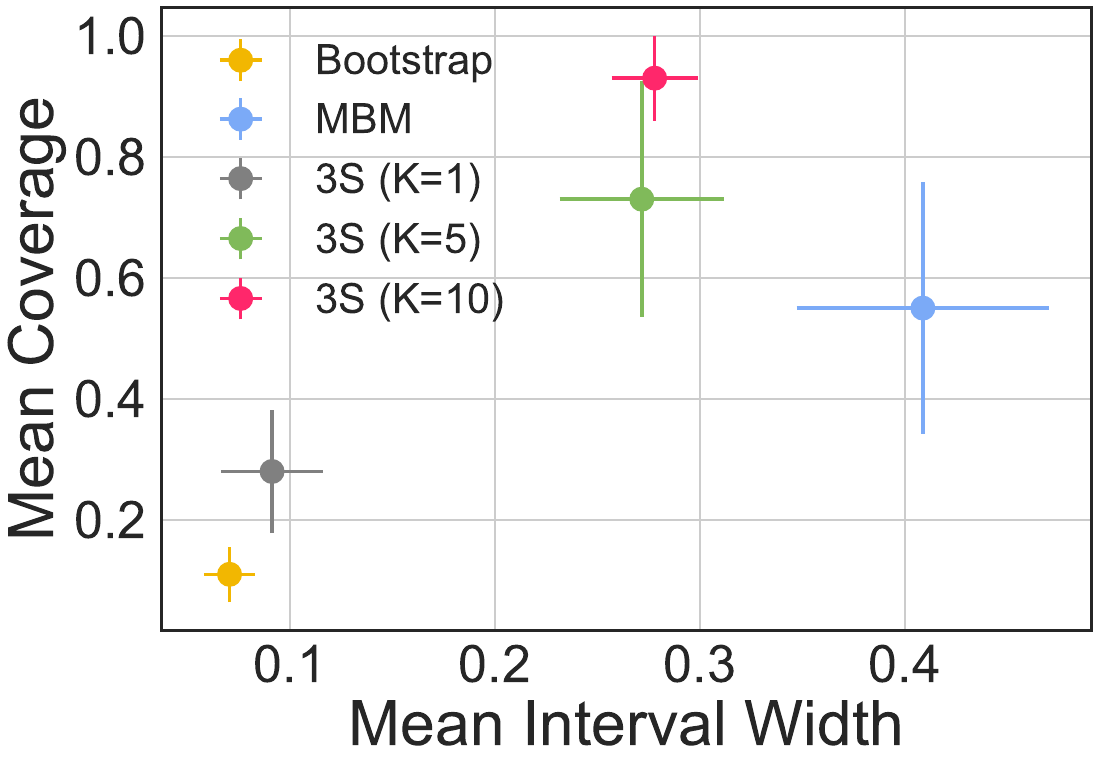}
 \vspace{-5mm}
\caption{3S is more reliable than baselines with higher coverage, lower widths --- with less variability.}\label{coverage}
\vspace{-1mm}
\rule{0.33\textwidth}{0.1pt}
\vspace{-6mm}
\end{wrapfigure}

\textbf{Goal.} 
In Fig.~\ref{fig:exp1}, we see that all methods, including \texttt{3S}, have errors in some cases, which warrants the desire to have confidence intervals at test time.
\texttt{3S} uses a deep generative ensemble to provide uncertainty estimates at test-time---see Sec. \ref{sec:method_subset}.

\textbf{Set-up.} We assess coverage over 20 random splits and seeds for \texttt{3S} vs the baselines (B1) bootstrapping \cite{efron1992bootstrap} confidence intervals for $\cD_{test,f}$ and (B2) MBM: which itself uses bootstrapping for the predictive distribution. For \texttt{3S}, we assess a Deep Generative Ensemble with $K=1,5,10$  randomly initialized generative models.  For each method, we take the average estimate $\pm$ 2 standard deviations. We evaluate the intervals based on the following two metrics defined in \cite{kompa2021empirical,seedat2022data,navratil2020uncertainty}:  (i) Coverage =  $\mathbb{E}\left[1_{x_{i} \in\left[l_{i}, r_{i} \right]}\right] $ 
(ii) Width =  $\mathbb{E}[|r_{i}-l_{i}|]$. Coverage measures how often the true label is in the prediction region, while width measures how specific that prediction region is. In the ideal case, we have high coverage with low width.
See Appendix \ref{appx:experimental_details} for more details. 

\textbf{Analysis.} Fig. \ref{coverage} shows the mean test set coverage and width averaged over the five datasets. \texttt{3S} (with K=5 and K=10) is more reliable, attaining higher coverage rates with lower width compared to baselines. In addition, the variability with \texttt{3S} is much lower for both coverage and width. We note that this comes at a price: computational cost scales linearly with $K$. For fair comparison, we set $K=1$ in the rest of the paper.

\textbf{\textcolor{ForestGreen}{Takeaway.}} \texttt{3S} includes uncertainty estimates at test time that cover the true value much better than baselines, allowing practitioners to decide when (not) to trust \threeS performance estimates.

\vspace{-2mm}

\subsection{(P2) Sensitivity to Distributional Shifts}\label{exp-d2}
\vspace{-2mm}
ML models deployed in the wild often encounter data distributions differing from the training set. We simulate distributional shifts to evaluate model performance under different potential post-deployment conditions. We examine two setups with varying knowledge of the potential shift.

\vspace{-2mm}
\subsubsection{No Prior Information}\label{exp-operating}
\textbf{Goal.} Assume we have no prior information for the (future) model deployment environment. In this case, we might still wish to stress test the sensitivity for different potential operating conditions, such that a practitioner understands model behavior under different conditions, which can guide as to when the model can and cannot be used. We wish to simulate distribution shifts using synthetic data and assess if it captures true performance.

\textbf{Set-up.}
We consider shifts in the marginal of some feature $X_i$, keeping $p(\tilde{X}_{\neg i}|\tilde{X}_i)$ fixed (see Sec.~\ref{sec:formulation}). For instance, a shift in the marginal $X_i$'s mean (see Sec.~\ref{sec:method_shifts_general}).  To assess performance for different degrees of shift, we compute three shift buckets around the mean of the original feature distribution: large negative shift from the mean (\textbf{-}), small negative/positive shift from the mean (\textbf{$\pm$}), and large positive shift from the mean (\textbf{+}). 

We define each in terms of the feature quantiles. We generate uniformly distributed shifts (between min(feature) and max(feature)). Any shift that shifts the mean to less than Q1 is (\textbf{-})
, any shift that shifts the mean to more than Q3 is (\textbf{+}) and any shift in between is (\textbf{$\pm$}). 

As before, we compare estimated accuracy w.r.t. a pseudo-oracle test set. We compare two baselines: (i) Mean-shift (MS) and (ii) Rejection sampling (RS);  both applied to the real data.

\textbf{Analysis.} Table \ref{table:shift_buckets} shows the potential utility of synthetic data to more accurately estimate performance for unknown distribution shifts compared to real data alone. This is seen both with an average lower mean error of estimates, but also across all three buckets. This implies that the synthetic data is able to closely capture the true performance across the range of feature shifts.

\begingroup

\setlength{\tabcolsep}{2pt} 
\renewcommand{\arraystretch}{1.1} 
\vspace{-0mm}
\begin{table}[t]
\caption{Mean error in estimated accuracy across shift quantile buckets for all 5 datasets. The results show \texttt{3S} indeed provides more reliable estimates even for distribution shift. $\downarrow$ is better.}
\scalebox{0.8}{\begin{tabular}{c|cccc|cccc|cccc|cccc|cccc}
\toprule
   & \multicolumn{4}{c|}{Adult}                                                           & \multicolumn{4}{c|}{Support}                                                         & \multicolumn{4}{c|}{Bank}                                                            & \multicolumn{4}{c|}{Drug}                                                            & \multicolumn{4}{c}{SEER}                                                                               \\ \hline\hline
      & Mean  & -   & $\pm$   & +   & Mean  & -    & $\pm$    & +   & Mean & -   & $\pm$   & +   & Mean & -   & $\pm$   & +   & Mean & -   & $\pm$   & +   \\ \hline \hline
\bf 3S & \bf 2.6                       & \bf  2.2                    & \bf 
 1.8                     & \bf  3.9  & \bf  2.0                       & \bf  2.6                     & \bf  2.0                     & \bf  1.1  & \bf  5.4                       & \bf  3.7                     & \bf  3.6                     & \bf  6.7  & \bf  5.6                       & \bf  5.7                     & \bf  4.4                     & \bf  7.8  & \bf 2.7                       & 5.5                     & \bf  3.0                     & \bf  2.0  \\ \hline
MS & 5.9                       & 5.2                    & 5.6                     & 6.9  & 19.3                      & 22.9                    & 18.5                    & 15.2 & 18.6                      & 18.6                    & 19.9                    & 17.9 & 18.5                      & 19.6                    & 19.0                    & 16.3 & 3.3                       & \bf  2.6                     & 3.9                     & 3.2  \\ \hline
RS & 15.9                      & 10.5                     & 17.9                    & 18.6 & 25.1                      & 27.8                    & 24.2                    & 22.3 & 18.9                      & 18.9                    & 20.1                    & 18.3 & 20.1                      & 21.7                    & 21.0                    & 18.2 & 20.0                      & 20.3                    & 23.6                    & 18.6 \\ 
\bottomrule
\end{tabular}}
\label{table:shift_buckets}
\vspace{-3mm}
\end{table}

\endgroup

\textbf{\textcolor{ForestGreen}{Takeaway.}}
Synthetic data can be used to more accurately characterize model performance across a range of possible distributional shifts.

\subsubsection{Incorporating Prior Knowledge on Shift} 
\label{exp:prior_knowledge}
\begin{wrapfigure}{r}{0.66\textwidth}
\vspace{-3mm}
\centering
    \begin{subfigure}[t]{.3\textwidth}
    \includegraphics[width=\textwidth]{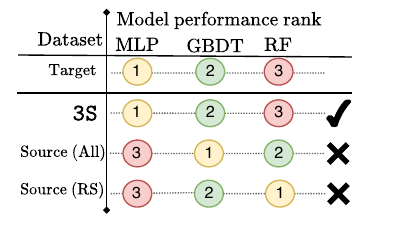}
        \vspace{-6mm}
        \caption{\footnotesize{Model performance rank compared.}}\label{fig:exp3a}
    \end{subfigure}%
    \hspace{3mm}
    \begin{subfigure}[t]{.33\textwidth}
        \includegraphics[width=\textwidth]{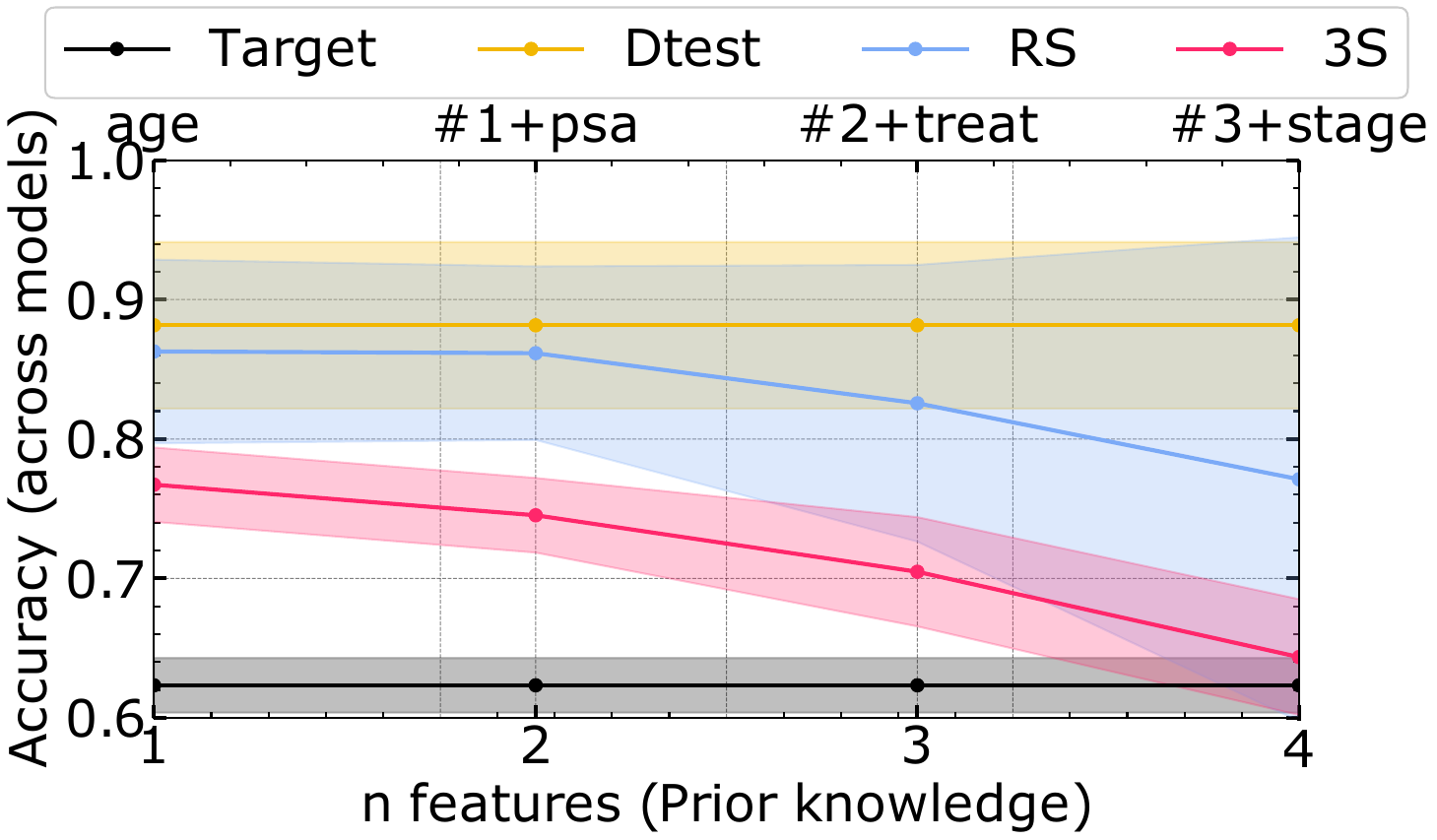}
        \vspace{-6mm}
        \caption{\footnotesize{Average accuracy vs increased prior knowledge}}\label{fig:exp3b}
    \end{subfigure}%
     \vspace{-0mm}
    \caption{Incorporating prior knowledge of the shift. (a) $\cD_{syn}$ matches the performance rank of the true target domain, which can help to select the best model to use in the target domain, and (b) $\cD_{syn}$ better estimates target domain performance compared to baselines. Performance improves and gets closer as more prior knowledge is incorporated via added features. \textit{Points are connected to highlight trends.}}\label{fig:exp3}
    \vspace{-1mm}
    \rule{0.63\textwidth}{.5pt}
\vspace{-3mm}
\end{wrapfigure}

\textbf{Goal.} Consider the scenario where we have \textit{some} knowledge of the shifted distribution and wish to estimate target domain performance. Specifically, here we assume we only have access to the feature marginals in the form of high-level info from the target domain, e.g. age (mean, std) or gender (proportions). We sample from this marginal and generate the other features conditionally (Sec.~\ref{sec:method_shifts_general}).  
\textbf{Set-up.}

We use datasets SEER (US) \cite{seer} and CUTRACT (UK) \cite{cutract}, two real cancer datasets with the same features, but with shifted distributions due to coming from different countries. We train models $f$ and $G$ on the source domain (USA). We then wish to estimate likely model performance in the shifted target domain (UK). We assume access to information from $n$ features in the target domain (features $X_c$), sample $X_c$ from this marginal, conditionally generate $X_{\Bar{c}}|X_c$. We estimate performance with $\cD_{syn}$. 

We use the CUTRACT dataset (Target) as the ground truth to validate our estimate. As baselines, we use estimates on the source test set, along with \emph{Source Rejection Sampling (RS)}, which achieves a distributional shift through rejection sampling the source data using the observed target features. We also compare to baselines which assume access to \textbf{more information than} \texttt{3S}, i.e. access to \emph{full} unlabeled data from the target domain and hence have an advantage over \texttt{3S} when predicting target domain performance. We benchmark ATC \cite{gargleveraging}, DOC \cite{guillory2021predicting} and IM \cite{chen2021mandoline}. Details on all baselines are in Appendix \ref{appx:experimental_details}.
Note, we repeat this experiment in Appendix \ref{appx:additional_experiments} for \emph{Covid-19 data} \cite{baqui2020ethnic}, where there is a shift between Brazil's north and south patients.

\begin{wraptable}{r}{0.66\textwidth}
\begingroup

\setlength{\tabcolsep}{4pt} 
\renewcommand{\arraystretch}{1.1} 
\vspace{-4mm}
 \captionsetup{font=footnotesize}
\caption{\texttt{3S} has lower performance estimate error in target domain for different downstream models. Rows yellow have access to more information than \texttt{3S}, in the form of unlabeled data from the target domain. $\downarrow$ is better}
\vspace{-2mm}
\scalebox{0.85}{
\begin{tabular}{lllllllll}
\toprule
\rowcolor[HTML]{FFFDFA} 
{\color[HTML]{343434} }                            & {\color[HTML]{343434} mean}                           & {\color[HTML]{343434} ada}                            & {\color[HTML]{343434} bag}                            & {\color[HTML]{343434} gbc}                            & {\color[HTML]{343434} mlp}                            & {\color[HTML]{343434} rf}     & {\color[HTML]{343434} knn}                            & {\color[HTML]{343434} lr}                             \\ \hline \hline
\rowcolor[HTML]{FFFDFA} 
{\color[HTML]{343434} 3S-Testing}                  & {\color[HTML]{343434} \bf 0.023}                         & {\color[HTML]{343434} 0.051}                         & {\color[HTML]{343434} \bf 0.012}                         & {\color[HTML]{343434} \bf 0.030}                         & {\color[HTML]{343434} \bf 0.009}                         & {\color[HTML]{343434} \bf 0.015} & {\color[HTML]{343434} \bf 0.020}                         & {\color[HTML]{343434} \bf 0.029}                         \\ \hline
\rowcolor[HTML]{FFFDFA} 
{\color[HTML]{343434} All (Source)}                & {\color[HTML]{343434} 0.258}                         & {\color[HTML]{343434} 0.207}                         & {\color[HTML]{343434} 0.327}                         & {\color[HTML]{343434} 0.207}                         & {\color[HTML]{343434} 0.170}                         & {\color[HTML]{343434} 0.346} & {\color[HTML]{343434} 0.233}                         & {\color[HTML]{343434} 0.211}                         \\ \hline
\rowcolor[HTML]{FFFDFA} 
{\color[HTML]{343434} RS (Source)}                 & {\color[HTML]{343434} 0.180}                         & {\color[HTML]{343434} \bf 0.028}                         & {\color[HTML]{343434} 0.298}                         & {\color[HTML]{343434} 0.096}                         & {\color[HTML]{343434} 0.014}                         & {\color[HTML]{343434} 0.373} & {\color[HTML]{343434} 0.213}                         & {\color[HTML]{343434} 0.094}                         \\ \hline
\rowcolor[HTML]{FFFBC8}  
{\color[HTML]{343434} ATC \cite{gargleveraging}} & {\color[HTML]{343434} 0.249} & {\color[HTML]{343434} 0.253} & {\color[HTML]{343434} 0.288} & {\color[HTML]{343434} 0.162} & {\color[HTML]{343434} 0.140} &  0.214                    & {\color[HTML]{343434} 0.369} & {\color[HTML]{343434} 0.165} \\ \hline
\rowcolor[HTML]{FFFBC8} 
{\color[HTML]{343434} IM \cite{chen2021mandoline}}                     & {\color[HTML]{343434} 0.215}                         & {\color[HTML]{343434} 0.206}                         & {\color[HTML]{343434} 0.278}                         & {\color[HTML]{343434} 0.156}                         & {\color[HTML]{343434} 0.126}                         & {\color[HTML]{343434} 0.268} & {\color[HTML]{343434} 0.131}                         & {\color[HTML]{343434} 0.163}                         \\ \hline
\rowcolor[HTML]{FFFBC8}  
{\color[HTML]{343434} DOC\cite{guillory2021predicting}}                   & {\color[HTML]{343434} 0.201}                         & {\color[HTML]{343434} 0.207}                         & {\color[HTML]{343434} 0.211}                         & {\color[HTML]{343434} 0.162}                         & {\color[HTML]{343434} 0.116}                         & {\color[HTML]{343434} 0.223} & {\color[HTML]{343434} 0.148}                         & {\color[HTML]{343434} 0.161}                         \\ \hline      
\end{tabular}}
\vspace{-4.5mm}
\label{shift-numbers}
\endgroup
\end{wraptable}

\textbf{Analysis.} 
In Fig.~\ref{fig:exp3a}, we show the model ranking of the different predictive models based on performance estimates of the different methods. Using the synthetic data from \texttt{3S}, we determine the same model ranking as the true ranking on the target---showcasing how \texttt{3S} can be used for model selection with distributional shifts. On the other hand, baselines provide incorrect rankings. 

Fig.~\ref{fig:exp3b} shows the average estimated performance of $f$ as a function of the number of observed features. We see that the \texttt{3S} estimates are closer to the oracle across the board compared to baselines. Furthermore, for an increasing number of features (i.e. increasing prior knowledge), we observe that \texttt{3S} estimates converge to the oracle. This is unsurprising: the more features we acquire target statistics of, the better we can model the true shifted distribution. Source RS does so too, but more slowly and with major variance issues. 

We also assess raw estimate errors in Table \ref{shift-numbers}. \texttt{3S} clearly has lower performance estimate errors for the numerous downstream models. Beyond having reduced error compared to rejection sampling, it is interesting that \texttt{3S} generally outperforms highly specialized methods (ATC, IM, DOC), which not only have access to \textbf{more information} but are also developed specifically to predict target domain performance. A likely rationale for this is that these methods rely on probabilities and hence do not translate well to the non-neural methods widely used in the tabular domain.

\textbf{\textcolor{ForestGreen}{Takeaway:}} High-level information about potential shifts can be translated into realistic synthetic data, to better estimate target domain model performance and select the best model to use.

\section{Discussion} \label{sec:discussion}
\vspace{-1mm}
\textbf{Synthetic data for model evaluation.} 
Accurate model evaluation is of vital importance to ML, but this is challenging when there is \emph{limited test data}. We have shown in Sec. \ref{sec:real-data-fails} that it is hard to accurately evaluate performance for small subgroups (e.g. minority race groups) and to understand how models would perform under distributional shifts using real data alone.
We have investigated the potential of synthetic data for model evaluation and found that \texttt{3S} can accurately evaluate the performance of a prediction model, even when the generative model is trained on the same test set. A deep generative ensemble approach can be used to quantify the uncertainty in \texttt{3S} estimates, which we have shown provides reliable coverage of the true model performance. Furthermore, we explored synthetic test sets with shifts, which provide practitioners with insight into how their model may perform in other populations or future scenarios.

\textbf{Model reports.} We envision evaluations using synthetic data could be published alongside models to give insight into when a model should and should not be used---e.g. to complete model evaluation templates such as Model Cards for Model Reporting \cite{gebru2021datasheets}. Appendix \ref{appx:modelreport} illustrates an example model report using \texttt{3S}.

\textbf{Practical considerations.} We discuss and explore limitations in detail in Appendix \ref{appx:limitations}. Let us highlight three practical considerations to the application of synthetic data for testing. 
Firstly, evaluating the performance under distributional shifts requires assumptions on the shift. 
These assumptions affect model evaluation and require careful consideration from the end-user. 
This is especially true for large shifts or scenarios where we do not have enough training data to describe the shifted distribution well enough. 
However, even if absolute estimates are inaccurate, we can still provide insight into trends of different scenarios. Secondly, synthetic data might have failure modes or limitations in certain settings, such as cases where there are only a handful of samples or with many samples. Thirdly, training and tuning a generative model is non-trivial. \texttt{3S}' DGE mechanism mitigates this by offering an uncertainty estimate that provides insight into the error of the generative learning process. The computational cost of such a generative ensemble may be justified by the cost of untrustworthy model evaluation.

\section*{Acknowledgements}
Boris van Bruegel is supported by ONR-UK. Nabeel Seedat is supported by the Cystic Fibrosis Trust. We would like to warmly thank the reviewers for their time and useful feedback. The authors also thank Alicia Curth, Alex Chan, Zhaozhi Qian, Daniel Jarrett, Tennison Liu and Andrew Rashbass for their comments and feedback on an earlier manuscript.


\bibliographystyle{unsrt}
\bibliography{example}

\begin{thebibliography}{10}

\bibitem{shergadwala2022human}
Murtuza~N Shergadwala, Himabindu Lakkaraju, and Krishnaram Kenthapadi.
\newblock A human-centric perspective on model monitoring.
\newblock In {\em Proceedings of the AAAI Conference on Human Computation and
  Crowdsourcing}, volume~10, pages 173--183, 2022.

\bibitem{oakden2020hidden}
Luke Oakden-Rayner, Jared Dunnmon, Gustavo Carneiro, and Christopher R{\'e}.
\newblock Hidden stratification causes clinically meaningful failures in
  machine learning for medical imaging.
\newblock In {\em Proceedings of the ACM Conference on Health, Inference, and
  Learning}, pages 151--159, 2020.

\bibitem{suresh2019framework}
Harini Suresh and John~V Guttag.
\newblock A framework for understanding unintended consequences of machine
  learning.
\newblock {\em arXiv preprint arXiv:1901.10002}, 2:8, 2019.

\bibitem{cabrera2019fairvis}
{\'A}ngel~Alexander Cabrera, Will Epperson, Fred Hohman, Minsuk Kahng, Jamie
  Morgenstern, and Duen~Horng Chau.
\newblock {FairVis}: Visual analytics for discovering intersectional bias in
  machine learning.
\newblock In {\em 2019 IEEE Conference on Visual Analytics Science and
  Technology (VAST)}, pages 46--56. IEEE, 2019.

\bibitem{cabreradiscovery}
Angel~Alexander Cabrera, Minsuk Kahng, Fred Hohman, Jamie Morgenstern, and
  Duen~Horng Chau.
\newblock Discovery of intersectional bias in machine learning using automatic
  subgroup generation.
\newblock In {\em ICLR Debugging Machine Learning Models Workshop}, 2019.

\bibitem{pianykh2020continuous}
Oleg~S Pianykh, Georg Langs, Marc Dewey, Dieter~R Enzmann, Christian~J Herold,
  Stefan~O Schoenberg, and James~A Brink.
\newblock Continuous learning {AI} in radiology: {I}mplementation principles
  and early applications.
\newblock {\em Radiology}, 297(1):6--14, 2020.

\bibitem{quinonero2008dataset}
Joaquin Quinonero-Candela, Masashi Sugiyama, Anton Schwaighofer, and Neil~D
  Lawrence.
\newblock {\em Dataset shift in machine learning}.
\newblock MIT Press, 2008.

\bibitem{koh2021wilds}
Pang~Wei Koh, Shiori Sagawa, Henrik Marklund, Sang~Michael Xie, Marvin Zhang,
  Akshay Balsubramani, Weihua Hu, Michihiro Yasunaga, Richard~Lanas Phillips,
  Irena Gao, Tony Lee, Etienne David, Ian Stavness, Wei Guo, Berton~A.
  Earnshaw, Imran~S. Haque, Sara Beery, Jure Leskovec, Anshul Kundaje, Emma
  Pierson, Sergey Levine, Chelsea Finn, and Percy Liang.
\newblock {WILDS}: A benchmark of in-the-wild distribution shifts.
\newblock In {\em International Conference on Machine Learning}, pages
  5637--5664. PMLR, 2021.

\bibitem{seedat2022dc}
Nabeel Seedat, Fergus Imrie, and Mihaela van~der Schaar.
\newblock Dc-check: A data-centric ai checklist to guide the development of
  reliable machine learning systems.
\newblock {\em arXiv preprint arXiv:2211.05764}, 2022.

\bibitem{gebru2021datasheets}
Timnit Gebru, Jamie Morgenstern, Briana Vecchione, Jennifer~Wortman Vaughan,
  Hanna Wallach, Hal~Daum{\'e} Iii, and Kate Crawford.
\newblock Datasheets for datasets.
\newblock {\em Communications of the ACM}, 64(12):86--92, 2021.

\bibitem{suresh2018learning}
Harini Suresh, Jen~J Gong, and John~V Guttag.
\newblock Learning tasks for multitask learning: Heterogenous patient
  populations in the {ICU}.
\newblock In {\em Proceedings of the 24th ACM SIGKDD International Conference
  on Knowledge Discovery \& Data Mining}, pages 802--810, 2018.

\bibitem{goel2020model}
Karan Goel, Albert Gu, Yixuan Li, and Christopher Re.
\newblock Model patching: Closing the subgroup performance gap with data
  augmentation.
\newblock In {\em International Conference on Learning Representations}, 2020.

\bibitem{saria2019tutorial}
Suchi Saria and Adarsh Subbaswamy.
\newblock Tutorial: safe and reliable machine learning.
\newblock {\em arXiv preprint arXiv:1904.07204}, 2019.

\bibitem{d2020underspecification}
Alexander D'Amour, Katherine Heller, Dan Moldovan, Ben Adlam, Babak Alipanahi,
  Alex Beutel, Christina Chen, Jonathan Deaton, Jacob Eisenstein, Matthew~D
  Hoffman, et~al.
\newblock Underspecification presents challenges for credibility in modern
  machine learning.
\newblock {\em arXiv preprint arXiv:2011.03395}, 2020.

\bibitem{cohen2021problems}
Joseph~Paul Cohen, Tianshi Cao, Joseph~D Viviano, Chin-Wei Huang, Michael
  Fralick, Marzyeh Ghassemi, Muhammad Mamdani, Russell Greiner, and Yoshua
  Bengio.
\newblock Problems in the deployment of machine-learned models in health care.
\newblock {\em CMAJ}, 193(35):E1391--E1394, 2021.

\bibitem{deo2015machine}
Rahul~C Deo.
\newblock Machine learning in medicine.
\newblock {\em Circulation}, 132(20):1920--1930, 2015.

\bibitem{savage2012better}
Neil Savage.
\newblock Better medicine through machine learning.
\newblock {\em Communications of the ACM}, 55(1):17--19, 2012.

\bibitem{nezhad2017subic}
Milad~Zafar Nezhad, Dongxiao Zhu, Najibesadat Sadati, Kai Yang, and Phillip
  Levi.
\newblock {SUBIC}: A supervised bi-clustering approach for precision medicine.
\newblock In {\em 2017 16th IEEE International Conference on Machine Learning
  and Applications (ICMLA)}, pages 755--760. IEEE, 2017.

\bibitem{patel2008investigating}
Kayur Patel, James Fogarty, James~A Landay, and Beverly Harrison.
\newblock Investigating statistical machine learning as a tool for software
  development.
\newblock In {\em Proceedings of the SIGCHI Conference on Human Factors in
  Computing Systems}, pages 667--676, 2008.

\bibitem{recht2019imagenet}
Benjamin Recht, Rebecca Roelofs, Ludwig Schmidt, and Vaishaal Shankar.
\newblock Do {ImageNet} classifiers generalize to {ImageNet}?
\newblock In {\em International Conference on Machine Learning}, pages
  5389--5400. PMLR, 2019.

\bibitem{miller2021model}
Andrew~C Miller, Leon~A Gatys, Joseph Futoma, and Emily Fox.
\newblock Model-based metrics: Sample-efficient estimates of predictive model
  subpopulation performance.
\newblock In {\em Machine Learning for Healthcare Conference}, pages 308--336.
  PMLR, 2021.

\bibitem{borisov2021deep}
Vadim Borisov, Tobias Leemann, Kathrin Se{\ss}ler, Johannes Haug, Martin
  Pawelczyk, and Gjergji Kasneci.
\newblock Deep neural networks and tabular data: A survey.
\newblock {\em arXiv preprint arXiv:2110.01889}, 2021.

\bibitem{shwartz2022tabular}
Ravid Shwartz-Ziv and Amitai Armon.
\newblock Tabular data: Deep learning is not all you need.
\newblock {\em Information Fusion}, 81:84--90, 2022.

\bibitem{kaggle_2017}
Kaggle.
\newblock Kaggle machine learning and data science survey, 2017.

\bibitem{breugel2023dge}
Boris van Breugel, Zhaozhi Qian, and Mihaela van~der Schaar.
\newblock Synthetic data, real errors: how (not) to publish and use synthetic
  data.
\newblock In {\em International Conference on Machine Learning}. PMLR, 2023.

\bibitem{flach2019performance}
Peter Flach.
\newblock Performance evaluation in machine learning: the good, the bad, the
  ugly, and the way forward.
\newblock In {\em Proceedings of the AAAI Conference on Artificial
  Intelligence}, volume~33, pages 9808--9814, 2019.

\bibitem{wiles2021fine}
Olivia Wiles, Sven Gowal, Florian Stimberg, Sylvestre-Alvise Rebuffi, Ira
  Ktena, Krishnamurthy~Dj Dvijotham, and Ali~Taylan Cemgil.
\newblock A fine-grained analysis on distribution shift.
\newblock In {\em International Conference on Learning Representations}, 2021.

\bibitem{hendrycks2018benchmarking}
Dan Hendrycks and Thomas Dietterich.
\newblock Benchmarking neural network robustness to common corruptions and
  perturbations.
\newblock In {\em International Conference on Learning Representations}, 2018.

\bibitem{ribeiro2020beyond}
Marco~Tulio Ribeiro, Tongshuang Wu, Carlos Guestrin, and Sameer Singh.
\newblock Beyond accuracy: Behavioral testing of nlp models with checklist.
\newblock In {\em Proceedings of the 58th Annual Meeting of the Association for
  Computational Linguistics}, pages 4902--4912, 2020.

\bibitem{rottger2021hatecheck}
Paul R{\"o}ttger, Bertie Vidgen, Dong Nguyen, Zeerak Waseem, Helen Margetts,
  Janet Pierrehumbert, et~al.
\newblock Hatecheck: Functional tests for hate speech detection models.
\newblock In {\em Proceedings of the 59th Annual Meeting of the Association for
  Computational Linguistics and the 11th International Joint Conference on
  Natural Language Processing (Volume 1: Long Papers)}, page~41. Association
  for Computational Linguistics, 2021.

\bibitem{gargleveraging}
Saurabh Garg, Sivaraman Balakrishnan, Zachary~Chase Lipton, Behnam Neyshabur,
  and Hanie Sedghi.
\newblock Leveraging unlabeled data to predict out-of-distribution performance.
\newblock In {\em International Conference on Learning Representations}, 2022.

\bibitem{guillory2021predicting}
Devin Guillory, Vaishaal Shankar, Sayna Ebrahimi, Trevor Darrell, and Ludwig
  Schmidt.
\newblock Predicting with confidence on unseen distributions.
\newblock In {\em Proceedings of the IEEE/CVF International Conference on
  Computer Vision}, pages 1134--1144, 2021.

\bibitem{chen2021mandoline}
Mayee Chen, Karan Goel, Nimit~S Sohoni, Fait Poms, Kayvon Fatahalian, and
  Christopher R{\'e}.
\newblock Mandoline: Model evaluation under distribution shift.
\newblock In {\em International Conference on Machine Learning}, pages
  1617--1629. PMLR, 2021.

\bibitem{breugel2023beyond}
Boris van Breugel and Mihaela van~der Schaar.
\newblock Beyond privacy: Navigating the opportunities and challenges of
  synthetic data.
\newblock {\em arXiv preprint arXiv:2304.03722}, 2023.

\bibitem{jordon2018pate}
James Jordon, Jinsung Yoon, and Mihaela Van Der~Schaar.
\newblock {PATE-GAN}: Generating synthetic data with differential privacy
  guarantees.
\newblock In {\em International Conference on Learning Representations}, 2018.

\bibitem{assefa2020generating}
Samuel~A Assefa, Danial Dervovic, Mahmoud Mahfouz, Robert~E Tillman, Prashant
  Reddy, and Manuela Veloso.
\newblock Generating synthetic data in finance: opportunities, challenges and
  pitfalls.
\newblock In {\em Proceedings of the First ACM International Conference on AI
  in Finance}, pages 1--8, 2020.

\bibitem{xu2019achieving}
Depeng Xu, Yongkai Wu, Shuhan Yuan, Lu~Zhang, and Xintao Wu.
\newblock Achieving causal fairness through generative adversarial networks.
\newblock In {\em Proceedings of the Twenty-Eighth International Joint
  Conference on Artificial Intelligence}, 2019.

\bibitem{breugel2021decaf}
Boris van Breugel, Trent Kyono, Jeroen Berrevoets, and Mihaela van~der Schaar.
\newblock {DECAF}: Generating fair synthetic data using causally-aware
  generative networks.
\newblock {\em Advances in Neural Information Processing Systems},
  34:22221--22233, 2021.

\bibitem{antoniou2017data}
Antreas Antoniou, Amos Storkey, and Harrison Edwards.
\newblock Data augmentation generative adversarial networks.
\newblock {\em arXiv preprint arXiv:1711.04340}, 2017.

\bibitem{dina2022effect}
Ayesha~S Dina, AB~Siddique, and D~Manivannan.
\newblock Effect of balancing data using synthetic data on the performance of
  machine learning classifiers for intrusion detection in computer networks.
\newblock {\em arXiv preprint arXiv:2204.00144}, 2022.

\bibitem{das2022conditional}
Hari~Prasanna Das, Ryan Tran, Japjot Singh, Xiangyu Yue, Geoffrey Tison,
  Alberto Sangiovanni-Vincentelli, and Costas~J Spanos.
\newblock Conditional synthetic data generation for robust machine learning
  applications with limited pandemic data.
\newblock In {\em Proceedings of the AAAI Conference on Artificial
  Intelligence}, volume~36, pages 11792--11800, 2022.

\bibitem{bing2022conditional}
Simon Bing, Andrea Dittadi, Stefan Bauer, and Patrick Schwab.
\newblock Conditional generation of medical time series for extrapolation to
  underrepresented populations.
\newblock {\em PLOS Digital Health}, 1(7):1--26, 07 2022.

\bibitem{wang2019learning}
Qi~Wang, Junyu Gao, Wei Lin, and Yuan Yuan.
\newblock Learning from synthetic data for crowd counting in the wild.
\newblock In {\em Proceedings of the IEEE/CVF Conference on Computer Vision and
  Pattern Recognition}, pages 8198--8207, 2019.

\bibitem{trigueros2021generating}
Daniel~S{\'a}ez Trigueros, Li~Meng, and Margaret Hartnett.
\newblock Generating photo-realistic training data to improve face recognition
  accuracy.
\newblock {\em Neural Networks}, 134:86--94, 2021.

\bibitem{ruiz2022simulated}
Nataniel Ruiz, Adam Kortylewski, Weichao Qiu, Cihang Xie, Sarah~Adel Bargal,
  Alan Yuille, and Stan Sclaroff.
\newblock Simulated adversarial testing of face recognition models.
\newblock In {\em Proceedings of the IEEE/CVF Conference on Computer Vision and
  Pattern Recognition}, pages 4145--4155, 2022.

\bibitem{khan2019procsy}
Samin Khan, Buu Phan, Rick Salay, and Krzysztof Czarnecki.
\newblock Procsy: Procedural synthetic dataset generation towards influence
  factor studies of semantic segmentation networks.
\newblock In {\em CVPR workshops}, pages 88--96, 2019.

\bibitem{xu2019modeling}
Lei Xu, Maria Skoularidou, Alfredo Cuesta-Infante, and Kalyan Veeramachaneni.
\newblock Modeling tabular data using conditional {GAN}.
\newblock {\em Advances in Neural Information Processing Systems}, 32, 2019.

\bibitem{lakshminarayanan2017}
Balaji Lakshminarayanan, Alexander Pritzel, and Charles Blundell.
\newblock Simple and scalable predictive uncertainty estimation using deep
  ensembles.
\newblock {\em Advances in Neural Information Processing Systems}, 30, 2017.

\bibitem{crenshaw1989demarginalizing}
Kimberley Crenshaw.
\newblock Demarginalizing the intersection of race and sex: A black feminist
  critique of antidiscrimination doctrine, feminist theory, and antiracist
  politics. the university of chicago legal forum, 1989 (1), 139-167.
\newblock {\em Chicago, IL}, 1989.

\bibitem{amodei2016concrete}
Dario Amodei, Chris Olah, Jacob Steinhardt, Paul Christiano, John Schulman, and
  Dan Man{\'e}.
\newblock Concrete problems in {AI} safety.
\newblock {\em arXiv preprint arXiv:1606.06565}, 2016.

\bibitem{wu2021online}
Ruihan Wu, Chuan Guo, Yi~Su, and Kilian~Q Weinberger.
\newblock Online adaptation to label distribution shift.
\newblock {\em Advances in Neural Information Processing Systems},
  34:11340--11351, 2021.

\bibitem{ji2023large}
Christina~X Ji, Ahmed~M Alaa, and David Sontag.
\newblock Large-scale study of temporal shift in health insurance claims.
\newblock {\em Conference on Health, Inference, and Learning (CHIL)}, 2023.

\bibitem{shimodaira2000improving}
Hidetoshi Shimodaira.
\newblock Improving predictive inference under covariate shift by weighting the
  log-likelihood function.
\newblock {\em Journal of Statistical Planning and Inference}, 90(2):227--244,
  2000.

\bibitem{moreno2012unifying}
Jose~G Moreno-Torres, Troy Raeder, Roc{\'\i}o Alaiz-Rodr{\'\i}guez, Nitesh~V
  Chawla, and Francisco Herrera.
\newblock A unifying view on dataset shift in classification.
\newblock {\em Pattern Recognition}, 45(1):521--530, 2012.

\bibitem{saerens2002adjusting}
Marco Saerens, Patrice Latinne, and Christine Decaestecker.
\newblock Adjusting the outputs of a classifier to new a priori probabilities:
  a simple procedure.
\newblock {\em Neural Computation}, 14(1):21--41, 2002.

\bibitem{gretton2012kernel}
Arthur Gretton, Karsten~M Borgwardt, Malte~J Rasch, Bernhard Sch{\"o}lkopf, and
  Alexander Smola.
\newblock A kernel two-sample test.
\newblock {\em The Journal of Machine Learning Research}, 13(1):723--773, 2012.

\bibitem{uci}
Arthur Asuncion and David Newman.
\newblock {UCI} machine learning repository, 2007.

\bibitem{baqui2020ethnic}
Pedro Baqui, Ioana Bica, Valerio Marra, Ari Ercole, and Mihaela van Der~Schaar.
\newblock Ethnic and regional variations in hospital mortality from {COVID}-19
  in {B}razil: a cross-sectional observational study.
\newblock {\em The Lancet Global Health}, 8(8):e1018--e1026, 2020.

\bibitem{knaus1995support}
William~A Knaus, Frank~E Harrell, Joanne Lynn, Lee Goldman, Russell~S Phillips,
  Alfred~F Connors, Neal~V Dawson, William~J Fulkerson, Robert~M Califf, Norman
  Desbiens, Peter Layde, Robert~K Oye, Paul~E Bellamy, Rosemarie~B Hakim, and
  Douglas~P Wagner.
\newblock The {SUPPORT} prognostic model: Objective estimates of survival for
  seriously ill hospitalized adults.
\newblock {\em Annals of Internal Medicine}, 122(3):191--203, 1995.

\bibitem{jesus2022turning}
S{\'e}rgio Jesus, Jos{\'e} Pombal, Duarte Alves, Andr{\'e} Cruz, Pedro Saleiro,
  Rita Ribeiro, Jo{\~a}o Gama, and Pedro Bizarro.
\newblock Turning the tables: Biased, imbalanced, dynamic tabular datasets for
  {ML} evaluation.
\newblock {\em Advances in Neural Information Processing Systems},
  35:33563--33575, 2022.

\bibitem{fehrman2017five}
Elaine Fehrman, Awaz~K Muhammad, Evgeny~M Mirkes, Vincent Egan, and Alexander~N
  Gorban.
\newblock The five factor model of personality and evaluation of drug
  consumption risk.
\newblock In {\em Data Science: Innovative Developments in Data Analysis and
  Clustering}, pages 231--242. Springer, 2017.

\bibitem{efron1992bootstrap}
Bradley Efron.
\newblock {\em Bootstrap methods: another look at the jackknife}.
\newblock Springer, 1992.

\bibitem{kompa2021empirical}
Benjamin Kompa, Jasper Snoek, and Andrew~L Beam.
\newblock Empirical frequentist coverage of deep learning uncertainty
  quantification procedures.
\newblock {\em Entropy}, 23(12):1608, 2021.

\bibitem{seedat2022data}
Nabeel Seedat, Jonathan Crabb{\'e}, and Mihaela van~der Schaar.
\newblock {Data-SUITE}: Data-centric identification of in-distribution
  incongruous examples.
\newblock In {\em International Conference on Machine Learning}, pages
  19467--19496. PMLR, 2022.

\bibitem{navratil2020uncertainty}
Jiri Navratil, Matthew Arnold, and Benjamin Elder.
\newblock Uncertainty prediction for deep sequential regression using meta
  models.
\newblock {\em arXiv preprint arXiv:2007.01350}, 2020.

\bibitem{seer}
M{\'a}ire~A Duggan, William~F Anderson, Sean Altekruse, Lynne Penberthy, and
  Mark~E Sherman.
\newblock The surveillance, epidemiology and end results ({SEER}) program and
  pathology: towards strengthening the critical relationship.
\newblock {\em The American Journal of Surgical Pathology}, 40(12):e94, 2016.

\bibitem{cutract}
Prostate~Cancer UK.
\newblock Cutract.
\newblock {\em https://prostatecanceruk.org}, 2019.

\bibitem{ribeiro2022adaptive}
Marco~Tulio Ribeiro and Scott Lundberg.
\newblock Adaptive testing and debugging of nlp models.
\newblock In {\em Proceedings of the 60th Annual Meeting of the Association for
  Computational Linguistics (Volume 1: Long Papers)}, pages 3253--3267, 2022.

\bibitem{sambasivan2021everyone}
Nithya Sambasivan, Shivani Kapania, Hannah Highfill, Diana Akrong, Praveen
  Paritosh, and Lora~M Aroyo.
\newblock “{E}veryone wants to do the model work, not the data work”: Data
  cascades in high-stakes ai.
\newblock In {\em Proceedings of the 2021 CHI Conference on Human Factors in
  Computing Systems}, pages 1--15, 2021.

\bibitem{seedat2022diq}
Nabeel Seedat, Jonathan Crabb{\'e}, Ioana Bica, and Mihaela van~der Schaar.
\newblock Data-iq: Characterizing subgroups with heterogeneous outcomes in
  tabular data.
\newblock {\em Advances in Neural Information Processing Systems},
  35:23660--23674, 2022.

\bibitem{ng2021}
Andrew Ng, Lora Aroyo, Cody Coleman, Greg Diamos, Vijay~Janapa Reddi, Joaquin
  Vanschoren, Carole-Jean Wu, and Sharon Zhou.
\newblock {NeurIPS} data-centric {AI} workshop, 2021.

\bibitem{polyzotis2021can}
Neoklis Polyzotis and Matei Zaharia.
\newblock What can data-centric {AI} learn from data and {ML} engineering?
\newblock {\em arXiv preprint arXiv:2112.06439}, 2021.

\bibitem{chan2021medkit}
Alex Chan, Ioana Bica, Alihan H{\"u}y{\"u}k, Daniel Jarrett, and Mihaela
  van~der Schaar.
\newblock The {Medkit-Learn}(ing) environment: Medical decision modelling
  through simulation.
\newblock In {\em Thirty-fifth Conference on Neural Information Processing
  Systems Datasets and Benchmarks Track (Round 2)}, 2021.

\bibitem{neal2020realcause}
Brady Neal, Chin-Wei Huang, and Sunand Raghupathi.
\newblock {RealCause}: Realistic causal inference benchmarking.
\newblock {\em arXiv preprint arXiv:2011.15007}, 2020.

\bibitem{ATHEY2021}
Susan Athey, Guido~W. Imbens, Jonas Metzger, and Evan Munro.
\newblock Using {W}asserstein generative adversarial networks for the design of
  {M}onte {C}arlo simulations.
\newblock {\em Journal of Econometrics}, 2021.

\bibitem{parikh2022validating}
Harsh Parikh, Carlos Varjao, Louise Xu, and Eric~Tchetgen Tchetgen.
\newblock Validating causal inference methods.
\newblock In {\em International Conference on Machine Learning}, pages
  17346--17358. PMLR, 2022.

\bibitem{kortylewski2019analyzing}
Adam Kortylewski, Bernhard Egger, Andreas Schneider, Thomas Gerig, Andreas
  Morel-Forster, and Thomas Vetter.
\newblock Analyzing and reducing the damage of dataset bias to face recognition
  with synthetic data.
\newblock In {\em Proceedings of the IEEE/CVF Conference on Computer Vision and
  Pattern Recognition Workshops}, 2019.

\bibitem{li2021discover}
Zhiheng Li and Chenliang Xu.
\newblock Discover the unknown biased attribute of an image classifier.
\newblock In {\em Proceedings of the IEEE/CVF International Conference on
  Computer Vision}, pages 14970--14979, 2021.

\bibitem{mcduff2019characterizing}
Daniel McDuff, Shuang Ma, Yale Song, and Ashish Kapoor.
\newblock Characterizing bias in classifiers using generative models.
\newblock {\em Advances in Neural Information Processing Systems}, 32, 2019.

\bibitem{howe2017synthetic}
Bill Howe, Julia Stoyanovich, Haoyue Ping, Bernease Herman, and Matt Gee.
\newblock Synthetic data for social good.
\newblock {\em arXiv preprint arXiv:1710.08874}, 2017.

\bibitem{saha2022data}
Diptikalyan Saha, Aniya Aggarwal, and Sandeep Hans.
\newblock Data synthesis for testing black-box machine learning models.
\newblock In {\em 5th Joint International Conference on Data Science \&
  Management of Data (9th ACM IKDD CODS and 27th COMAD)}, pages 110--114, 2022.

\bibitem{sivep}
SIVEP Brazil Ministry~of Health.
\newblock {M}inistry of {H}ealth {SIVEP-Gripe} public dataset.

\bibitem{efron1983estimating}
Bradley Efron.
\newblock Estimating the error rate of a prediction rule: improvement on
  cross-validation.
\newblock {\em Journal of the American Statistical Association},
  78(382):316--331, 1983.

\bibitem{diciccio1996bootstrap}
Thomas~J DiCiccio and Bradley Efron.
\newblock Bootstrap confidence intervals.
\newblock {\em Statistical science}, 11(3):189--228, 1996.

\bibitem{carpenter2000bootstrap}
James Carpenter and John Bithell.
\newblock Bootstrap confidence intervals: when, which, what? {A} practical
  guide for medical statisticians.
\newblock {\em Statistics in medicine}, 19(9):1141--1164, 2000.

\bibitem{borji2019pros}
Ali Borji.
\newblock Pros and cons of {GAN} evaluation measures.
\newblock {\em Computer Vision and Image Understanding}, 179:41--65, 2019.

\bibitem{sutherland2017generative}
Dougal~J Sutherland, Hsiao-Yu Tung, Heiko Strathmann, Soumyajit De, Aaditya
  Ramdas, Alexander~J Smola, and Arthur Gretton.
\newblock Generative models and model criticism via optimized maximum mean
  discrepancy.
\newblock In {\em International Conference on Learning Representations}, 2017.

\bibitem{bounliphone2016test}
Wacha Bounliphone, Eugene Belilovsky, Matthew~B Blaschko, Ioannis Antonoglou,
  and Arthur Gretton.
\newblock A test of relative similarity for model selection in generative
  models.
\newblock In {\em International Conference on Learning Representations}, 2016.

\bibitem{varshney2021trustworthy}
Kush~R Varshney.
\newblock Trustworthy machine learning.
\newblock {\em Chappaqua, NY}, 2021.
\newblock Section 9.2.1.

\bibitem{alaa2022faithful}
Ahmed Alaa, Boris Van~Breugel, Evgeny~S Saveliev, and Mihaela van~der Schaar.
\newblock How faithful is your synthetic data? sample-level metrics for
  evaluating and auditing generative models.
\newblock In {\em International Conference on Machine Learning}, pages
  290--306. PMLR, 2022.

\end{thebibliography}

\clearpage
\appendix
\addcontentsline{toc}{section}{Appendix}
\part{Appendix: Can You Rely on Your Model? A Case for Synthetic-Data-Based Model Evaluation}
\mtcsetdepth{parttoc}{5} 
\parttoc

\newpage
\section{Extended Related Work} \label{appx:related_work}
We present a comparison of our framework, \threeS, and provide further contrast to related work.
Table \ref{related_work}, highlights that related methods often do not permit automated test creation, in particular, they are often human labor-intensive ---see Table \ref{related_work}. These benchmark tasks are tailored to specific datasets and tasks and hence cannot be customized and/or personalized to an end-user's specific task, dataset, or trained model. Finally, both benchmarking methods and behavioral testing require additional data to be collected or created. This contrasts \threeS which only requires the original test data, already available.

\begin{table*}[hbt]
\centering
\caption{Comparison of \threeS to human-crafted testing frameworks. (i) Automated test creation, (ii) Tests can be personalised to user dataset and use case(e.g. distributional shift), and (iii) No additional data or info required.}
\rotatebox{0}{
\scalebox{0.7}{
\begin{tabular}{@{}l|l|c|c|ccccc@{}}
\toprule
        \textbf{Method}
      & \textbf{Approach}
      & \textbf{Assumption}
      & \textbf{Use cases}  
      & \textbf{(i)}
      & \textbf{(ii)}
      & \textbf{(iii)}
                        \\ \midrule
\threeS Testing (Ours) & Synthetic test data & \makecell[l]{Generative model fits the data}   & \makecell[l]{ (P1) Subgroup testing  \\ (P2) Distributional shift testing}  &  \cmark & \cmark  & \cmark  \\ \hline\hline
\multicolumn{7}{c}{\textbf{BENCHMARK TASKS} }   \\\hline\hline
Imagenet-C/P \citep{hendrycks2018benchmarking} & Create corrupted images &  \makecell[l]{Synthetic corruption \\ reflects the real-world}& \makecell[l]{Image corruption testing} &   \xmark & \xmark  & \xmark \\
Wilds \citep{koh2021wilds}  & Collect data with real shifts &  \makecell[l]{Collected `wild' data reflects \\sufficient use cases}  & \makecell[l]{Distribution shift testing} &   \xmark & \xmark  & \xmark  \\
\hline\hline
\multicolumn{7}{c}{\textbf{MODEL BEHAVIOURAL TESTING}}   \\\hline\hline
CheckList \citep{ribeiro2020beyond} & Human crafted test scenarios & \makecell[l]{Know a-priori scenarios to test} & \makecell[l]{Crafted scenario tests} &  \xmark & \cmark & \xmark    \\
HateCheck \citep{rottger2021hatecheck} & Human crafted test scenarios  & \makecell[l]{Know a-priori scenarios to test}   & \makecell[l]{Crafted scenario tests} & \xmark & \cmark & \xmark  \\
AdaTest \citep{ribeiro2022adaptive} & GPT-3 creates tests, human refines  & Human-in-the-loop  & \makecell[l]{Weakness probing} &  \cmark & \cmark & \xmark    \\

\bottomrule
\end{tabular}}}
\label{related_work}
\vspace{-5pt}
\end{table*}

\textbf{Data-centric AI.} The usage and assessment is vital for ML models, yet is often overlooked as operational \citep{sambasivan2021everyone, seedat2022diq}. The recent focus on data-centric AI \citep{ng2021, seedat2022dc} aims to build systematic tools to improve the quality of data used to train and evaluate ML models \citep{polyzotis2021can, seedat2022data}. \threeS contributes to this nascent body of work, specifically around the usage and generation of data to better evaluate and test ML models

\textbf{Simulators and counterfactual generators.} Simulators have been used to benchmark algorithms in settings such as causal effect estimation and sequential decision-making \citep{chan2021medkit}. The work on simulators is not directly related, as the goal is often to test scenarios that are not available in real-world data. For example, in \citep{chan2021medkit}, the goal is customization of the decision-making policy in order to evaluate methods to better understand human decision-making. Similarly, there has been work on generating realistic synthetic data in the causal inference domain \citep{neal2020realcause,ATHEY2021,parikh2022validating}, but these methods focus on benchmarking causal inference methods, do not consider distributional shifts nor subgroup evaluation, and do not explore the value of the synthetic data beyond realistic, ground-truth counterfactuals.

\textbf{Synthetic data in computer vision.}
Synthetic data has been used in computer vision both to improve model training and to test weaknesses in models. These methods can be grouped as follows by their motivations:

\begin{itemize}
    \item Generate synthetic data for training, to reduce the reliance on collecting and annotating large training sets— This is different from \threeS as they focus on constructing better training sets, rather than constructing better test sets for model evaluation
    \item Generate synthetic data to improve the model by augmenting the real dataset with synthetic examples—*Again, this is different from \threeS as they focus on training better-performing models, rather than the evaluation of an already trained model
    \item Generate synthetic data to probe models on different dataset attributes. For example, in face recognition, how the model might perform on faces with long vs short hair. This is most similar to \threeS, but there are clear differences in both the goal for and approach to generating the data. We compare \threeS to (i)  CGI- or physics-based simulators and (ii) deep generative models for probing in computer vision.
\end{itemize}

In Table \ref{cv-related_work}, we contrast both simulators and generative approaches where synthetic data to probe models on different dataset attributes.

\begingroup
\setlength{\tabcolsep}{4pt}
\renewcommand{\arraystretch}{1.1}
\begin{table*}[hbt]
\centering
\caption{Comparing \threeS to computer vision synthetic data approaches. $A$ is a performance metric (Eq.1) and $f$ the trained predictive model. $\cD_{syn}(x)$ denotes synthetic data is created dependent on $x$}
\rotatebox{0}{
\scalebox{0.5}{
\begin{tabular}{l|l|c|c|c|c}
\toprule
        Examples
      & \makecell[c]{Data and/or generator input}
      & Conditioning info
      & \makecell{Does not require \\ pre-trained simulator/generator}
      & \makecell{$\cD_{syn}$ used for \\ training or testing}
      & Goal
                        \\ \midrule
\threeS - Subgroup  &  \multirow{2}{*}{$\cD_{train,G} = \cD_{test,f}$ (Any dataset)}   & Subgroups $\mathcal{S}$   & \multirow{2}{*}{$\checkmark$} &  \multirow{2}{*}{Testing} & \makecell[l]{Reliable subgroup performance estimates for $f$, i.e. choose \\ $\cD_{syn}\sim p_G$ s.t. $A(f;\cD_{syn}, \mathcal{S})] \approx \mathbb{E}_{\cD\overset{iid}{\sim} p_R}[A(f;\cD, \mathcal{S})]$ }   \\ \cline{1-1} \cline{3-3} \cline{6-6} 
\threeS - Shift  & & Shift information $T$  &  & & \makecell[l]{Estimate performance of $f$ under shift $T$, i.e. choose \\ $\cD_{syn}\sim p_G$ s.t. $A(f;\cD_{syn}, \mathcal{S})] \approx \mathbb{E}_{\cD\overset{iid}{\sim} p^s}[A(f;\cD, \mathcal{S})]$}     \\
\hline\hline
\multicolumn{6}{c}{\textbf{Computer Vision} }   \\\hline\hline
\cite{wang2019learning} & \makecell[l]{Video game engine (GTA5) \\ Real-world data} & Scene info $\mathcal{S}$ in virtual world  & $\times$ & Training & \makecell[l]{Improve crowd counting performance on diff. scenes by \\ generating semi-synthetic data for training $f$, \\i.e. $\text{max}_f A(f;\cD_{test} (\mathcal{S}))$} \\\hline 
\cite{trigueros2021generating} & $\cD_{train,f}$ = VGGFace & Identity attributes   & $\checkmark$ & Training & \makecell[l]{Improve overall performance of facial recognition  \\ 
i.e. $\text{max}(A(f;\cD_{test}))$} \\ \hline
\cite{kortylewski2019analyzing} & 3D face model & Nuisance transforms $\mathcal{N}$   & $\times$ & Testing & \makecell[l]{Report face recognition robustness to different \\ nuisances $\mathcal{N}$, $\mathcal{D}_{syn}(N)$ \\and report $A(f; \cD_{syn}(N)), \forall N\in \mathcal{N}$}\\\hline 
\cite{ruiz2022simulated} & 3D face model & Simulator parameters $\rho$   & $\times$ & Testing & \makecell[l]{Find adversarial failures for face recognition, i.e. \\ find  $\rho=\argmin_\rho A(f; D_{syn}(\rho))$} \\ \hline 
\cite{khan2019procsy} & \makecell[l]{CityEngine, Unreal Engine, \\CARLA} &  Weather conditions $\mathcal{S}$    & $\times$ & Testing & \makecell[l]{Report segmentation performance for self-driving \\ cars under different weather conditions, \\
i.e. $\mathcal{D}_{syn}(S)$ and report $A(f; \cD_{syn}(S)), \forall S\in \mathcal{S}$} \\ \hline 
\cite{li2021discover} & Pretrained StyleGAN & \makecell{Implicit attributes $S$ \\ (e.g. age, lighting)}  & $\times$ & Testing &  \makecell[l]{Find attributes with poor performance, \\ i.e. $\text{argmin}_S(A(f; \cD_{syn}(S)))$} \\\hline 
\cite{mcduff2019characterizing} & $\cD$ = MS-CELEB-1M  & Subgroups $S$   & $\checkmark$ & Testing & \makecell[l]{Find $S$ with poor face recognition performance, \\
i.e. $\text{argmin}_S(A(f; \cD_{syn}(S))$}\\ 
\bottomrule
\end{tabular}}}
\label{cv-related_work}
\end{table*}
\endgroup

\clearpage
\textbf{Synthetic data and tabular approaches.}

We contrast \threeS  to two works DataSynthesizer \citep{howe2017synthetic} and AITEST \citep{saha2022data}, which while seemingly similar have specific differences to \threeS. A side-by-side contrast is presented in Table \ref{tabular-related_work}.

\textbf{Data Synthesizer}

We believe \threeS is significantly different from DataSynthesizer, in terms of aims, assumptions, and algorithmically. 

\textbf{\textit{Aim and assumptions.}}
Data Synthesizer primarily focuses on privacy-preserving generation of tabular synthetic data. 
The closest component to our work is the extension the paper proposes around adversarial fake data generation. While there are no experiments, the adversarial fake data consists of three areas. We contrast them to \threeS. 

The major difference is Data Synthesizer assumes access to full knowledge about the shift/distributional change. In contrast, \threeS operates in a different setting - (1) No prior knowledge on the shift and (2) high-level partial knowledge about the shift through observing some variables in the target domain.

\begin{enumerate}
    \item Edit the distribution: this assumes the user knows exactly the shift [Full knowledge of the shift].  \threeS covers two different settings: (1) No prior knowledge on the shift, where only minimal assumptions on means of variables allow us to create characteristic curves like in Section 5.2. and (2) Incorporating prior knowledge, in which some features are observed from the shifted distribution and we use these to generate the full data from the shifted distribution, like in Section 5.2.2. Consequently, the difference is that \threeS tackles the no and partial information settings, whereas Data Synthesizer tackles the full info setting of editing the distribution. 
\item  Preconfigured pathological distributions — this requires full and exact knowledge about the shift, which differs from \threeS of partial knowledge and no prior knowledge settings.
\item  Injecting missing data, extreme values — either such an approach is possible to incorporate in \threeS. We see these ideas as complementary.
\end{enumerate}

\textbf{\textit{Algorithmic.}}

The authors propose three methods, one with random features, one with independent features, and one with correlated features. Due to the absence of correlation in the first two, these reduce the data utility. Let us thus focus on the third method, that does include correlation. This approach uses Bayesian Networks and is only applicable to discrete data, hence needing to discretize continuous variables. This loses utility when a coarse discretization is chosen, while a fine discretization is often intractable and data-inefficient due to the ordinal information being lost, e.g. results for $age=31$ and $age=32$ will generally be similar—exactly the reason why the independent approach was also introduced. Bayesian Nets are also limited in other ways, e.g. results can be influenced by the feature generation order deviating from the real data generation process' ordering, as indicated by the authors of DataSynthesizer in Figures 5 and 6. 

\textbf{AITEST}

 We contrast \threeS to AITEST in terms of aims and assumptions, algorithm, and use cases. 

\textbf{\textit{Aims and assumptions.}}
AITEST has a significantly different aim and method compared to \threeS. As mentioned by the reviewer, AITEST can test for adversarial robustness by generating realistic data with user-defined constraints, but this is different from our work that aims to generate synthetic test data for granular evaluation and distributional shifts. 

Additionally, the assumptions on user input are quite different: AITEST enables users to define constraints on features and associations between features, whereas \threeS requires information in terms of which subgroups to test or shifts to generate.  We do see possibilities to combine both frameworks, e.g. through including constraints similar to the ones AITEST uses within the \threeS method, or using fairness as a downstream task. 

We have taken a step in this direction and added fairness as an additional experiment and have included this experiment in the new Appendix D.4. 

\textbf{\textit{Algorithmic}}
AITEST requires a decision tree surrogate of the black-box model, whilst \threeS does not need to model the black-box predictive model.
AITEST defines data constraints by fitting different distributions to the features and using statistical testing to select the correct distribution. The dependencies are then captured by a DAG. \threeS does not require predefined constraints and dependencies, but aims to learn these implicitly with the generative model.

\textbf{\textit{Use cases}}

\begin{itemize}
    \item Group fairness:  AITEST aims to probe if a model does have a group fairness issue or not. The goal of \threeS is different — even if models don’t have group bias issues, with \threeS we desire reliable performance metric estimates (accuracy or even fairness) which are similar to the oracle estimates on small and intersectional subgroups for which we have limited real test data. 
\item Adversarial robustness: AITEST does this by generating more inputs in the neighborhood of a specific sample and seeing if they behave the same. In reality, this is analogous to group-wise testing with $n=1$, a very specific type of group testing. In contrast, with \threeS we explore multiple definitions of groups from specific sensitive attributes, to intersectional groups, to points of interest (i.e. $n=1$), to high- and low density regions.
\item AITEST does not account for distribution shift, unlike \threeS which looks at distribution shift with no prior knowledge and high-level knowledge.
\end{itemize}

\begingroup
\setlength{\tabcolsep}{4pt} 
\renewcommand{\arraystretch}{1.1} 
\begin{table*}[!h]
\centering
\caption{Comparing \threeS to other tabular approaches of generating synthetic test data. $f$ is the trained predictive model and we abbreviate $A(f;\cD) = A(f;\cD, \Omega)$ for evaluating $f$ over all of $\cD$ (Eq. 1). (i) used for evaluating subgroups, (ii) used for evaluating shifts, (iii) does not require discretization of continuous features, (iv) does not require modeling black-box $f$}
\rotatebox{0}{
\scalebox{0.6}{
\begin{tabular}{l|l|c|c|c|c|c|c}
\toprule
        Examples
      & Inputs 
      & (i) 
      & (ii)
      & (iii) 
      & (iv)
      & Generator Type
      & Goal
                        \\ \midrule
\multirow{2}{*}{\threeS}  &  \multirow{2}{*}{ \makecell[l]{$\cD_{train,G} = \cD_{test,f}$ (Any dataset) \\ Subgroups: $S$,  Shifts: No/Partial knowledge}}    & \multirow{2}{*}{$\checkmark$} & \multirow{2}{*}{$\checkmark$} & \multirow{2}{*}{$\checkmark$} & \multirow{2}{*}{$\checkmark$} & \multirow{2}{*}{GAN} & \makecell[l]{Reliable subgroup performance estimates for $f$, i.e. choose \\ $\cD_{syn}\sim p_G$ s.t. $A(f;\cD_{syn}, \mathcal{S})] \approx \mathbb{E}_{\cD\overset{iid}{\sim} p_R}[A(f;\cD, \mathcal{S})]$}  \\ \cline{8-8} 
& & & & & & & \makecell[l]{Estimate performance of $f$ under shift $T$, i.e. choose \\ $\cD_{syn}\sim p_G$ s.t. $A(f;\cD_{syn}, \mathcal{S})] \approx \mathbb{E}_{\cD\overset{iid}{\sim} p^s}[A(f;\cD, \mathcal{S})]$} \\
\hline \hline
\makecell[l]{DataSynthesizer \\ \cite{howe2017synthetic}} & \makecell[l]{Privacy-sensitive $\cD$ \\ Full knowledge of shift} & $\times$ & $\checkmark$ & $\times$ & $\checkmark$ & Bayesian network & \makecell[l]{Generate private data, extensions for generating \\ pathological data through (i) manual editing, (ii) inserting \\ extreme values/missingness, and (iii) .} \\\hline 
\makecell[l]{AITEST \\ \cite{saha2022data}} & \makecell[l]{$\cD_{train,G} = \cD_{train,f}$, \\ Constraints and dependencies (DAG)} & $\checkmark$ & $\times$ & $\times/\checkmark$ & $\times$ & \makecell[l]{Sample features \\ sequentially \\ following DAG} & \makecell[l]{Subgroup performance, e.g. fairness between two\\ sensitive groups ($S_1,S_2$),
i.e.  $\frac{A(f; \cD_{Syn}, S_1)}{A(f; \cD_{Syn}, S_2)}$}\\  
\bottomrule
\end{tabular}}}
\label{tabular-related_work}
\end{table*}
\endgroup
\clearpage
\section{Experimental Details} \label{appx:experimental_details}
This appendix includes details on the experiments, including (i) the datasets, and (ii) the different settings of the experiments, including the implementation of baselines. 
\subsection{Datasets}
Here we describe the real-world datasets used in greater detail.

\textbf{ADULT Dataset}
The ADULT dataset \citep{uci} has 32,561 instances with a total of 13 attributes capturing demographic (age, gender, race), personal (marital status) and financial (income) features amongst others. The classification task predicts whether a person earns over \$50K or not. We encode the features (e.g. race, sex, gender, etc.) and a summary can be found in Table \ref{tab:adult_features}.

Note that there is an imbalance across certain features, such as across different race groups, which is what we evaluate. 

\begin{table}[hbt]
\centering
\caption{Summary of features for the Adult Dataset \citep{uci}}
\begin{tabular}{ll}
\toprule
Feature & Values/Range \\ 
\midrule
Age & $17-90$ \\ 
education-num & $1-16$ \\ 
marital-status & $0, 1$ \\ 
relationship & $0, 1, 2, 3, 4$ \\ 
race & $0, 1, 2, 3, 4$ \\ 
sex & $0,1$ \\ 
capital-gain & $0,1$ \\ 
capital-loss & $0,1$ \\ 
hours-per-week & $1-99$ \\ 
country & $0,1$ \\ 
employment-type & $0, 1, 2, 3$ \\ 
salary & $0,1$ \\ 
\bottomrule
\end{tabular} 
\vspace{.5cm}
\label{tab:adult_features}
\end{table}

\textbf{Covid-19 Dataset}
The Covid-19 dataset \citep{baqui2020ethnic} consists of Covid patients from Brazil. The dataset is publicly available and based on SIVEP-Gripe data \citep{sivep}. The dataset consists of 6882 patients from Brazil recorded between February 27-May 4 2020. The dataset captures risk factors including comorbidities, symptoms, and demographic characteristics. There is a mortality label from Covid-19 making it a binary classification task.  A summary of the characteristics of the covariates can be found in Table \ref{tab:covid_features}.

\begin{table}[ht]
\centering
\caption{Summary of features for the Covid-19 Dataset \citep{baqui2020ethnic}}
\scalebox{0.8}{
\begin{tabular}{ll}
\toprule
Feature & Range \\ 
\midrule
Sex & 0 (Female), 1(Male) \\
Age & $1-104$ \\ 
Fever & $0,1$ \\ 
Cough & $0,1$ \\ 
Sore throad & $0,1$ \\ 
Shortness of breath & $0,1$ \\ 
Respiratory discomfort & $0,1$ \\ 
SPO2 & $0-1$ \\ 
Diharea & $0,1$ \\ 
Vomitting & $0,1$ \\ 
Cardiovascular & $0,1$ \\ 
Asthma & $0,1$ \\ 
Diabetes & $0,1$ \\ 
Pulmonary & $0,1$ \\ 
Immunosuppresion & $0,1$ \\ 
Obesity & $0,1$ \\ 
Liver & $0,1$ \\ 
Neurologic & $0,1$ \\ 
Branca (Region) & $0,1$ \\ 
Preta (Region) & $0,1$ \\ 
Amarela (Region) & $0,1$ \\ 
Parda (Region) & $0,1$ \\ 
Indigena (Region) & $0,1$ \\ 
\bottomrule
\end{tabular}}
\vspace{.5cm}
\label{tab:covid_features}
\end{table}

\textbf{SEER Dataset} 
The SEER dataset is a publicly available dataset consisting of 240,486 patients enrolled in the American SEER program~\citep{seer}. The dataset consists of features used to characterize prostate cancer, including age, PSA (severity score), Gleason score, clinical stage, and treatments. A summary of the covariates can be found in Table \ref{tab:seer_features}. The classification task is to predict patient mortality, which is a binary label.

The dataset is highly imbalanced, where $~94\%$ of patients survive. Hence, we extract a balanced subset of 20,000 patients (i.e. 10,000 with label=0 and 10,000 with label=1).

\begin{table}[h]
\caption{Summary of features for the SEER \citep{seer} and CUTRACT \cite{cutract} datasets. \textit{Note: the range of age starts slightly lower for SEER (37-95) compared to CUTRACT (44-95).}}
\scalebox{0.8}{
\begin{tabular}{ll}
\toprule
Feature & Range \\ 
\midrule
Age & $37-95$ \\ 
PSA & $0-98$ \\ 
Comorbidities & $0, 1, 2, \geq 3$ \\ 
Treatment & Hormone Therapy (PHT), Radical Therapy - RDx (RT-RDx), \newline Radical Therapy -Sx (RT-Sx), CM \\ 
Grade & $1, 2, 3, 4, 5$ \\ 
Stage & $1, 2, 3, 4$ \\ 
Primary Gleason & $1, 2, 3, 4, 5$ \\ 
Secondary Gleason & $1, 2, 3, 4, 5$ \\ 
\bottomrule
\end{tabular}}
\label{tab:seer_features}
\end{table}

\textbf{CUTRACT Dataset}
The CUTRACT dataset is a private dataset consisting of 10,086 patients enrolled in the British Prostate Cancer UK program~\citep{cutract}. It includes the same features as SEER and also uses mortality as the label, see Table \ref{tab:seer_features}.

The dataset is highly imbalanced in its labels, hence we choose to extract a balanced subset of 2,000 patients (i.e. 1000 with label=0 and 1000 with label=1).

\subsection{Experiments}
For specifics on how $G$ is evaluated, tuned, and selected, please see Appendix \ref{quality-discussion}.

\subsubsection{Experiment 5.1: Subgroups}
In this experiment, we evaluate the performance estimates on different subgroups based on the mean absolute error compared to the estimates of subgroup performance using the oracle dataset. In order, to represent potential variation of selecting different test sets, we repeat the experiment 10 times, where we sample a different test set in each run. That being said, we keep the proportions in each dataset fixed such that $\{\cD_{train, f}$,$\cD_{test, f}$,$\cD_{oracle}\}$ = $\{8.4k, 2.1k, 19.6k\}$.  Given that minority subgroups have few samples, in this experiment, we generate $n$ samples for each subgroup, where $n$ is the size of the largest subgroup in $\cD_{test, f}$. This allows us to ``balance'' the evaluation dataset.

We provide more details on the subgroups below for each dataset:

\begin{table}
    \caption{Details on dataset specific subgroups}
    \centering
    \scalebox{0.65}{
    \begin{tabular}{ccc}
    \hline
        Dataset & Subgroup & Specifics \\
          \hline
      Adult   & Race &   " White",
            "Amer-Indian-Eskimo",
            "Asian-Pac-Islander",
            "Black",
            " Other" \\
      Drug   & Ethnicity & 'White', 'Other', 'Mixed-White/Black', 'Asian',
       'Mixed-White/Asian', 'Black', 'Mixed-Black/Asian' \\
     Covid    &  Region & 'Branca', 'Preta', 'Amarela', 'Parda', 'Indigena' \\
     Support    & Race & 'white','black', 'asian', 'hispanic' \\
     Bank    & Employment status (Anon) & 'CA', 'CB', 'CC', 'CD', 'CE', 'CF', 'CG' \\
       \hline
    \end{tabular}}
    \label{tab:my_label}
\end{table}

In producing the intersectional performance matrix, we slice the data for these intersections. However, as we slice the data into finer intersections, the intersectional groups naturally become smaller. Hence, to ensure we have reliable estimates, we set a cut-off wherein we only evaluate performance for intersectional groups where there are 100 or more samples. In computing the mean absolute error, we do not include the corresponding intersections for which there were insufficient samples.

For Section \ref{exp:coverage},
we use bootstrapping for the naive $\cD_{test,f}$ baseline. Bootstrapping is a prevalent method for providing uncertainty \cite{efron1983estimating,efron1992bootstrap,diciccio1996bootstrap,carpenter2000bootstrap,miller2021model}. In our case, we sample a dataset $\cD_{test,f}^k$ the same size as $\cD_{test,f}$, uniformly with replacement from $\cD_{test,f}$. Model $f$ is evaluated on each $\cD_{test,f}^k$, and the mean and standard deviation across $\cD_{test,f}^k$ is used to construct confidence intervals. For all methods, intervals are chosen as the mean $\pm$ 2 standard deviations.

\subsubsection{Experiments 5.2 Distributional shifts}
\textbf{No prior knowledge: characterizing sensitivity across operating ranges.}
In this experiment, we assess performance for different degrees of shift, we compute three shift buckets around the mean of the original feature distribution: large negative shift from the mean (\textbf{-}), small negative/positive shift from the mean (\textbf{$\pm$}), and large positive shift from the mean (\textbf{+}). 

We define each in terms of the feature quantiles. We generate uniformly distributed shifts (between min(feature) and max(feature)). Any shift that shifts the mean to less than Q1 is (\textbf{-})
, any shift that shifts the mean to more than Q3 is (\textbf{+}) and any shift in between is (\textbf{$\pm$}). The Oracle target is created using rejection sampling of the oracle source data, see Section \ref{appx:rejection_sampling}.

\textbf{Prior Knowledge.}
In this experiment, we assume we observe some of the features in the target domain, i.e. we observe the empirical marginal distribution of $X_c$. This empirical marginal distribution is used to sample from, and conditioned on when generating the other features. The Source RS target is created using rejection sampling of the test data, see Section \ref{appx:rejection_sampling}.

\subsection{Rejection sampling for creating shifted datasets} \label{appx:rejection_sampling}
In experiments 5.2 and 5.3, we use rejection sampling for the oracle and source baselines, respectively. Let us briefly explain how this is achieved. 

Let $\mathcal{D}$ be some dataset with distribution $p^0(X)$ that can be split into parts $p(X_{\Bar{c}}|X_c)$ and $p(X_c)$. As noted in Section \ref{sec:formulation}, we assume the latter changes (inducing a shift in distribution), while the former is fixed. We denote the shifted marginal distribution as $p^s(X_c)$ and the full distribution as $p^s(X) = p^s(X_c)p(X_{\Bar{c}}|X_c)$.

In experiment 5.2, we desire a ground-truth target dataset for a given shift. We do not have data from $p^s(X)$, however we can use rejection sampling to \emph{create} such dataset, which we denote by $\mathcal{D}^{s}$. Since we do not know $p^0(X)$ either, we sample from the empirical distribution, i.e. from data $\cD$ itself, which will converge to the true distribution when $|\cD|$ becomes large. To approximate $p^0(X_c)$, we train a simple KDE model and $p^s(X_c)$ is defined by shifting this distribution (see Section \ref{sec:method_shifts_general}). This gives the following algorithm:

\begin{algorithm}[H]
\begin{algorithmic}
    \State \textbf{Input} Source dataset $\cD$, shift $T$ and desired shifted set size $n_s$\;
    \State Fit density model $\hat{p}^0(X_c)$ to $\{\bf{x}_c|\bf{x}\in \cD\}$\;
    \State $\hat{p}^s(X_c)\leftarrow T(\hat{p}^0)(X_c)$\; 
    \State $M \leftarrow \max_{\bf{x}_c\in \cD}\frac{\hat{p}^s(\bf{x}_c)} {\hat{p}^0(\bf{x}_c)}$\;
    \State $\cD^s\leftarrow \{ \}$\;
    \While
    {$|\cD^s|<n_s$}{
        \State Sample $\bf{x}$ from $\cD$ uniformly\;
        \State Sample $u\sim U(0,1)$\;
        \If{$\frac{\hat{p}^s(\bf{x}_c)}{\hat{p}^0(\bf{x}_c)}>M u$}{
            $\cD^s \leftarrow \cD^s\cup \{\bf{x}\}$\;   
        }
        \EndIf
        }
        \EndWhile
    \State \textbf{return} $\cD^s$
\end{algorithmic}
    \caption{Rejection sampling from source dataset $\cD$, given a predefined marginal shift $T$ and desired test set size.}
\end{algorithm}

In Experiment 5.2 we run the above with $\cD$ an oracle test set. Since the oracle test set is very large, it covers $p^0$ relatively well. This allows us to approximate $p^0(X_c)$, and also means that draws from the empirical distribution are distributed approximately like the true underlying distribution. 

In experiment 5.3 we use a similar set-up for creating baseline \emph{Source (RS)} based on $\cD_{test,f}$ alone, and to have a fair comparison we use rejection sampling to weigh the points.\footnote{Effectively, this reduces to an importance weighted estimate of the performance.} In this case, however, the distribution $p^0(X_c)$, and in turn $p^s(X_c)$, cannot be approximated accurately. In addition, we may have very little data such that the same points need to be included many times (in regions with large $p^s(X_c)/p^0(X_c)$). As a result, although we see that \emph{Source (RS)} performs better than unshifted \emph{Source (all)}, it is a poor evaluation approach.

\clearpage
\section{Generative Model Choice}

\label{appx:generative_model}

Any generative model can be used to produce the synthetic test data, but some models may be better or worse than others. \threeS uses CTGAN \citep{xu2019modeling} since this model is designed specifically for tabular data and has shown good performance. In this section, we explain how it is tuned and include a comparison to other generative models.

\subsection{Assessing the quality of generative model G in 
3S.}\label{quality-discussion} 
Approaches to model selection and 
quality assessment of generative models often measure the distance between the 
generated and the true distributions \citep{borji2019pros}. 
In \threeS, we use Maximum Mean Discrepancy (MMD) \citep{gretton2012kernel}, a popular choice for synthetic data quality \citep{sutherland2017generative,bounliphone2016test}. 

MMD performs a statistical test on distributions $P^r$ (Real) and $P^g$ (Generated),  measuring the difference of their expected function values, with a lower MMD implying
$P^g$ is closer to  $P^r$.  

We use MMD in our auto-tuning and model selection step,  comparing the generated data to a held-out test set, with $G$ selected as the model with the lowest MMD.  This step also serves to ensure that the data generated by \threeS is indeed close to the real-world reference dataset of interest.
Specifically, hyper-parameters of \threeS when training $G$ are tuned via a Tree-structured Parzen Estimator. We search over the number of epochs of training [100, 200, 300, 500], learning rate [2e-4, 2e-5, 2e-6], and embedding dimension [64,128,256]. For all methods, we have a small hyper-parameter validation set with a size of 10\% of the training dataset. Our objective is based on MMD minimization.

That said, of course, alternative widely used metrics such as Inverse KL-Divergence or the Jensen-Shannon divergence could also be used as metrics of assessment.

\subsection{Influence of model choice.} Any generative model can be used as the core of \threeS. For efficiency, a conditional generative model is highly desirable; this allows direct conditioning on subgroup or shift information, and not e.g. post-generation rejection sampling. Furthermore, some generative models may provide more or less realistic data. Here we compare \threeS estimates provided by CTGAN, vs estimates given by TVAE and Normalizing Flows.

We assess these different base models for $G$ on the race subgroup task from Sec. \ref{exp-d1}. Using \threeS can assess the generative models based on MMD, but for completeness we also show inverse KL-divergence and Jensen-Shannon Divergence (JSD), where the metrics are computed vs a held-out validation dataset. 

We show in Table \ref{table:diff-G}, that the better quality metric  does indeed translate into better performance when we use the synthetic data for model evaluation. We find specifically that CTGAN outperforms the other approaches, serving as validation for our selection. 

Additionally, the results highlight that for practical application, one could evaluate the quality of $G$ first using metrics such as MMD, inverse KLD, or JSD, as a proxy for how well the generative model should perform.

We assume for the purposes of this experiment that the three classes of models are trained with the same optimization hyperparameters (epochs=200, learning rate=2e-4).

\begin{table}[hbt]
    \centering
\caption{\footnotesize{Assessing the influence of model choice for $G$ and illustrating how our quality assessment metrics in \threeS can be used to select the best model which indeed will provide the best performance. We see that indeed CTGAN performs best in this case.}}
\scalebox{0.8}{

\vspace{-5mm}

\begin{tabular}[b]{c|ccc|c|c}

     Base $G$            & MMD $\downarrow$ & Inverse KLD $\uparrow$ & JSD $\downarrow$     & Subgroup (\%)              &  Mean Absolute Error \% $\downarrow$   \\ \hline
\multirow{4}{*}{CTGAN} & \multirow{4}{*} {\textbf{0.0014}} & \multirow{4}{*}{\textbf{0.995}} & \multirow{4}{*}{\textbf{0.03}}
 & \#1 (86\%)            &       \textbf{ 0.28 $\pm$ 0.24}                     \\ 
& & &  & \#2 (9\%)                    & \textbf{ 17.64 $\pm$ 0.29 }                         \\ 
& & &  & \#3 (3\%)                     & \textbf{2.96 $\pm$ 1.02         }                               \\ 
& & & & \#4 (1\%)                     & \textbf{1.14 $\pm$ 0.62         }                         \\ 
& & & & \#5 (1\%)                     & \textbf{1.03 $\pm$ 0.85      }                              \\ 
 \hline
 
\multirow{4}{*}{NF} & \multirow{4}{*}{0.0034} & \multirow{4}{*}{0.970} & \multirow{4}{*}{0.09}
 & \#1 (86\%)            &        16.25 $\pm$ 0.53                  \\ 
& & &  & \#2 (9\%)                     & 26.35 $\pm$ 1.07                                   \\ 
& & &  & \#3 (3\%)                     & 20.50 $\pm$ 3.31                                   \\ 
& & & & \#4 (1\%)                     & 27.14 $\pm$ 0.97                                 \\ 
& & & & \#5 (1\%)                     & 26.04 $\pm$ 2.87                                     \\ 
 \hline
\multirow{4}{*}{TVAE} & \multirow{4}{*}{0.4557} & \multirow{4}{*}{0.4987} & \multirow{4}{*}{0.505}
 & \#1 (86\%)            &        25.93 $\pm$ 1.34                        \\ 
& & &  & \#2 (9\%)                     & 35.40 $\pm$ 0.83                          \\ 
& & &  & \#3 (3\%)                     & 24.05 $\pm$ 1.12                                         \\ 
& & & & \#4 (1\%)                     & 33.0 $\pm$ 0.31                                     \\ 
& & & & \#5 (1\%)                     & 37.69 $\pm$ 0.61                                             \\ 
 \hline

\end{tabular}

}

\label{table:diff-G}
\end{table}

\textbf{Why do small changes in metrics, lead to large MAE differences?} One might wonder that small changes in for example JSD lead to large MAE. For instance, CTGAN to NF of 0.03 to 0.09. We assess the sensitivity of performance estimates to small changes in divergence measures. 

To do so, we conduct an experiment where we synthetically increase the JSD of the synthetic dataset through corruption. We then compute the MAE performance estimate on the synthetic dataset compared to the Oracle for different JSD values.

\begin{figure}[H]
\vspace{-0mm}
\captionsetup{font=footnotesize}
\centering
 \includegraphics[width=0.45\textwidth]{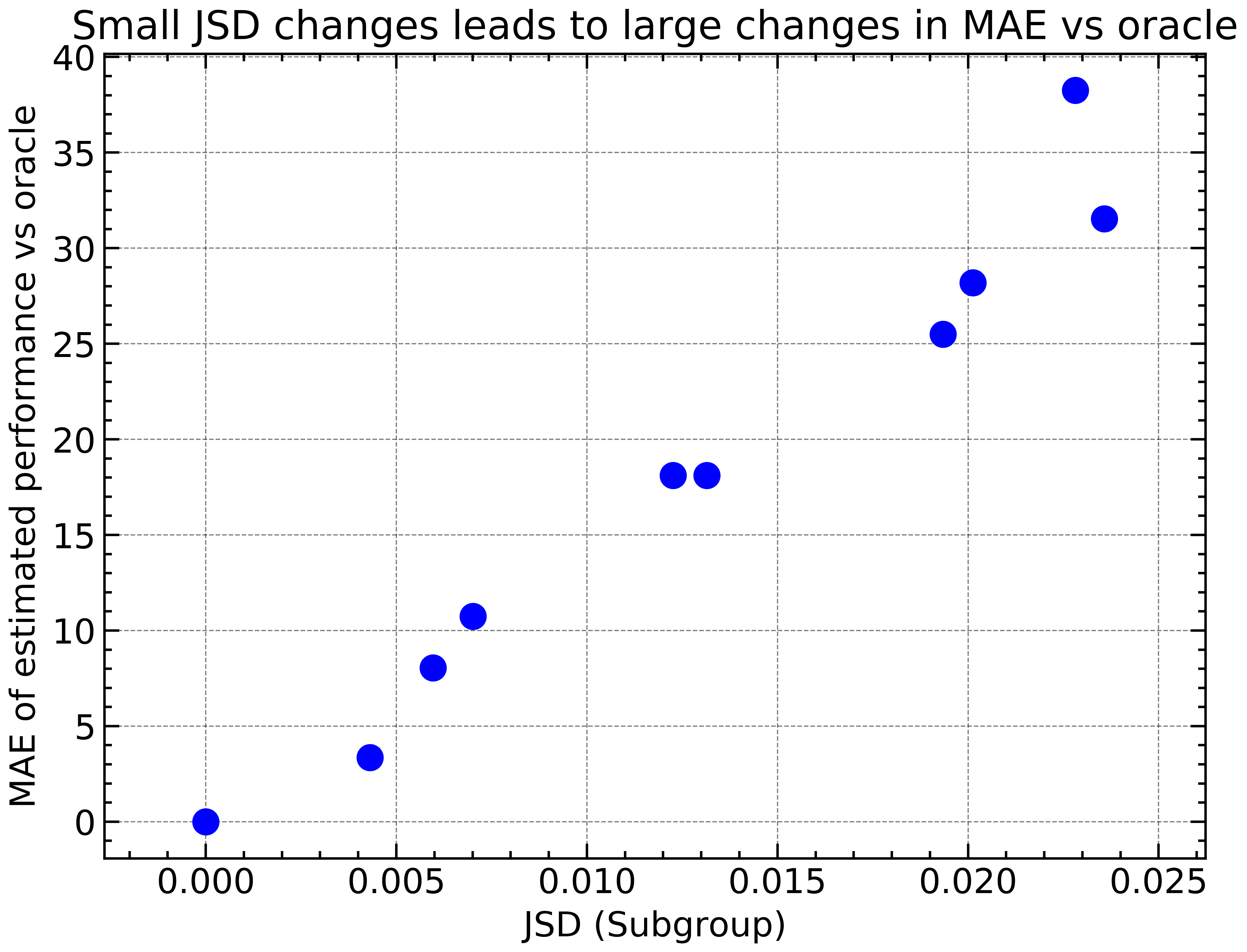}
 \vspace{-0mm}
\caption{\footnotesize{Small changes in JSD lead to large changes in MAE vs oracle.}}\label{jsd}
\vspace{-0mm}
\rule{0.33\textwidth}{0.1pt}
\vspace{-0mm}
\end{figure}

The result in Figure \ref{jsd} shows that even small changes in the JSD can significantly harm the MAE. This provides an explanation of why a method with seemingly relatively similar JSD (or divergence metric), with only a minor difference, may perform very differently as measured by MAE. This motivates why we select the generative model, for example with the lowest metric despite them looking similar.

\clearpage
\section{Limitations and Failure Cases} \label{appx:limitations}
\subsection{Limitations}
We summarise 3S’s main limitations (where applicable referencing limitations already outlined in the paper).
\begin{enumerate}
    \item \textbf{Subgroups in real test set are large and 3S is unnecessary.} When the aim is subgroup evaluation and there is sufficient real data for this subgroup, the real data estimate on $\mathcal{D}_{test,f}$ will be sufficiently accurate and there is no need to use 3S given the computational overhead. See*Failure Case 1 for new experiments and further discussion. Note, even for very large datasets, there can be very sparse regions or small subgroups for which using 3S is still beneficial. Also note that this limitation mostly applies to subgroup evaluation and less to Generating Synthetic Data with Shifts, because performance estimates for possible shifts is less trivial using real data alone (e.g. require reweighting or resampling of test data) and we show (Sec. 5.2, Table \ref{table:shift_buckets} and Fig. \ref{fig:exp3b}) that 3S beats real data baselines consistently.

\item \textbf{Possible errors in the generative process}. Errors in the generative process can affect downstream evaluation. This is especially relevant for groups or regions with few samples, as it is more likely a generative model does not fit the distribution perfectly here. By replacing the generative model by a deep generative ensemble \cite{breugel2023dge}, 3S provides insight into its own generative uncertainty. When there is very little data and 3S’s uncertainty bounds are too large, practitioners should consider gathering additional data. See Failure Case 2 below.

\item \textbf{Not enough knowledge or data to generate realistic shifts.} The success of modelling data with distributional shifts relies on the validity of the distributional shift’s assumptions. This is true for 3S as much as for supervised works on distributional shifts (e.g. see \cite{varshney2021trustworthy}). In Sec. 5.2.2 we show how a lack of shift knowledge may affect evaluation. We aimed to generate data to test a model on UK patients, using mostly data from US patients. We do not define the shift explicitly—we only assume that we observe a small number (1 to 4) of features $X_c$ (hence the conditional distribution of the rest of the features conditional on these features is the same across countries). In Fig. 6b, we show the assumption does not hold when we only see one feature—this lack of shift knowledge is a failure case. When we add more features, the assumption is more likely to approximately hold and 3S converges to a better estimate. We reiterate that invalid shift assumptions are a limitation of any distributional shift method, e.g. the rejection sampling (RS) baseline using only real data is worse than 3S overall.

\item \textbf{Computational cost}. The computational cost and complexity of using generative models is always higher than using real test data directly. For typical tabular datasets, the computational requirements are often very manageable: under 5 min for most datasets and under 40 min for the largest dataset Bank (Table \ref{time}). The reported times correspond to the dataset sizes in the paper.  We would like to emphasise that the cost at deployment time for a poorly evaluated model can be unexpectedly high, which warrants 3S’s higher evaluation cost. 
Additionally, pre-implemented generative libraries (e.g. Patki 2016, Qian 2023) can accelerate the generative process, automating generator training with minimal user input.
\end{enumerate}

\begin{table}[!h]
\centering
\caption{Training time for generator on an NVIDIA GeForce RTX3080 for main paper settings.}
\begin{tabular}{lccccc}
\hline
Dataset           & Adult & Covid & Drug & Support & Bank \\ \hline
Tuning time (min) & 4.5   & 4     & 1.5  & 2       & 38   \\ \hline
\end{tabular}
\label{time}
\end{table}

\subsection{Failure cases}

\begin{figure*}[!h]

\centering
    \begin{subfigure}[t]{.3\textwidth}
        \includegraphics[width=\textwidth]{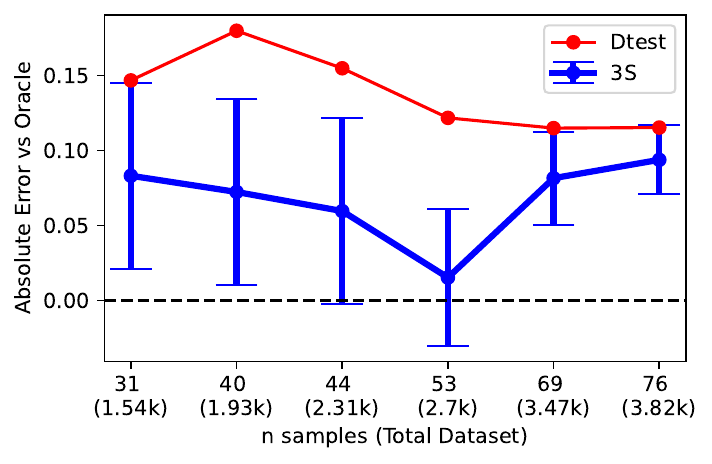}
        \caption{Covid}
    \end{subfigure}%
    ~
     \begin{subfigure}[t]{.3\textwidth}
        \includegraphics[width=\textwidth]{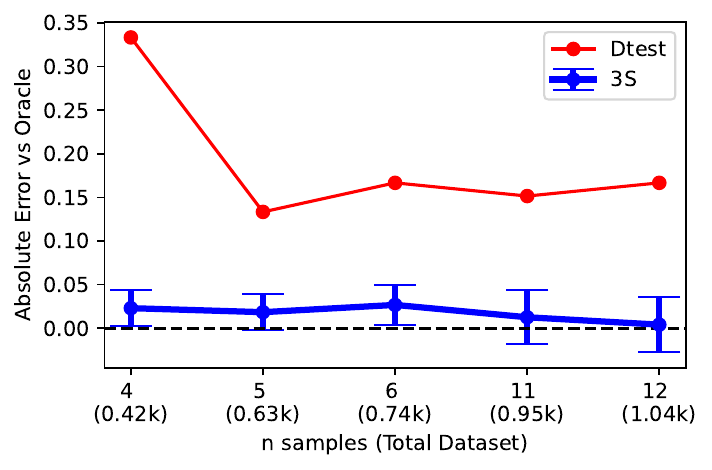}
        \caption{Drug}
    \end{subfigure}%
~
    \begin{subfigure}[t]{.3\textwidth}
        \includegraphics[width=\textwidth]{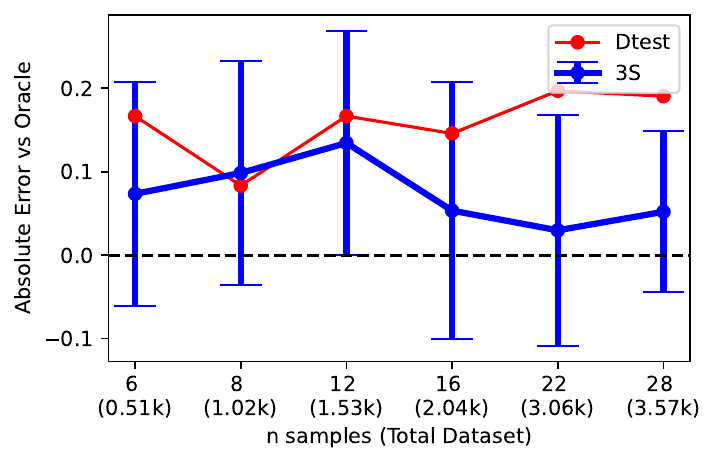}
        \caption{Support}
    \end{subfigure}%
     \vspace{-0mm}
    \caption{\footnotesize{Limitation Analysis: Extremely small sample size}}\label{fig:ultralow}
    \vspace{-0.5mm}
    \vspace{-0mm}
\vspace{-5mm}\end{figure*}

\begin{figure*}[hbt]
\vspace{-2mm}
\centering
 
   \begin{subfigure}[t]{.25\textwidth}
        \includegraphics[width=\textwidth]{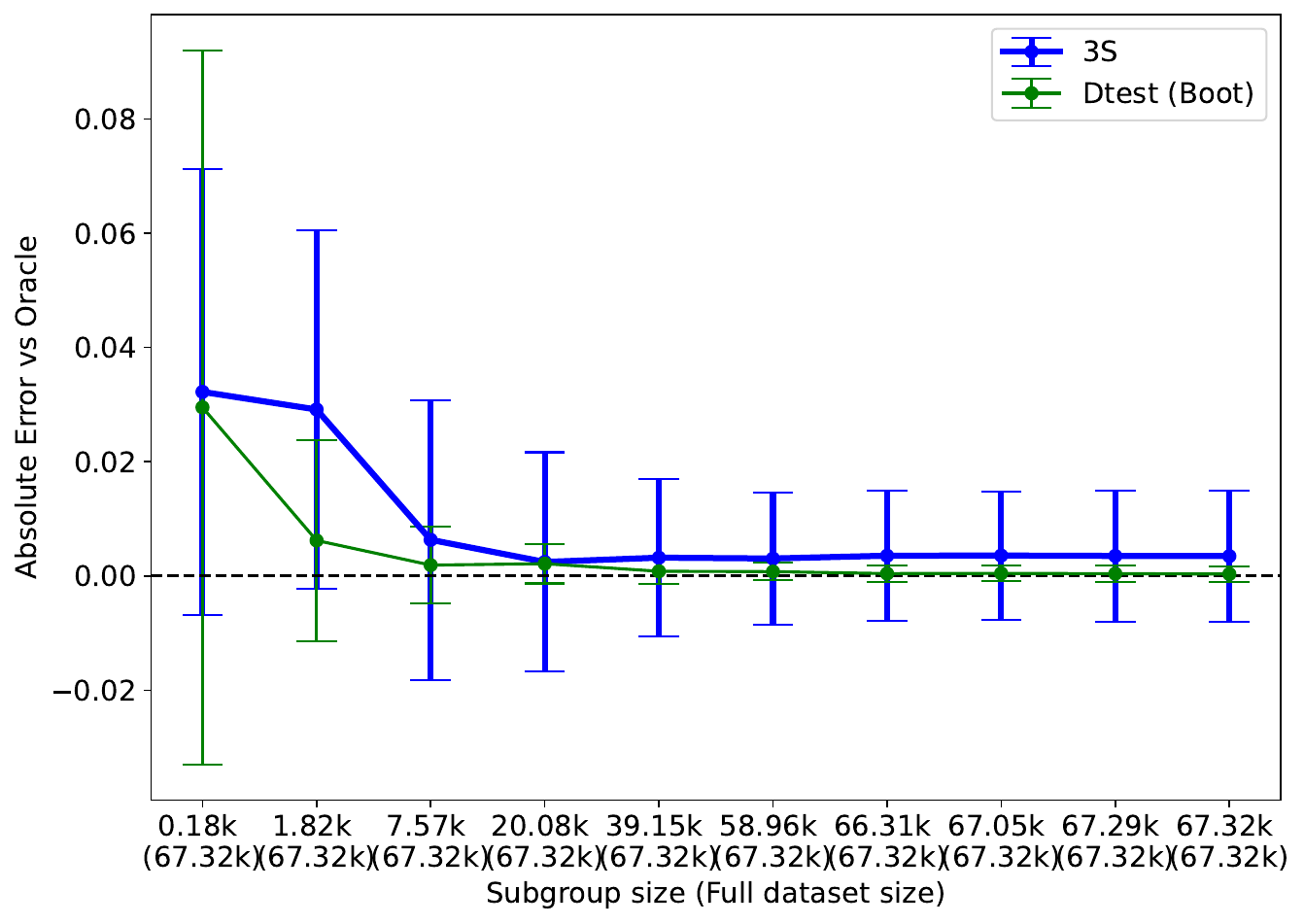}
        \caption{Credit risk vs Oracle}
    \end{subfigure}%
~
    \begin{subfigure}[t]{.25\textwidth}
        \includegraphics[width=\textwidth]{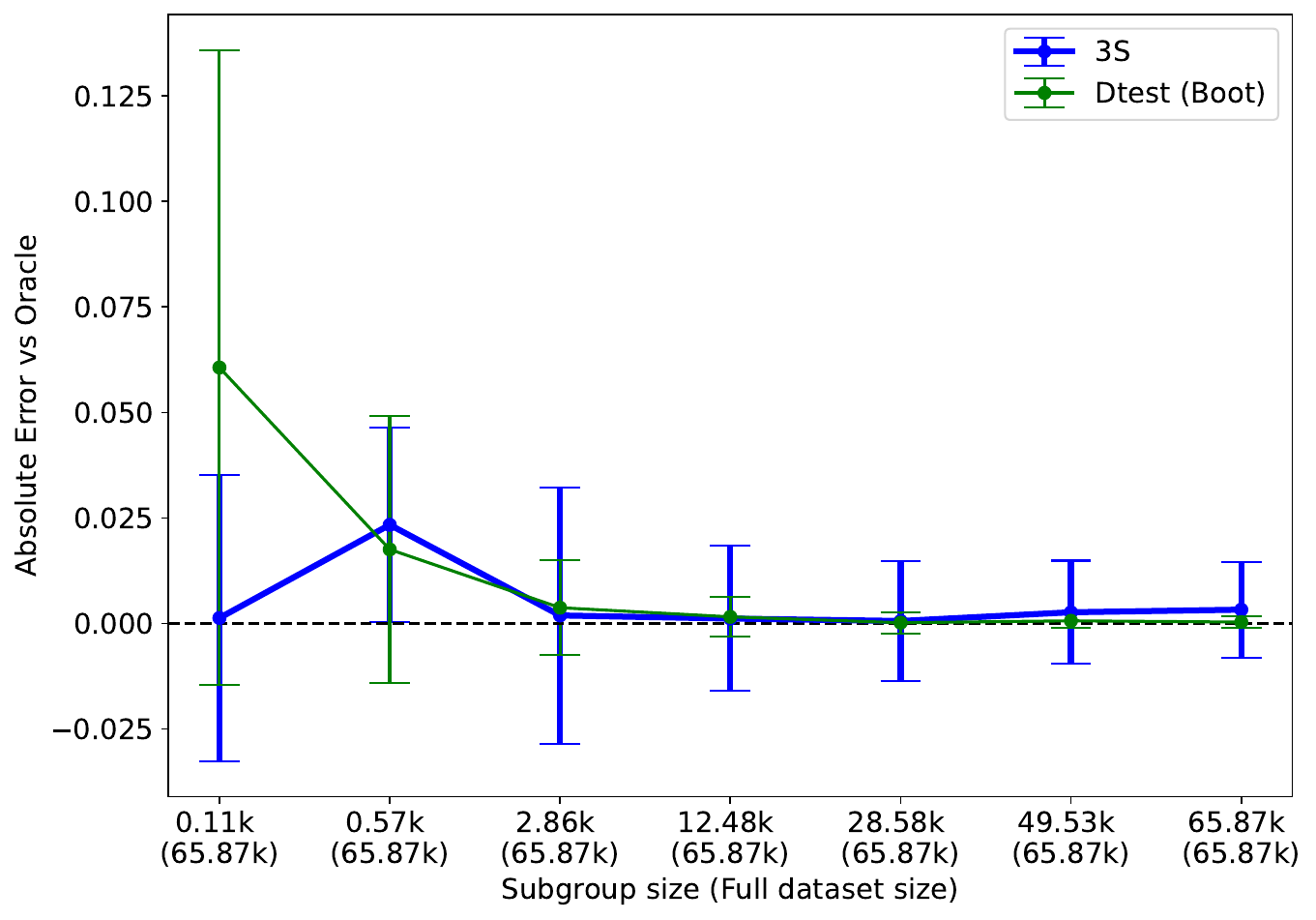}
        \caption{Age vs Oracle}
    \end{subfigure}%

     \vspace{-0mm}
    \caption{\footnotesize{Limitation analysis: Large sample size }}\label{fig:large}
    \vspace{-0.5mm}
    \vspace{-0mm}
\vspace{-5mm}\end{figure*}

For limitations 1 and 2 mentioned above, we include two new experiments highlighting failure cases on two extreme settings:

\begin{enumerate}
    \item \textbf{3S is similarly accurate to real data evaluation when the real test data is large. }
With sufficiently large real data, 3S provides no improvement despite higher complexity. We can determine sufficient size by estimating the variance $Var(A)$ of the performance metric $A(f;D,S)$ w.r.t. \textit{the random variable} denoting the test data $D$. With only access to one real dataset $D_{test}$ however, we can approximate $Var(A)$ via bootstrapping (Fig. 5). If $Var(A)$ falls below a small threshold $\alpha$ (e.g. 0.05), practitioners may for example decide to trust their real data estimate and not use 3S, as further evaluation improvements are unlikely. In our Bank dataset experiment (500k examples), we vary age and credit risk thresholds, retaining samples above each cut-off to shrink the test set (see Figure \ref{fig:large}). Note that for large datasets but very small subgroups, the $D_{test,f}$ estimate still has a high variance, (reflected in the large bootstrapped $Var(A)$ ), hence this should urge a practitioner to use 3S.

\item \textbf{Large uncertainty for very small test sets.}
At the other extreme, when there are too few samples the uncertainty of 3S (quantified through a DGE \cite{breugel2023dge}) can become too large. In Figure \ref{fig:ultralow} in which we reduce subgroups to fewer than 10 samples in the test data. We train $G$ in 3S on the overall test set (which includes the small subgroup) with $n_{samples}$. Despite good performance versus an oracle, the uncertainty intervals from 3S's ensemble span 0.1-0.2.  These wide intervals make the 3S estimates unreliable and less useful, and would urge a practitioner to consider gathering additional data. The key takeaway: With extremely sparse subgroups, the large uncertainties signal that more data should be gathered before relying on 3S's uncertain estimates. 
\end{enumerate}

\clearpage
\section{Additional Experiments} \label{appx:additional_experiments}
This appendix includes a number of additional results. First, we include additional results for the main paper's experiments, using different downstream models, predictors, metrics, and baselines. Second, we discuss other types of subgroup definitions, including subgroups based on points of interest and density of regions. Third, we include results for \threeS when we have access to the training set of predictor model $f$; since this dataset is usually larger, it can help train a more accurate generative model.

\subsection{Experimental additions to main paper results}

\subsubsection{Sanity check}
\begin{figure}[H]
\centering
    \begin{subfigure}[t]{.3\textwidth}
        \includegraphics[width=\textwidth]{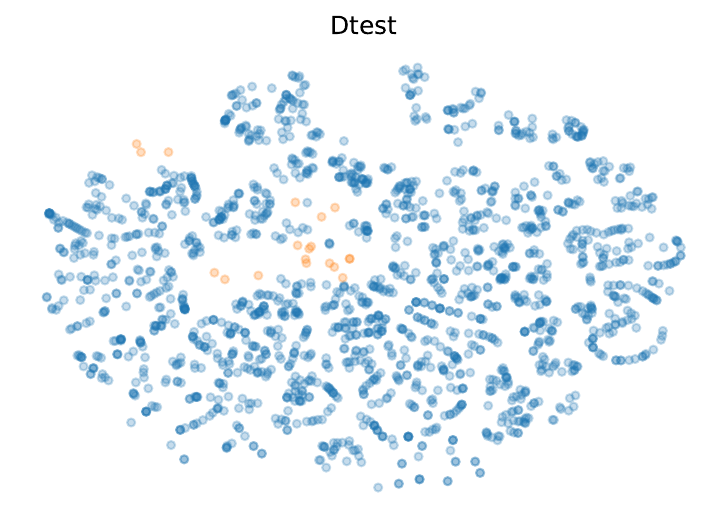}
        \caption{$\cD_{test, f}$}
    \end{subfigure}%
    ~
    \begin{subfigure}[t]{.3\textwidth}
        \includegraphics[width=\textwidth]{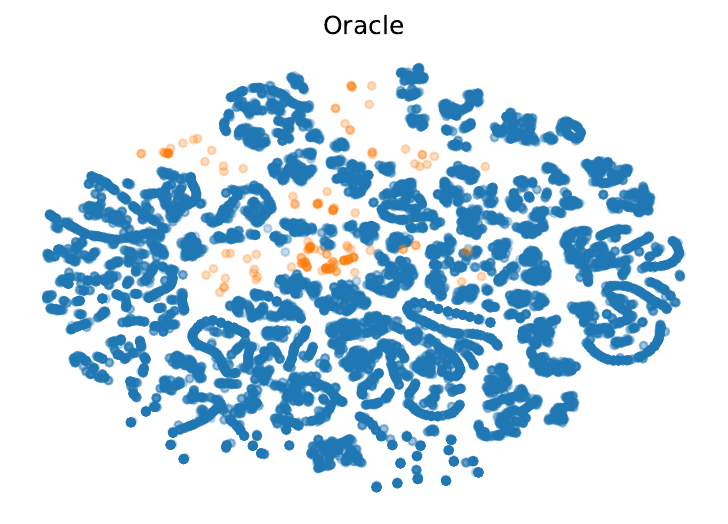}
        \caption{$\cD_{oracle}$}
    \end{subfigure}%
    ~
     \begin{subfigure}[t]{.3\textwidth}
        \includegraphics[width=\textwidth]{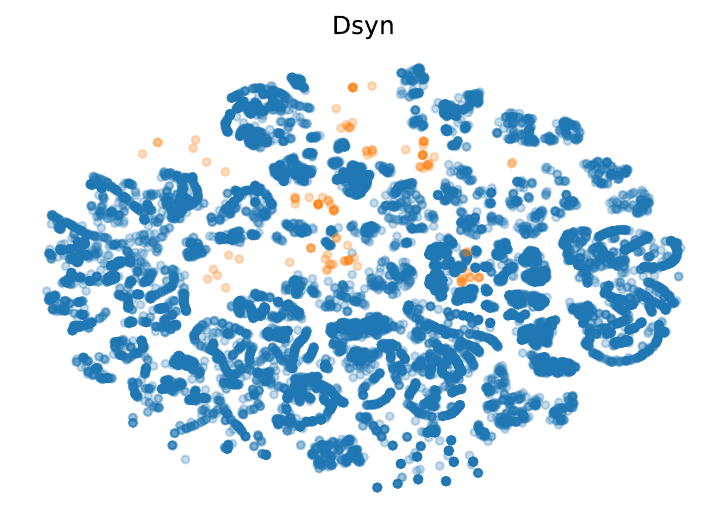}
        \caption{\threeS}
    \end{subfigure}%

    \caption{Qualitative assessment of synthetic data. T-SNE on the Adult dataset comparing the real test data $\cD_{test,f}$, oracle data $\cD_{oracle}$ and \threeS data $\cD_{syn}$. We find that \threeS generates synthetic test data that covers the oracle dataset well, despite only having access to $\cD_{test, f}$ during training. The data, evaluating subgroups where blue (Race 1 - majority) and orange (Race 5 - minority)}\label{fig:sanity}

    \rule{0.65\textwidth}{0.05pt}
\end{figure}

We perform a sanity check on the data generated by \threeS as compared to $\D_{oracle}$ and $D_{test}$. The results are visualized in Figure \ref{fig:sanity} and show that \threeS generates synthetic test data that covers the oracle dataset well, even on the minority subgroup in blue, capturing the distribution appropriately.

\subsubsection{Subgroup performance evaluation: more metrics and downstream models}

\textbf{Motivation.} The performance of the model, on subgroups, is likely influenced by the class of downstream predictive model $f$. We aim to assess the performance of the granular subgroups for a broader class of downstream models $f$. In addition, to accuracy, we assess F1-score performance estimates as another showcase of downstream performance estimates.

\textbf{Setup.} This experiment evaluates the mean performance difference for (i) \threeS, (ii) \threeS+, and (iii) $\cD_{test,f}$. We follow the same setup as the granular subgroup experiment in Section \ref{exp-d1}. We increase the predictive models beyond RF, MLP, and GBDT and further include SVM, AdaBoost, Bagging Classifier, and Logistic Regression.

\begin{figure}
\centering
    \begin{subfigure}[t]{1\textwidth}
        \includegraphics[width=\textwidth]{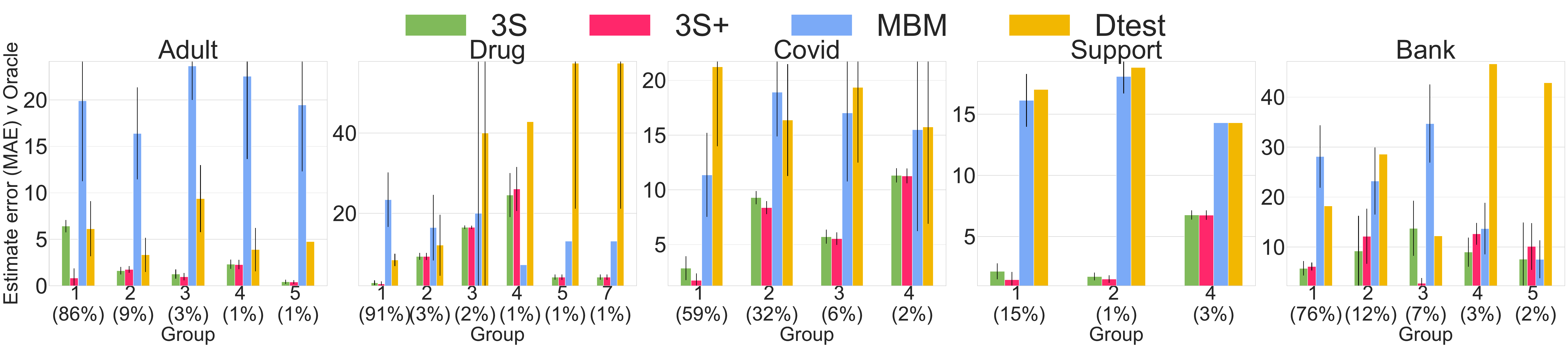}
        \caption{Random Forest}
    \end{subfigure}%
    \quad
    \begin{subfigure}[t]{1\textwidth}
        \includegraphics[width=\textwidth]{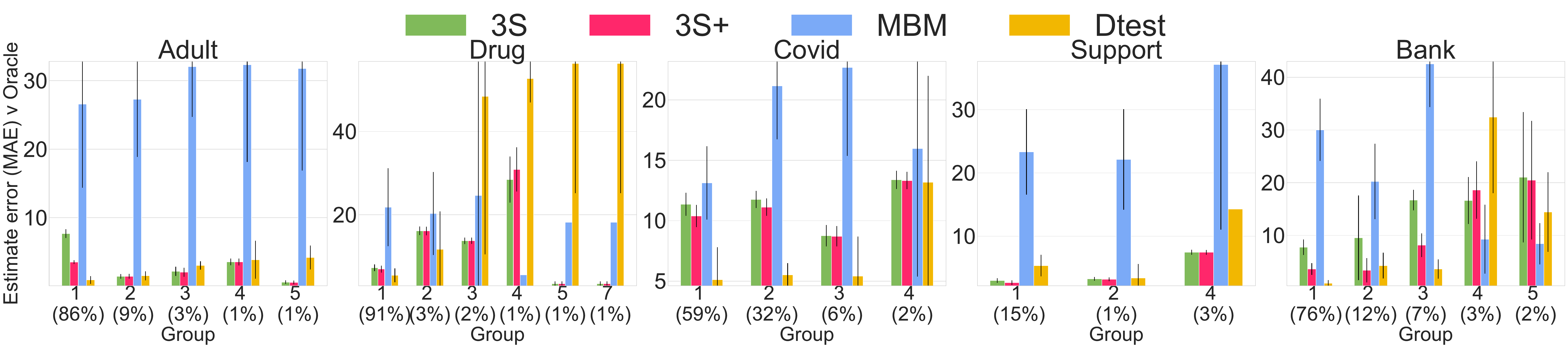}
        \caption{Gradient Boosting}
    \end{subfigure}%
    \quad
     \begin{subfigure}[t]{1\textwidth}
        \includegraphics[width=\textwidth]{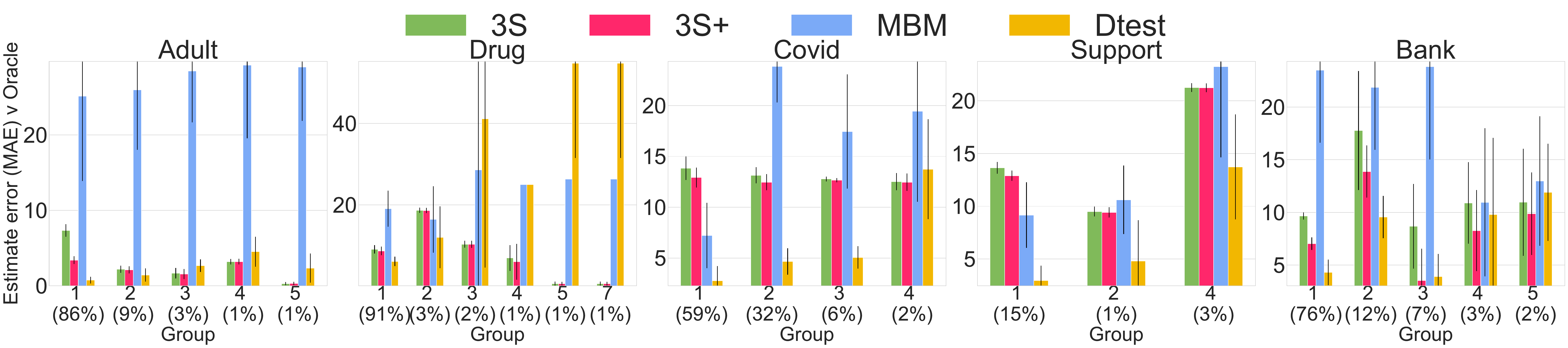}
        \caption{MLP}
    \end{subfigure}
    \quad
     \begin{subfigure}[t]{1\textwidth}
        \includegraphics[width=\textwidth]{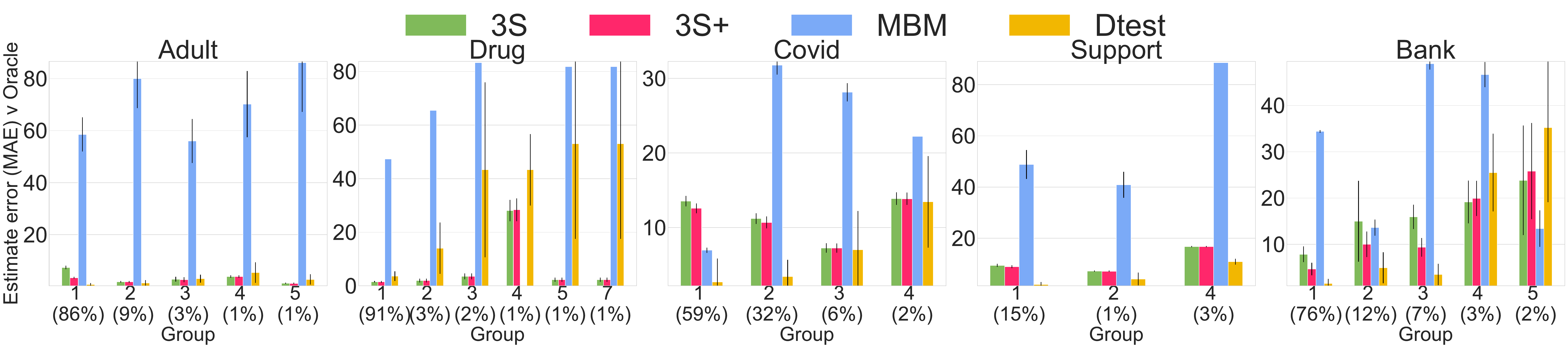}
        \caption{Adaboost}
    \end{subfigure}
    \quad
     \begin{subfigure}[t]{1\textwidth}
        \includegraphics[width=\textwidth]{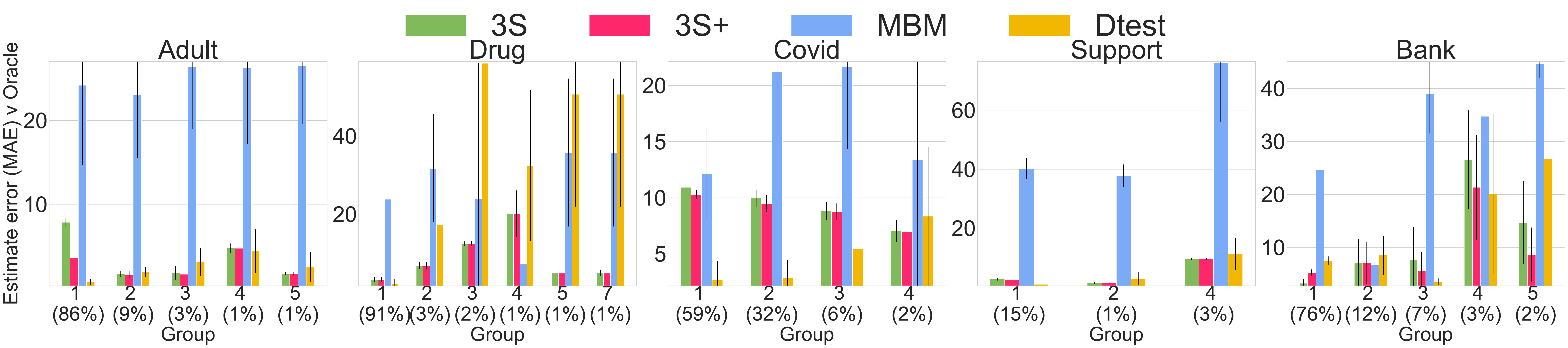}
        \caption{Logistic regression}
    \end{subfigure}
        \quad
    \quad
    \begin{subfigure}[t]{1\textwidth}
        \includegraphics[width=\textwidth]{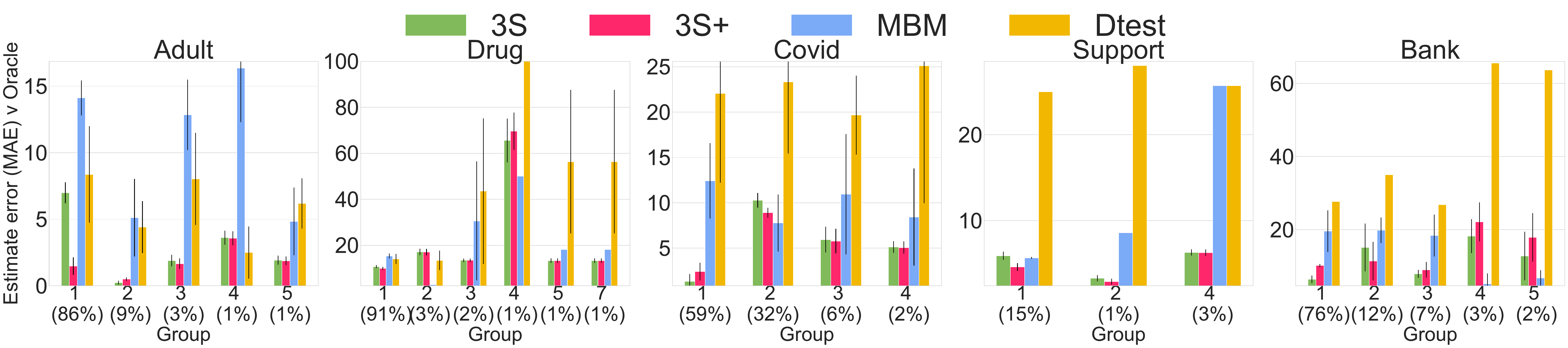}
        \caption{Decision tree}
    \end{subfigure}%
    
    \caption{Performance estimate error wrt an Oracle: F1-score }\label{more-models-f1}  

\end{figure}

\begin{figure}[H]
\centering
    \begin{subfigure}[t]{1\textwidth}
        \includegraphics[width=\textwidth]{figures/subgroups.pdf}
        \caption{Random Forest}
    \end{subfigure}%
    \quad
    \begin{subfigure}[t]{1\textwidth}
        \includegraphics[width=\textwidth]{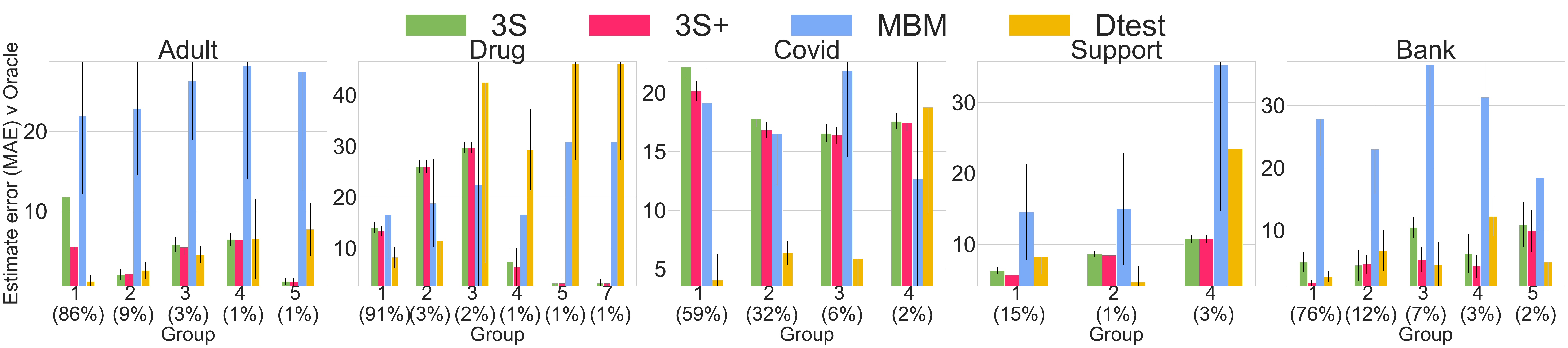}
        \caption{Gradient Boosting}
    \end{subfigure}%
    \quad
     \begin{subfigure}[t]{1\textwidth}
        \includegraphics[width=\textwidth]{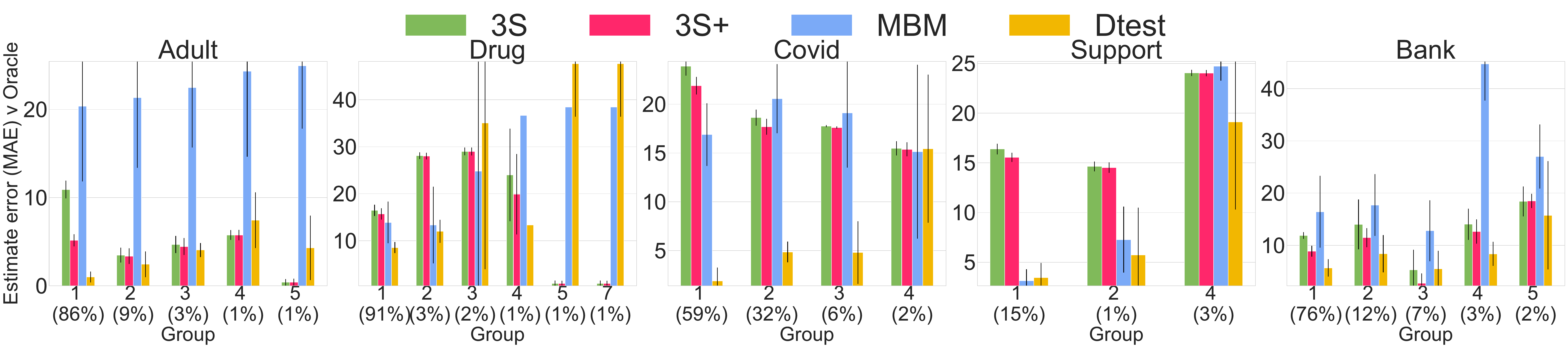}
        \caption{MLP}
    \end{subfigure}
    \quad
     \begin{subfigure}[t]{1\textwidth}
        \includegraphics[width=\textwidth]{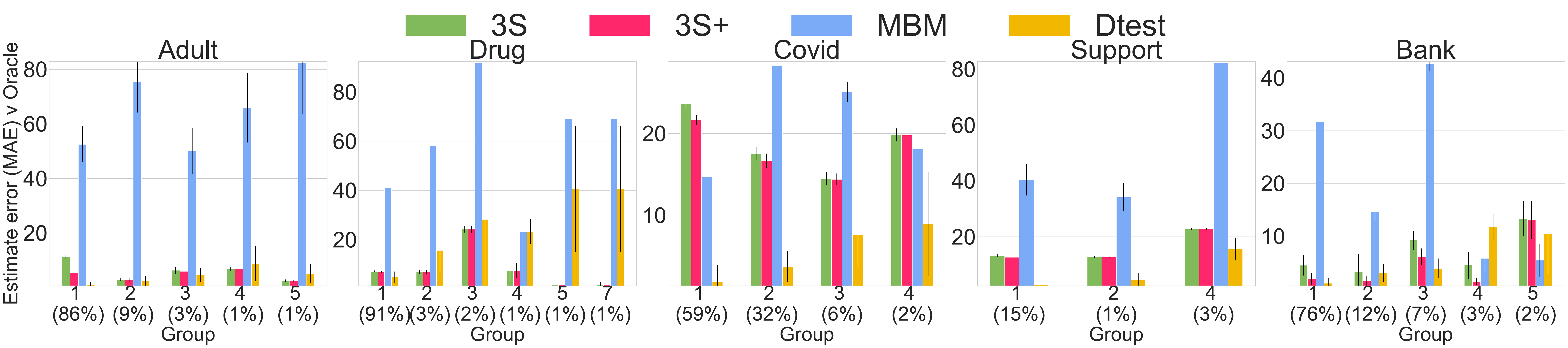}
        \caption{Adaboost}
    \end{subfigure}
    \quad
     \begin{subfigure}[t]{1\textwidth}
        \includegraphics[width=\textwidth]{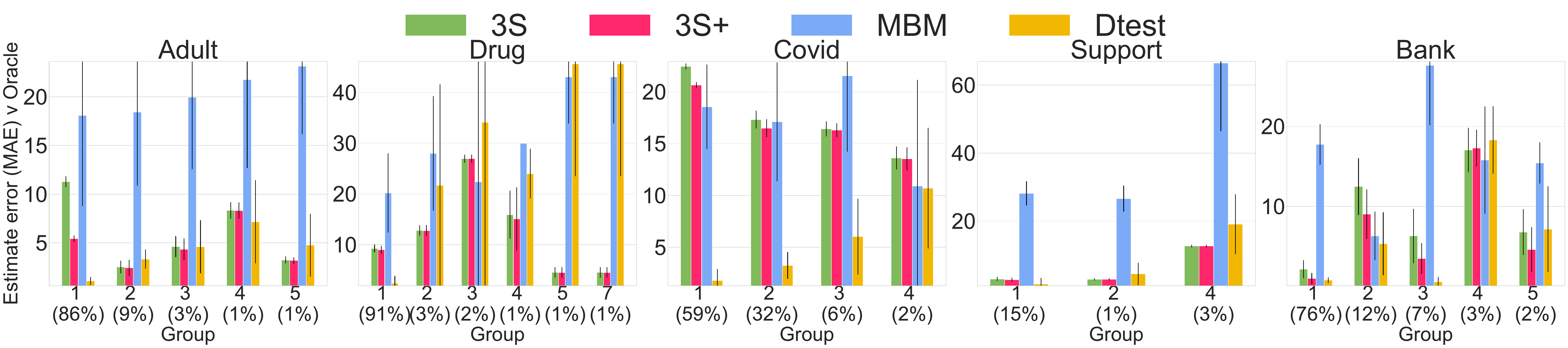}
        \caption{Logistic regression}
    \end{subfigure}
        \quad
    \quad
    \begin{subfigure}[t]{1\textwidth}
        \includegraphics[width=\textwidth]{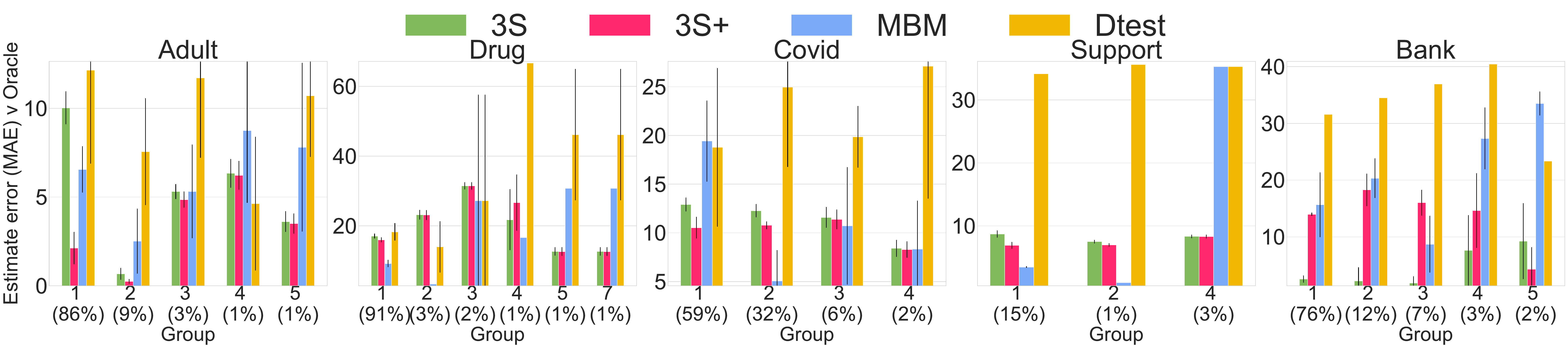}
        \caption{Decision tree}
    \end{subfigure}%
    
    \caption{Performance estimate error wrt an Oracle: Accuracy}\label{more-models-acc}  

\end{figure}

\textbf{Analysis.}
Figures \ref{more-models-f1} and \ref{more-models-acc}  illustrate that \threeS, when evaluated with more models, still better approximates the true performance on minority subgroups,
compared to test data alone. This is in terms of mean absolute performance difference between predicted performance and performance evaluated by the oracle.

\textbf{See next page for results figures}

\newpage

\subsubsection{Subgroup worst-case performance evaluation}\label{appx:worst-case-adult}

\textbf{Motivation.} When estimating sub-group performance, of course we want to have as low error as possible on average (i.e. low mean performance difference). That said, average performance glosses over the worst-case scenario. We desire that the worst-case mean performance difference is also low. This is to ensure that, by chance, the performance estimates are not wildly inaccurate. This scenario is particularly relevant, as by chance the testing data could either over- or under-estimate model performance, leading us to draw incorrect conclusions. 

\textbf{Setup.} This experiment evaluates the worst-case mean performance difference for (i) \threeS, (ii) \threeS+, and (iii) $\cD_{test,f}$. We follow the same setup as the granular subgroup experiment in Section \ref{exp-d1}.

\begin{figure}[H]
\centering
    \begin{subfigure}[t]{1\textwidth}
        \includegraphics[width=\textwidth]{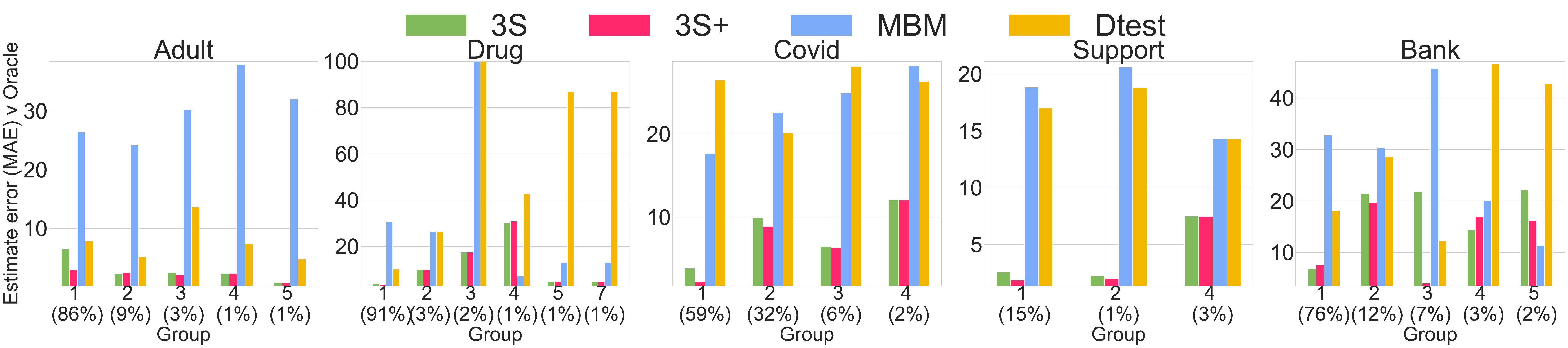}
        \caption{Random Forest}
    \end{subfigure}%
    \quad
    \begin{subfigure}[t]{1\textwidth}
        \includegraphics[width=\textwidth]{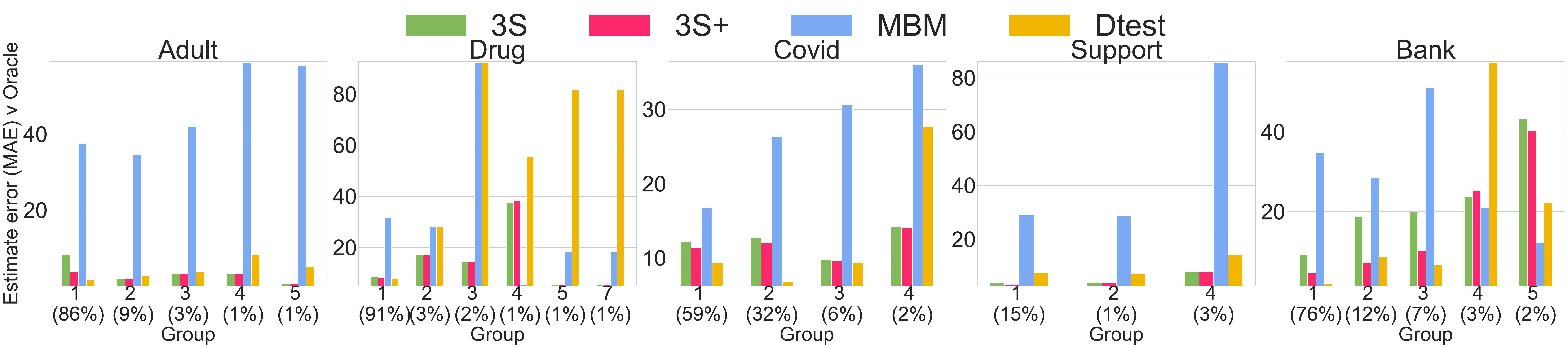}
        \caption{Gradient Boosting}
    \end{subfigure}%
    \quad
     \begin{subfigure}[t]{1\textwidth}
        \includegraphics[width=\textwidth]{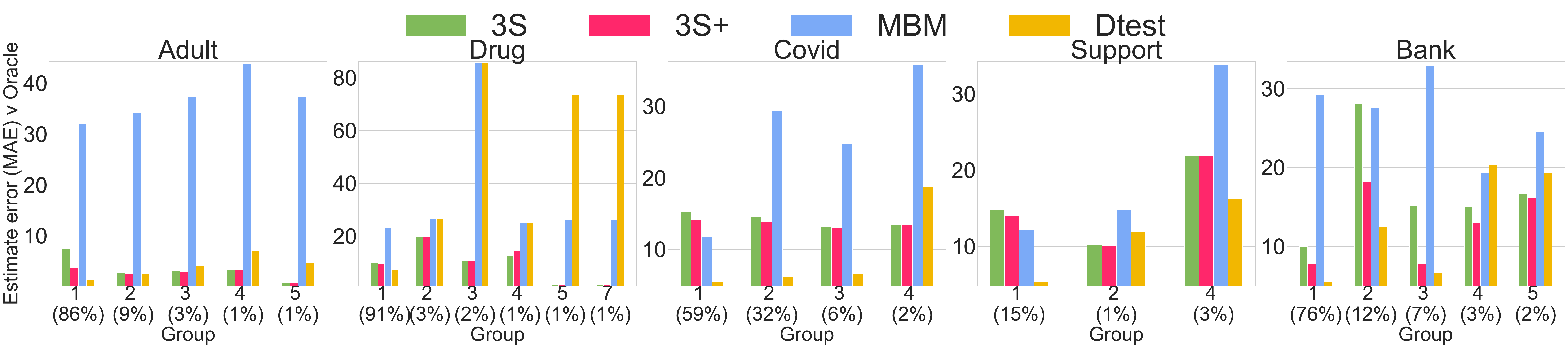}
        \caption{MLP}
    \end{subfigure}
    \quad
     \begin{subfigure}[t]{1\textwidth}
        \includegraphics[width=\textwidth]{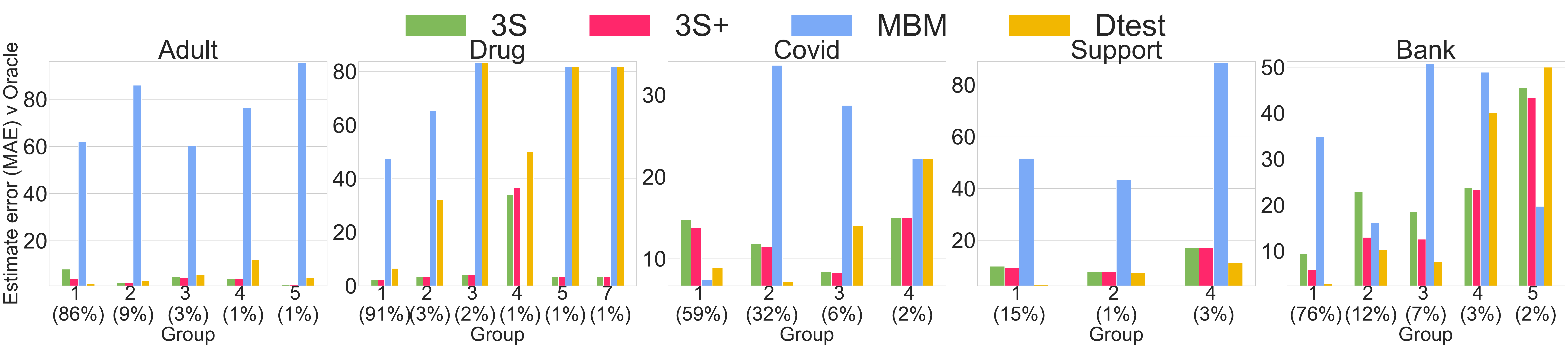}
        \caption{Adaboost}
    \end{subfigure}
    \quad
     \begin{subfigure}[t]{1\textwidth}
        \includegraphics[width=\textwidth]{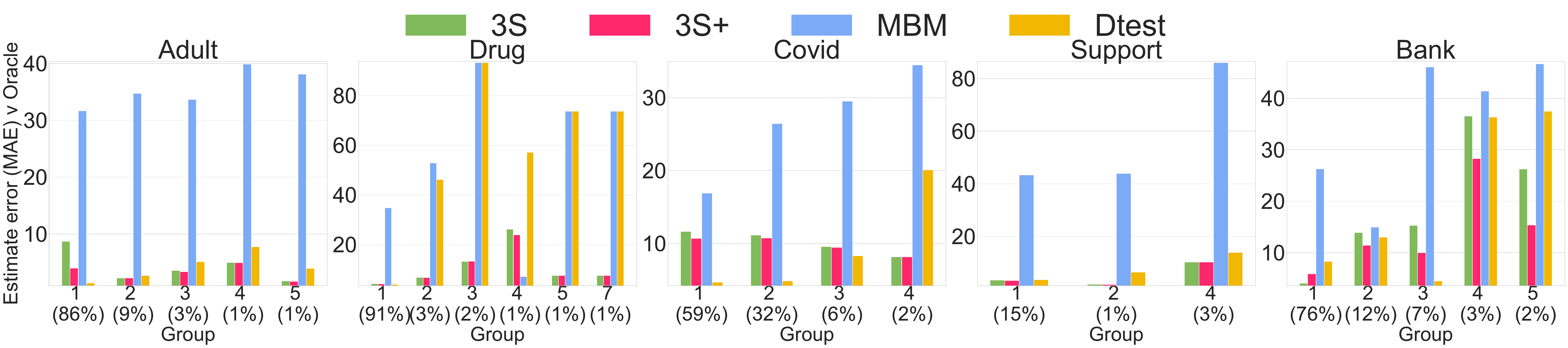}
        \caption{Logistic regression}
    \end{subfigure}
        \quad
    \quad
    \begin{subfigure}[t]{1\textwidth}
        \includegraphics[width=\textwidth]{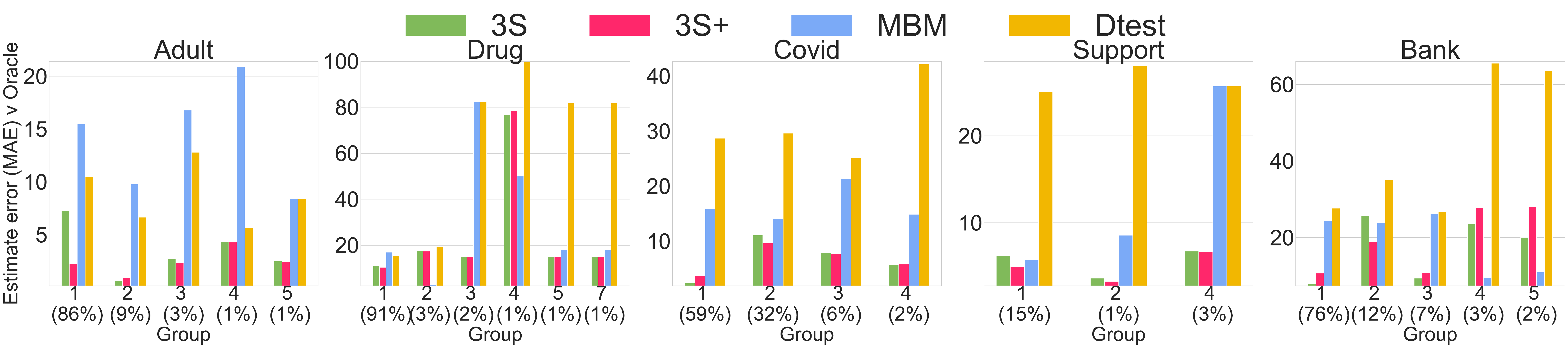}
        \caption{Decision tree}
    \end{subfigure}%
    
    \caption{Worst case, Performance estimate error wrt an Oracle: F1 Score}\label{fig:worst-case-f1}

\end{figure}

\begin{figure}[H]
\centering
    \begin{subfigure}[t]{1\textwidth}
        \includegraphics[width=\textwidth]{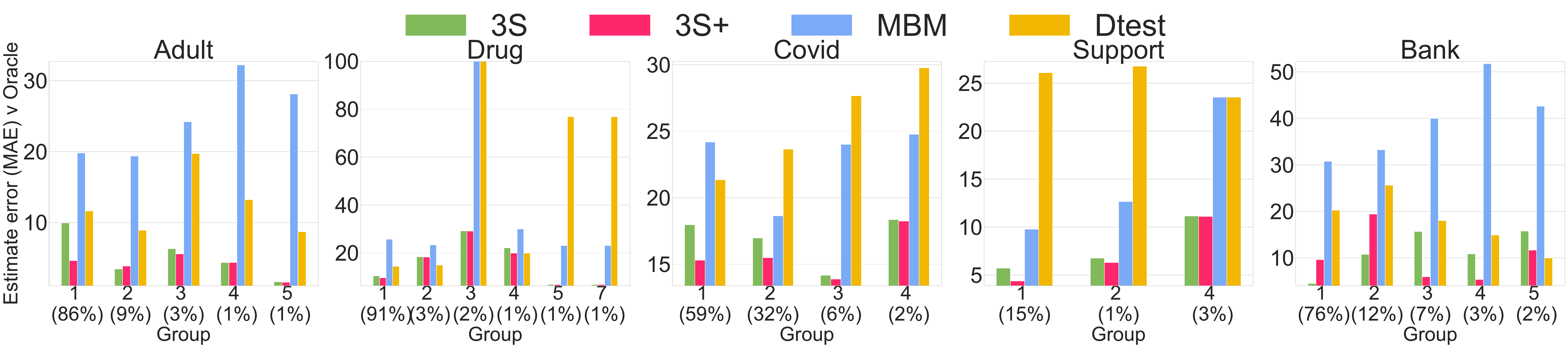}
        \caption{Random Forest}
    \end{subfigure}%
    \quad
    \begin{subfigure}[t]{1\textwidth}
        \includegraphics[width=\textwidth]{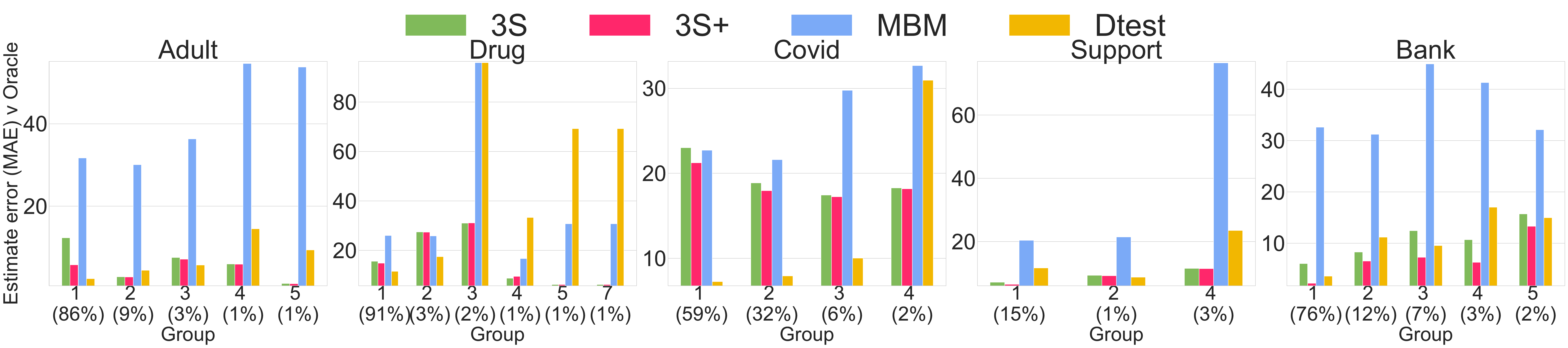}
        \caption{Gradient Boosting}
    \end{subfigure}%
    \quad
     \begin{subfigure}[t]{1\textwidth}
        \includegraphics[width=\textwidth]{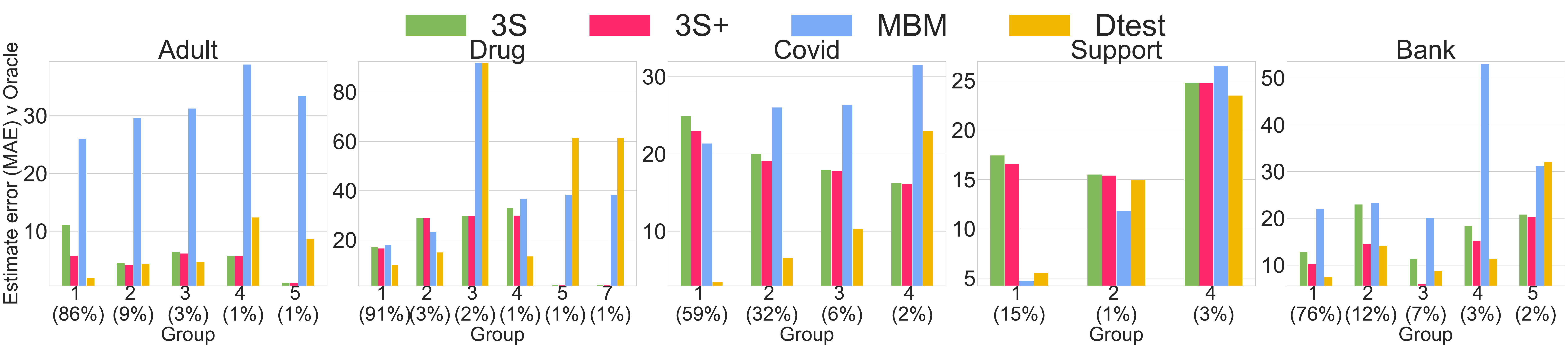}
        \caption{MLP}
    \end{subfigure}
    \quad
     \begin{subfigure}[t]{1\textwidth}
        \includegraphics[width=\textwidth]{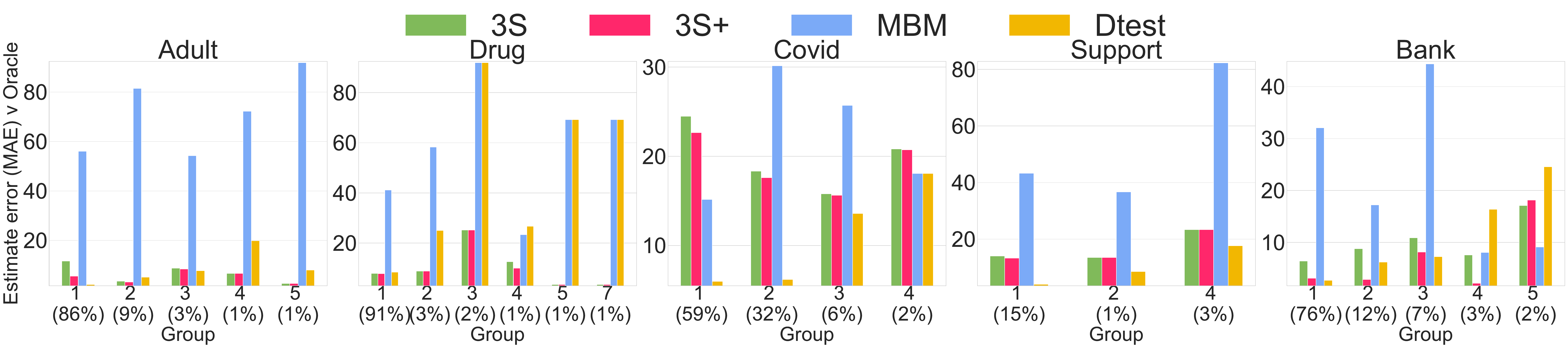}
        \caption{Adaboost}
    \end{subfigure}
    \quad
     \begin{subfigure}[t]{1\textwidth}
        \includegraphics[width=\textwidth]{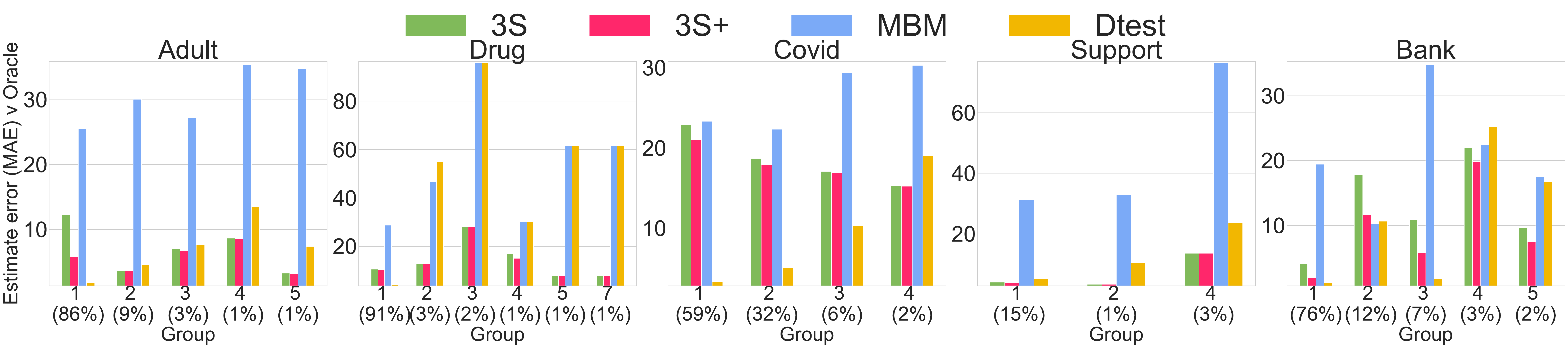}
        \caption{Logistic regression}
    \end{subfigure}
        \quad
    \quad
    \begin{subfigure}[t]{1\textwidth}
        \includegraphics[width=\textwidth]{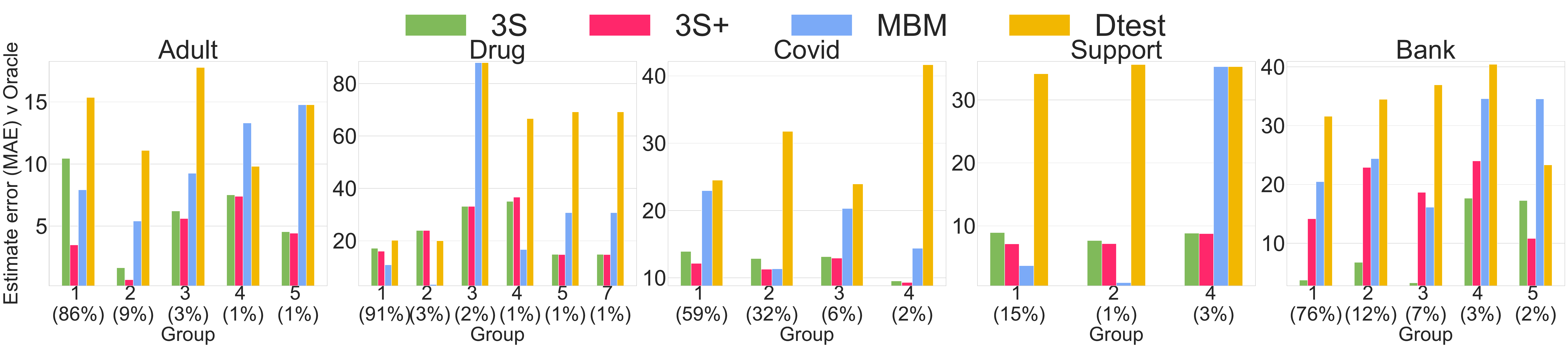}
        \caption{Decision tree}
    \end{subfigure}%
    
    \caption{Worst case, Performance estimate error wrt an Oracle: Accuracy }\label{fig:worst-case-acc}

\end{figure}

\textbf{Analysis.} Figures \ref{fig:worst-case-f1} and \ref{fig:worst-case-acc} illustrates that \threeS and the augmented \threeS+ have a lower worst-case performance compared to evaluation with real test data. This further shows that, by chance, evaluation with real data can severely over- or under-estimate performance, leading to incorrect conclusions about the model’s abilities. \threeS's lower worst-case error, means even in the worst scenario, that \threeS's estimates are still closer to true performance.

\textbf{See next page for results figures}

\newpage

\subsubsection{Fairness metrics: subgroup evaluation}

\textbf{Motivation.} We have primarily studied reliable estimation of model performance on different subgroups. This is easily generalized to estimate fairness metrics of ML models on specific subgroups. This can provide further insight into the use of synthetic data for model testing.

\textbf{Setup.} This experiment evaluates the mean performance difference for (i) \threeS, (ii) \threeS+, and (iii) $\cD_{test,f}$. We follow the same setup as the granular subgroup experiment in Section \ref{exp-d1}. We evaluate an RF model. We assess the following fairness metrics: (i) Disparate Impact (DI) ratio (demographic parity ratio) and (ii) Equalized-Odds (EO) ratio. When estimating these metrics for each subgroup (e.g. race group), we then condition on sex as the sensitive attribute. 

The DI ratio is: the ratio between the smallest and the largest group-level selection rate $E[f(X)|A=a]$ , across all values of the sensitive feature(s) $a \in A$. 

The EO ratio is the smaller of two metrics between TPR ratio (smallest and largest of $P[f(X)=1|A=a,Y=1]$ , across all values  of the sensitive feature(s)) and FPR ratio (similar but defined for $P[f(X)=1|A=a,Y=0]$), , across all values of the sensitive feature(s) $a \in A$.

\begin{figure}[H]
\centering
    \begin{subfigure}[t]{1\textwidth}
        \includegraphics[width=\textwidth]{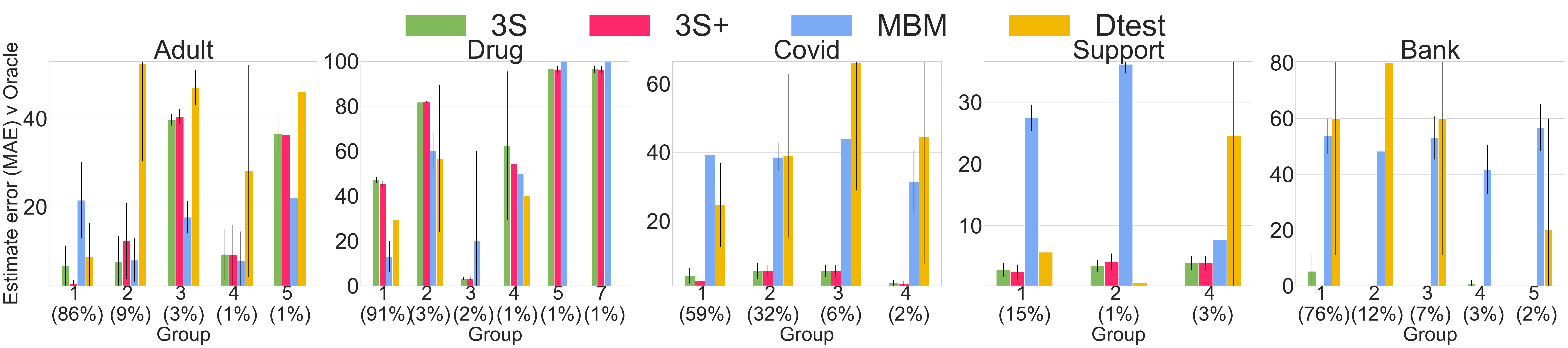}
        \caption{Random Forest}
    \end{subfigure}%
    \quad
    \begin{subfigure}[t]{1\textwidth}
        \includegraphics[width=\textwidth]{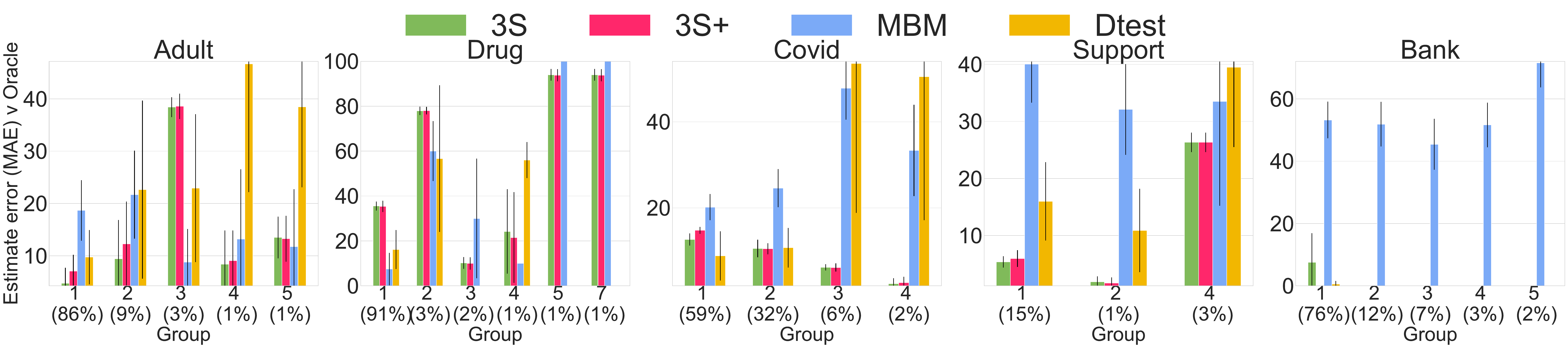}
        \caption{Gradient Boosting}
    \end{subfigure}%
    \quad
     \begin{subfigure}[t]{1\textwidth}
        \includegraphics[width=\textwidth]{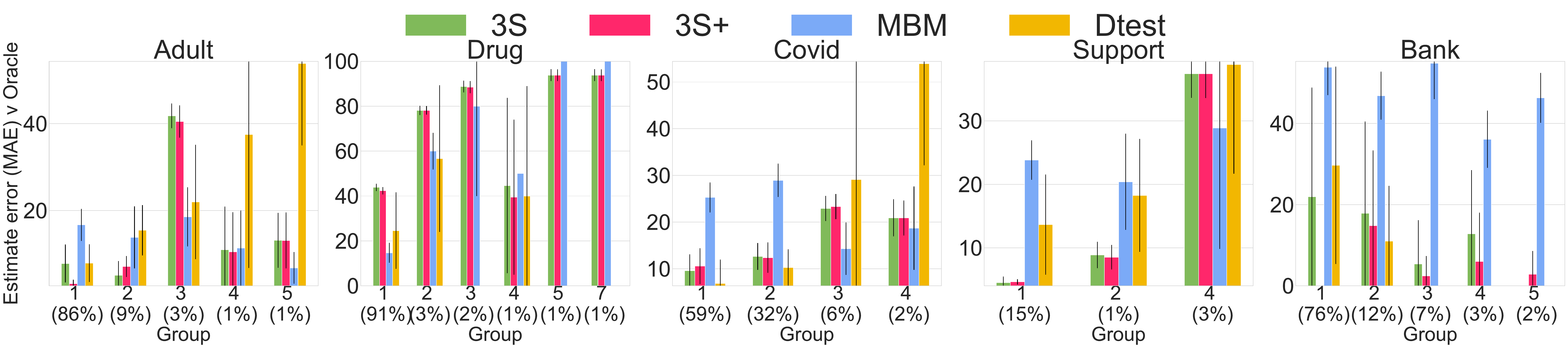}
        \caption{MLP}
    \end{subfigure}
    \quad
     \begin{subfigure}[t]{1\textwidth}
        \includegraphics[width=\textwidth]{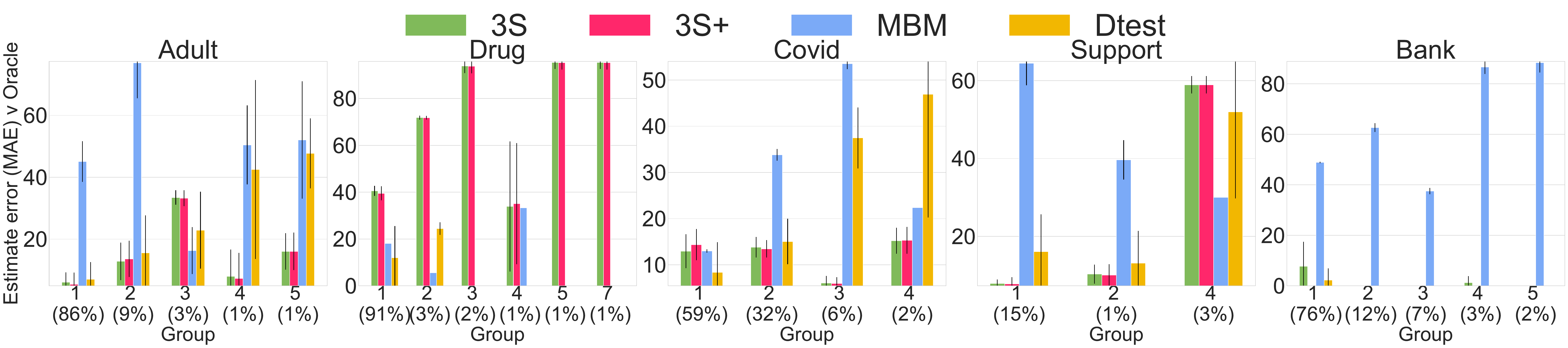}
        \caption{Adaboost}
    \end{subfigure}
    \quad
     \begin{subfigure}[t]{1\textwidth}
        \includegraphics[width=\textwidth]{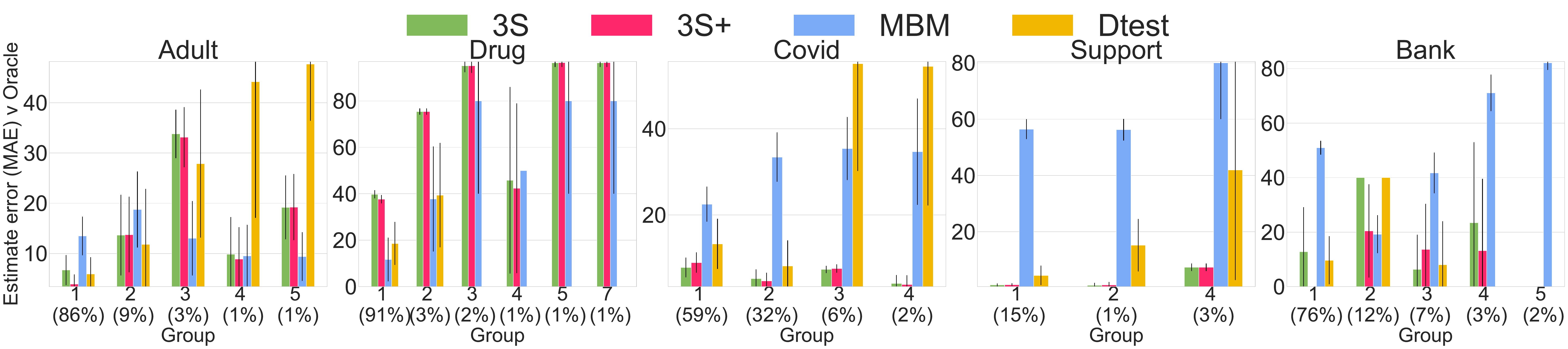}
        \caption{Logistic regression}
    \end{subfigure}
        \quad
    \quad
    \begin{subfigure}[t]{1\textwidth}
        \includegraphics[width=\textwidth]{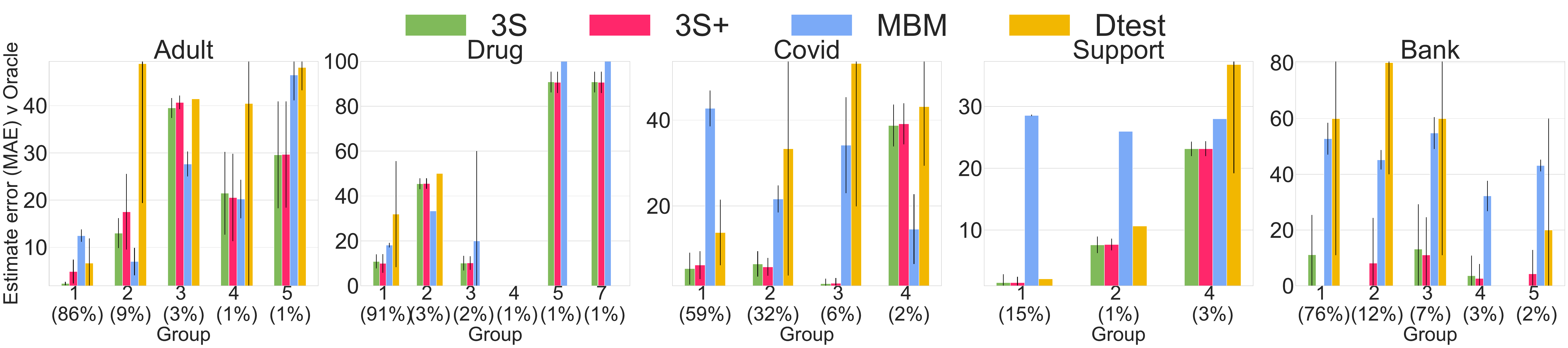}
        \caption{Decision tree}
    \end{subfigure}%
    
    \caption{Fairness (in terms of EO) estimate error}\label{fig:eo}

\end{figure}

\begin{figure}[H]
\centering
    \begin{subfigure}[t]{1\textwidth}
        \includegraphics[width=\textwidth]{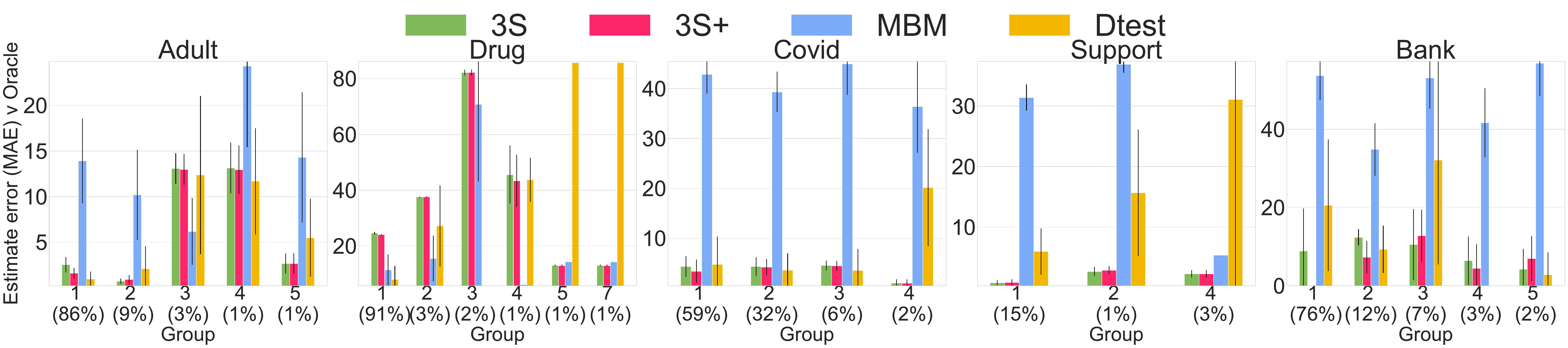}
        \caption{Random Forest}
    \end{subfigure}%
    \quad
    \begin{subfigure}[t]{1\textwidth}
        \includegraphics[width=\textwidth]{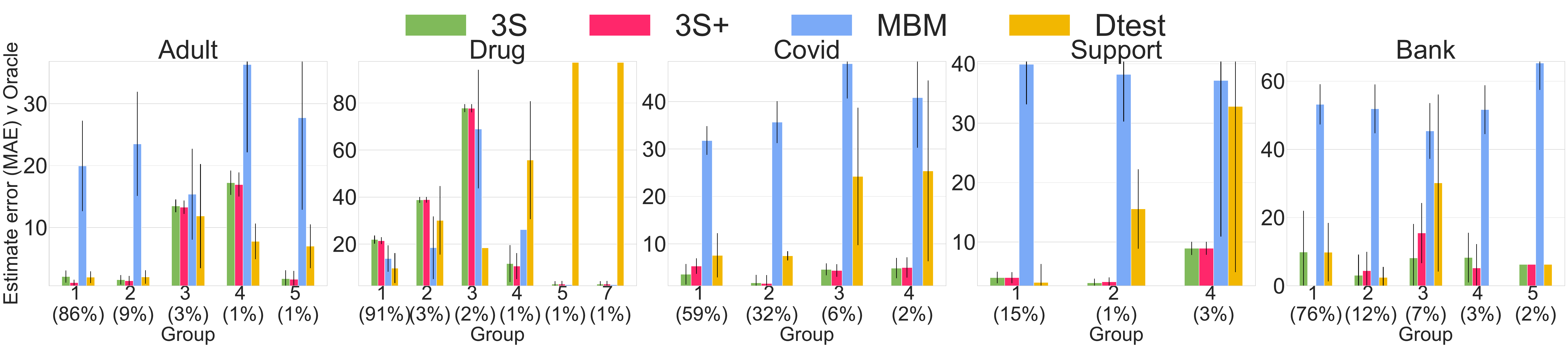}
        \caption{Gradient Boosting}
    \end{subfigure}%
    \quad
     \begin{subfigure}[t]{1\textwidth}
        \includegraphics[width=\textwidth]{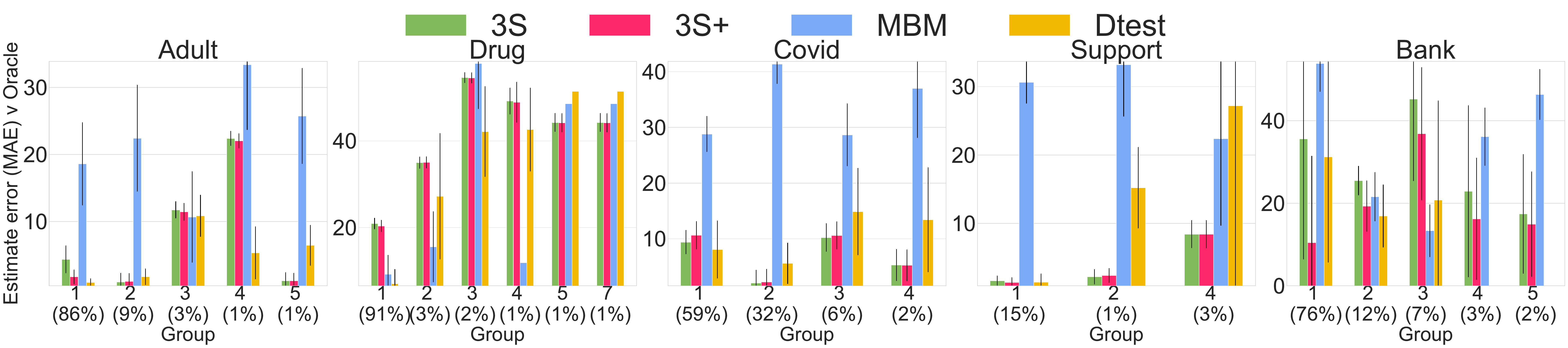}
        \caption{MLP}
    \end{subfigure}
    \quad
     \begin{subfigure}[t]{1\textwidth}
        \includegraphics[width=\textwidth]{figures/subgroup/subgroups_ada_acc.pdf}
        \caption{Adaboost}
    \end{subfigure}
    \quad
     \begin{subfigure}[t]{1\textwidth}
        \includegraphics[width=\textwidth]{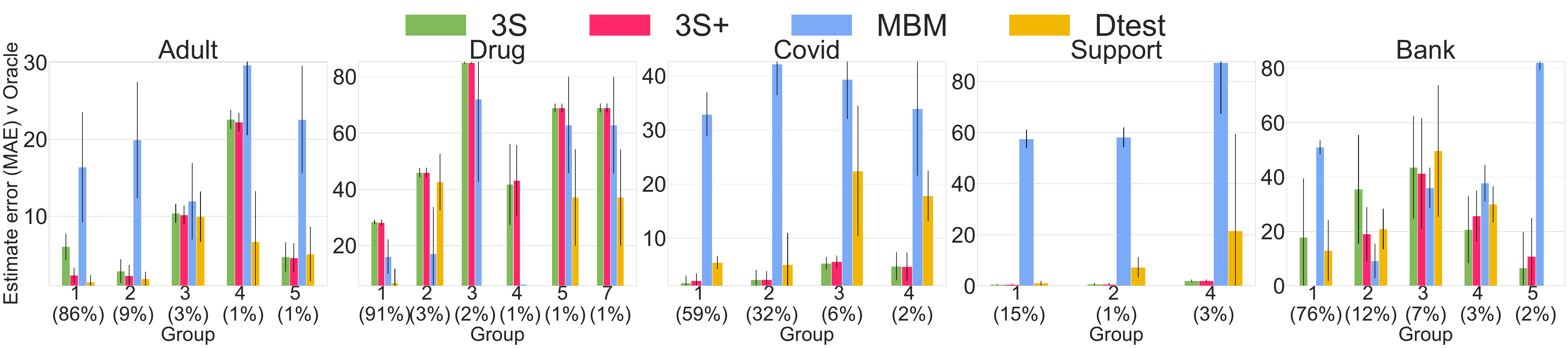}
        \caption{Logistic regression}
    \end{subfigure}
        \quad
    \quad
    \begin{subfigure}[t]{1\textwidth}
        \includegraphics[width=\textwidth]{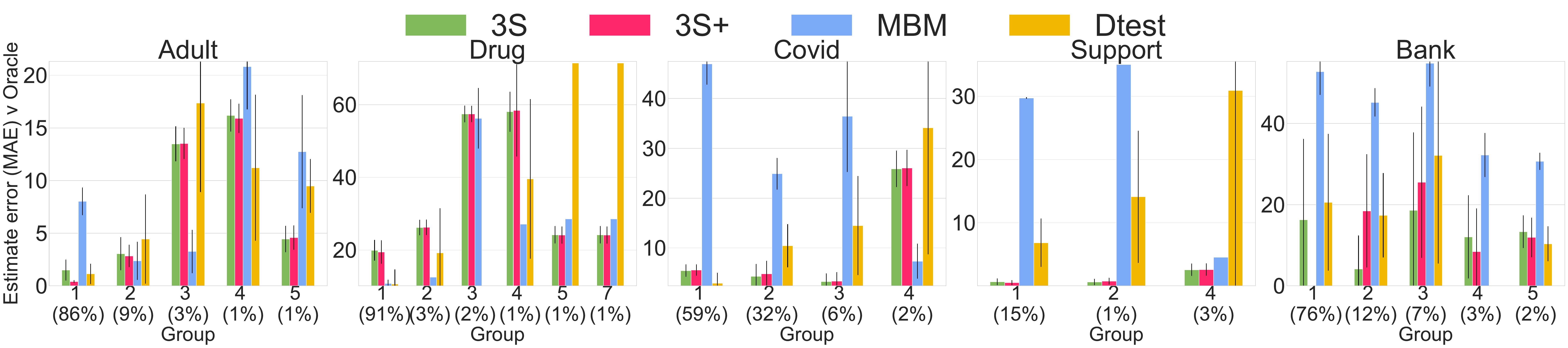}
        \caption{Decision tree}
    \end{subfigure}%
    
    \caption{DI estimate error}\label{fig:dp}

\end{figure}

\textbf{Analysis.}
Figures \ref{fig:eo} and \ref{fig:dp} illustrate that \threeS's performance on both fairness metrics, better approximates the true oracle metric on minority subgroups, compared to test data alone. This is in terms of mean absolute performance difference between predicted performance and performance evaluated by the oracle.

Note, we also assess the worst-case scenario as well, as done previously. 

\begin{figure}[H]
\centering
    \begin{subfigure}[t]{1\textwidth}
        \includegraphics[width=\textwidth]{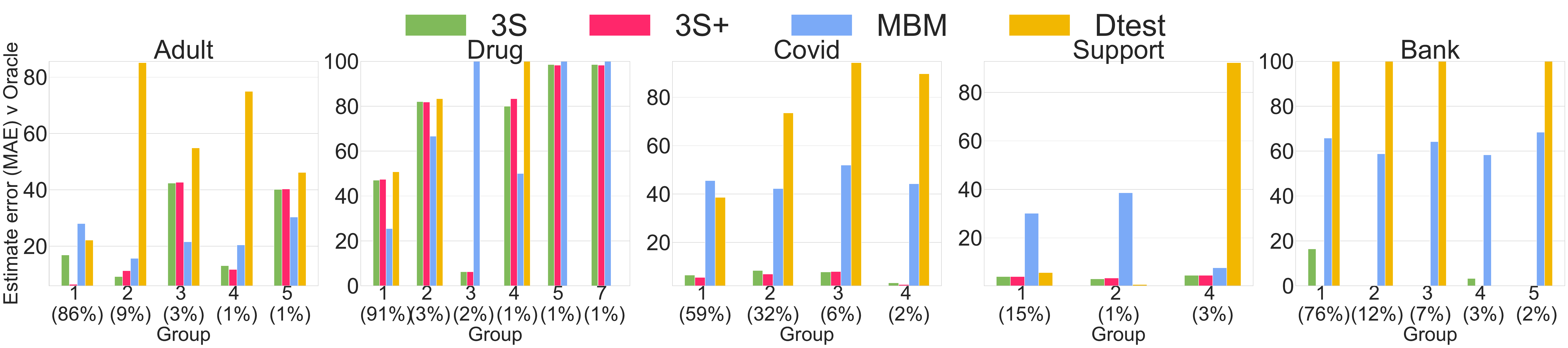}
        \caption{Random Forest}
    \end{subfigure}%
    \quad
    \begin{subfigure}[t]{1\textwidth}
        \includegraphics[width=\textwidth]{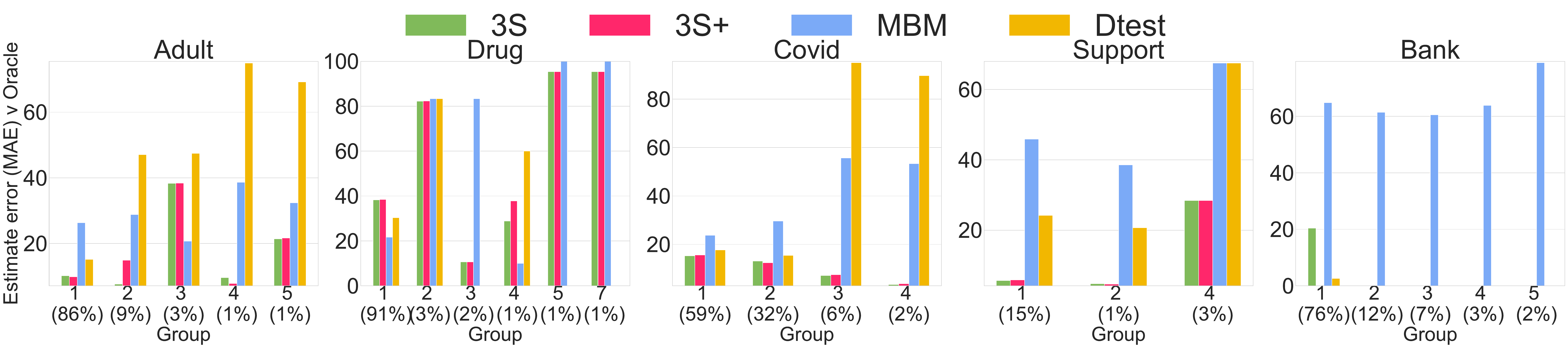}
        \caption{Gradient Boosting}
    \end{subfigure}%
    \quad
     \begin{subfigure}[t]{1\textwidth}
        \includegraphics[width=\textwidth]{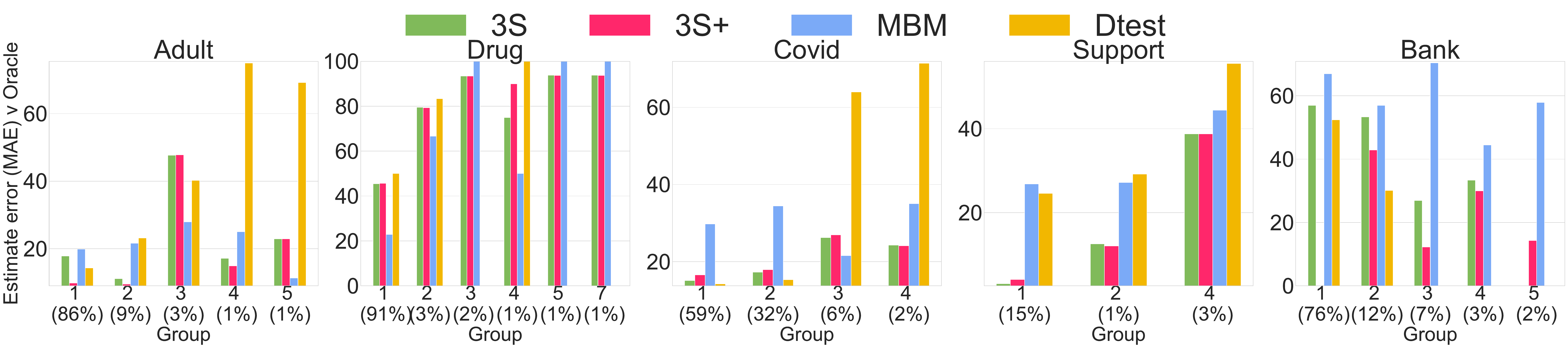}
        \caption{MLP}
    \end{subfigure}
    \quad
     \begin{subfigure}[t]{1\textwidth}
        \includegraphics[width=\textwidth]{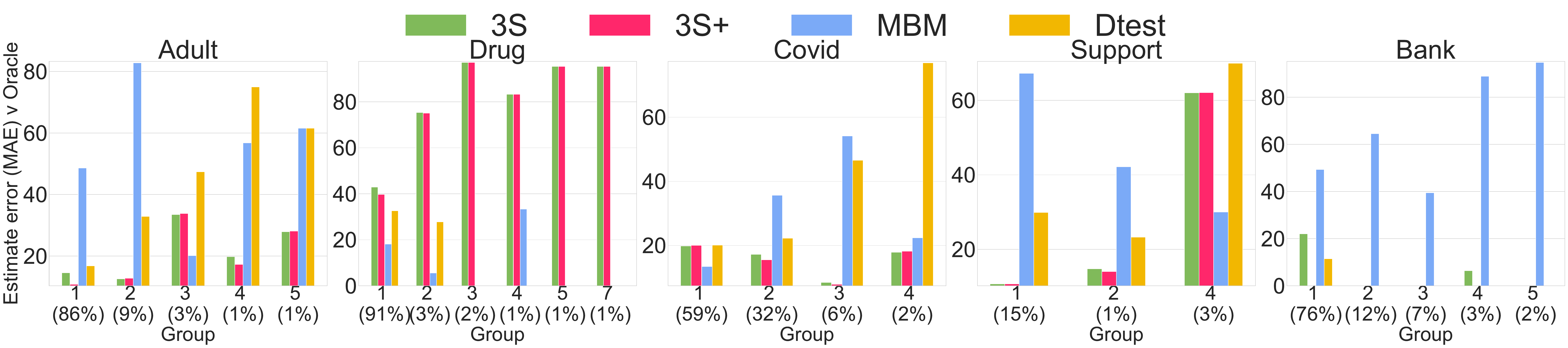}
        \caption{Adaboost}
    \end{subfigure}
    \quad
     \begin{subfigure}[t]{1\textwidth}
        \includegraphics[width=\textwidth]{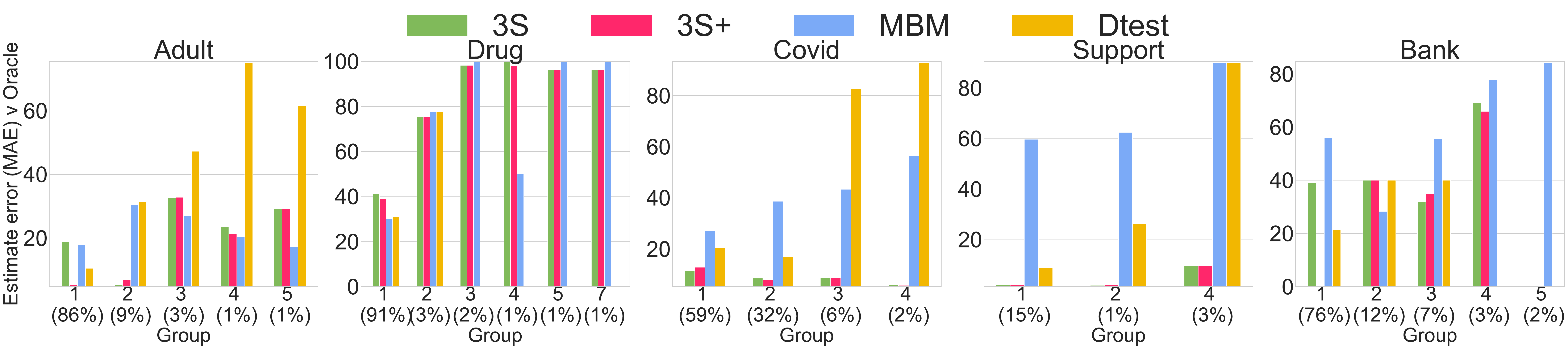}
        \caption{Logistic regression}
    \end{subfigure}
        \quad
    \quad
    \begin{subfigure}[t]{1\textwidth}
        \includegraphics[width=\textwidth]{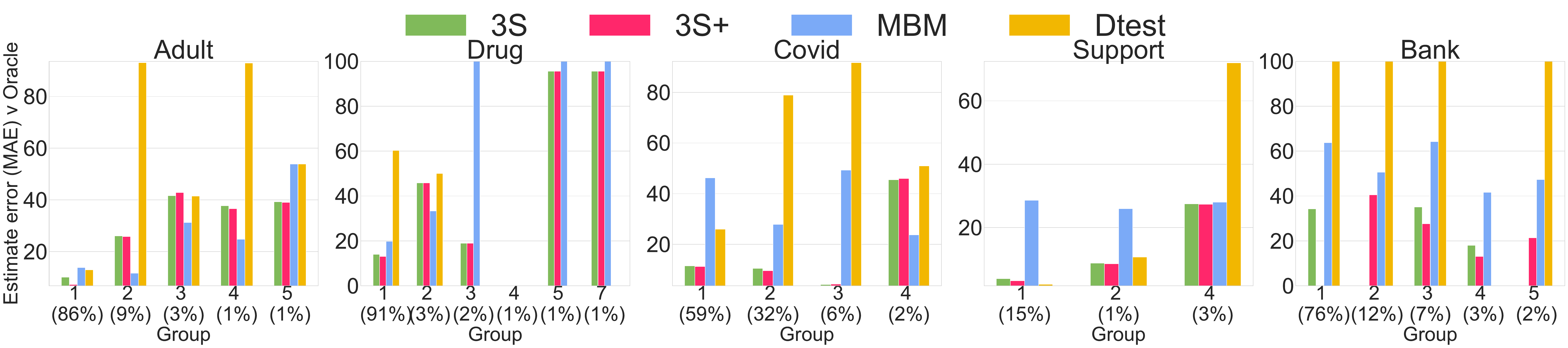}
        \caption{Decision tree}
    \end{subfigure}%
    
    \caption{Worst case, EO estimate error}\label{fig:worst-eo}

\end{figure}

\begin{figure}[H]
\centering
    \begin{subfigure}[t]{1\textwidth}
        \includegraphics[width=\textwidth]{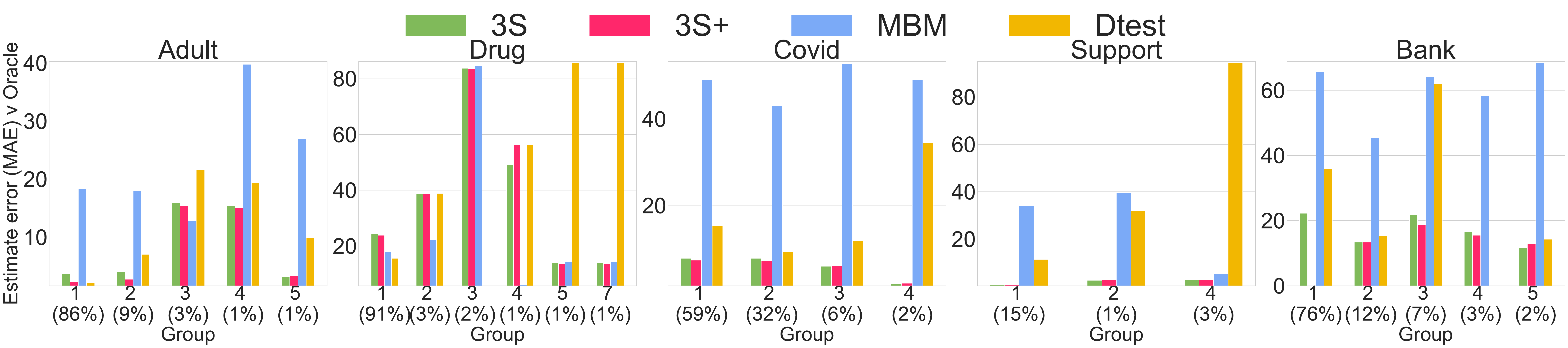}
        \caption{Random Forest}
    \end{subfigure}%
    \quad
    \begin{subfigure}[t]{1\textwidth}
        \includegraphics[width=\textwidth]{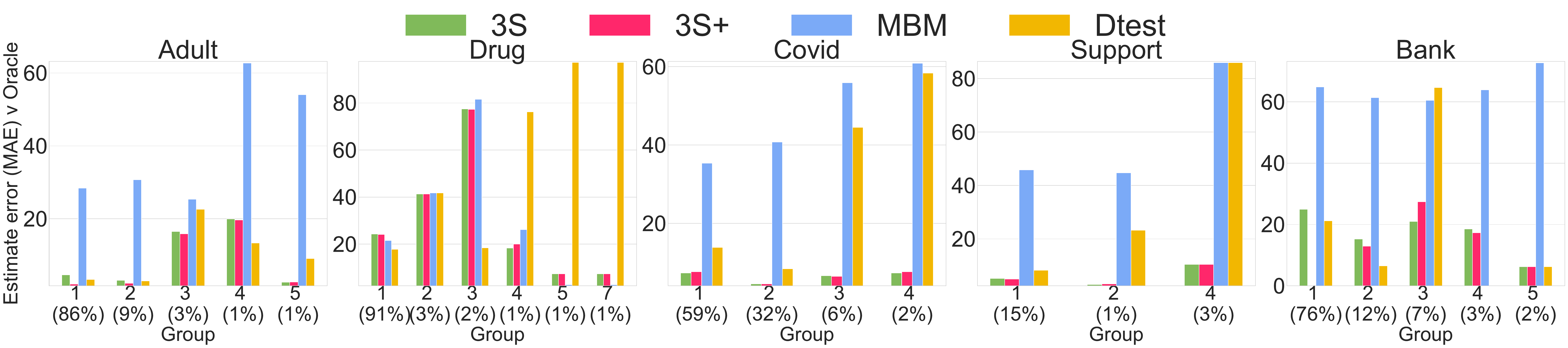}
        \caption{Gradient Boosting}
    \end{subfigure}%
    \quad
     \begin{subfigure}[t]{1\textwidth}
        \includegraphics[width=\textwidth]{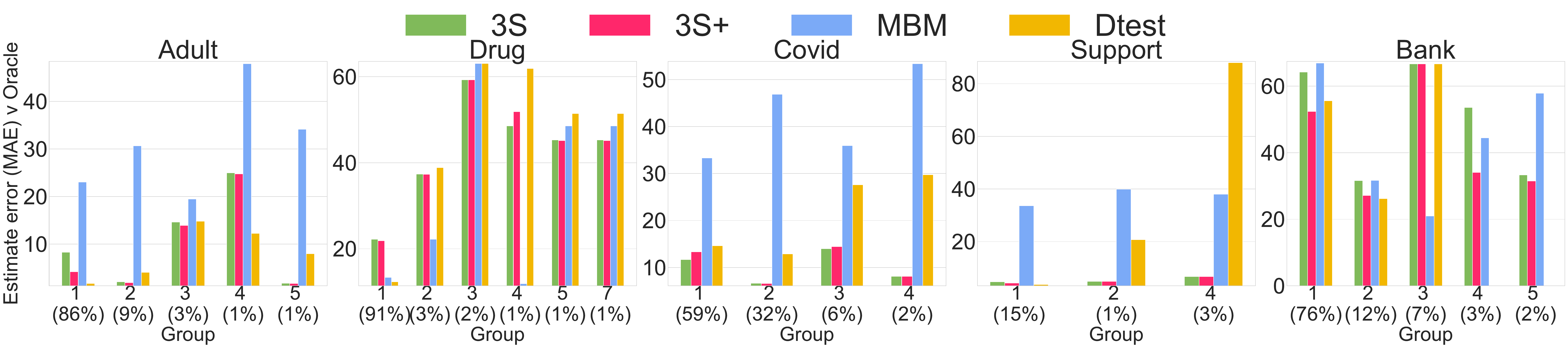}
        \caption{MLP}
    \end{subfigure}
    \quad
     \begin{subfigure}[t]{1\textwidth}
        \includegraphics[width=\textwidth]{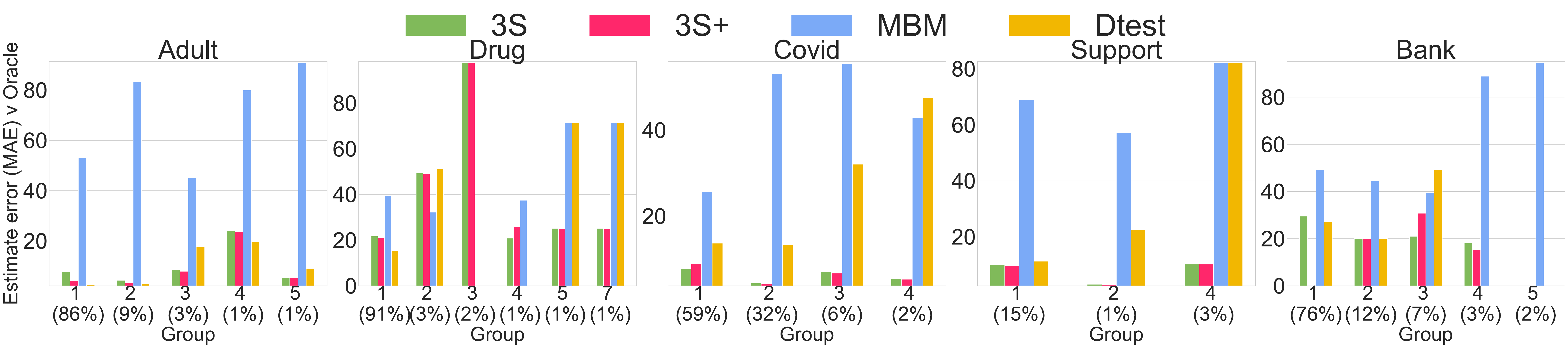}
        \caption{Adaboost}
    \end{subfigure}
    \quad
     \begin{subfigure}[t]{1\textwidth}
        \includegraphics[width=\textwidth]{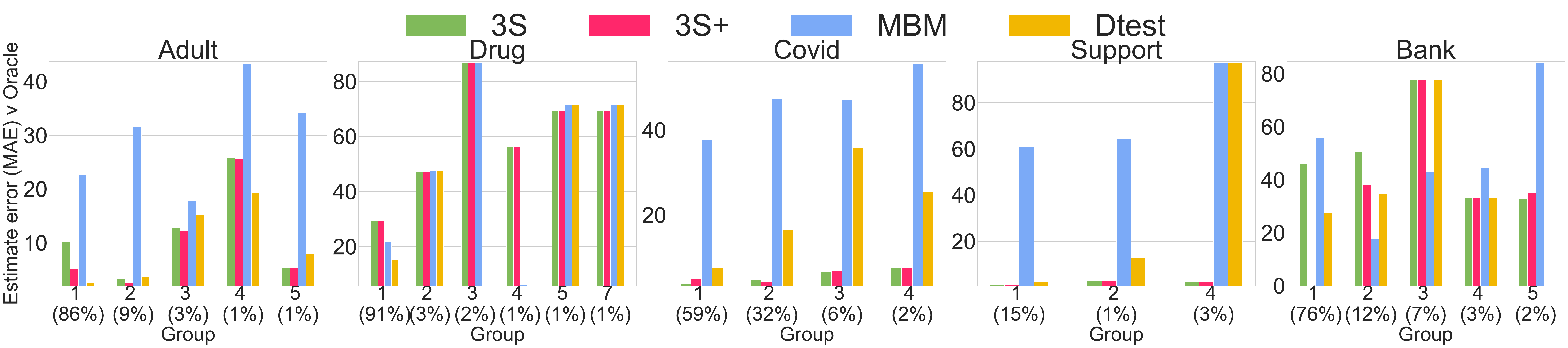}
        \caption{Logistic regression}
    \end{subfigure}
        \quad
    \quad
    \begin{subfigure}[t]{1\textwidth}
        \includegraphics[width=\textwidth]{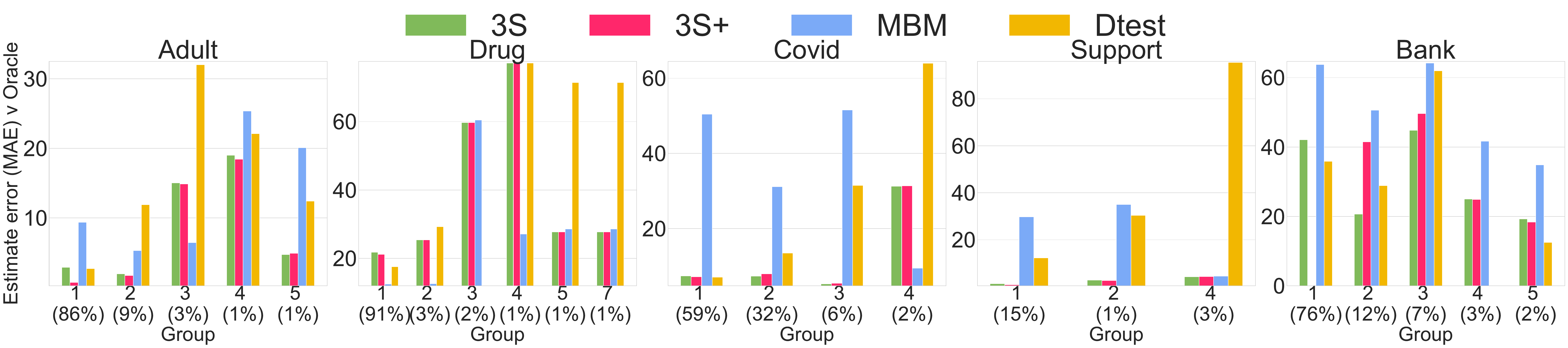}
        \caption{Decision tree}
    \end{subfigure}%
    
    \caption{Worst case, DI estimate error}\label{fig:worst-di}

\end{figure}

Figures \ref{fig:worst-eo} and \ref{fig:worst-di} illustrate that \threeS and the augmented \threeS+ have a lower worst-case estimated difference compared to evaluation with real test data. This further shows that, by chance, evaluation with real data can over- or under-estimate fairness, leading to incorrect conclusions about the model’s abilities. \threeS's lower worst-case error, means even in the worst scenario, that \threeS's estimates are still closer to the true fairness metric.

\textbf{See next page for results figures}

\newpage

\subsubsection{Intersectional subgroups: performance matrix}
\textbf{Motivation.} We perform a deep-dive of the intersectional performance matrices generated by \threeS, simply, using test data alone and the oracle.

\textbf{Analysis.} We first see in Table \ref{intersect-deepdive} that, in general, the intersectional matrix \threeS has a much \emph{lower} error when estimating performance for intersectional subgroups. i.e. this is more similar to the oracle, compared to using $\cD_{test, f}$. Note that we set the minimum number of samples required for validation $= 100$ samples. This induces sparseness, of course, but is necessary in order to prevent evaluation on too few data points. However, in cases where $\cD_{test, f}$ does not have data for the intersection (i.e. $n<100$), we do not consider these NaN blocks as part of our calculation; in fact, this makes it easier for $\cD_{test, f}$. 

The rationale is evident when evaluating the intersectional performance matrices for each group. We present the following findings.
\begin{itemize}
    \item \textbf{\threeS's insights are correct}: the underperforming subgroups, as noted by \threeS, match the oracle. Therefore, it serves as further validation.
    \item \textbf{$\cD_{test, f}$ is very sparse after cut-offs}: the 100 sample cut-off highlights the key challenge of evaluation on a test set. We may not have sufficient samples for each intersection to perform an evaluation.
\end{itemize}

\begin{table}[!h]
\centering
\caption{Adult: Intersectional performance matrix difference vs the oracle}
\begin{tabular}{l|c|c}
Model   & \threeS & $\cD_{test, f}$ \\
\hline \hline
Average &     0.13 $\pm$ 0.005
       &    0.21 $\pm$ 0.002        \\
\hline
RF      &   0.133          &  0.211  \\
GBDT      &    0.128         &  0.207 \\
MLP      &    0.128         &  0.207 \\
SVN      &     0.138        &  0.209 \\
AdaBoost      &    0.126         &  0.206 \\
Bagging      &    0.138         &  0.211 \\
LR      &    0.126         &  0.207 \\
\hline
\end{tabular}
\label{intersect-deepdive}
\end{table}

\subsubsection{Incorporating prior knowledge on shift: Covid-19}

\textbf{Motivation.} We have the ability to assess distributional shift where we have \textit{some} knowledge of the shifted distribution. Specifically, here we assume we only observe a few of the features, from the target domain. 

\textbf{Setup.} 
Our setup is similar to Section 5.2.2, however on a different dataset - i.e. Covid-19. There are known distributional differences between the north and south of Brazil. For example, different prevalence of respiratory issues, sex proportions, obesity rates, etc. Hence we train the predictive model on patients from the South (larger population) and seek to evaluate potential performance on patients from the North. We take the largest sub-regions for each. 

To validate our estimate, we use the actual northern dataset (Target) as ground-truth. Our baselines are as in Section 5.2.2.
Since the features are primarily binary, we parameterize the distributions as binomial with a probability of prevalence for their features. We can then sample from this distribution. 

\begin{figure}[H]
    \centering
    \includegraphics[width=0.5\textwidth]{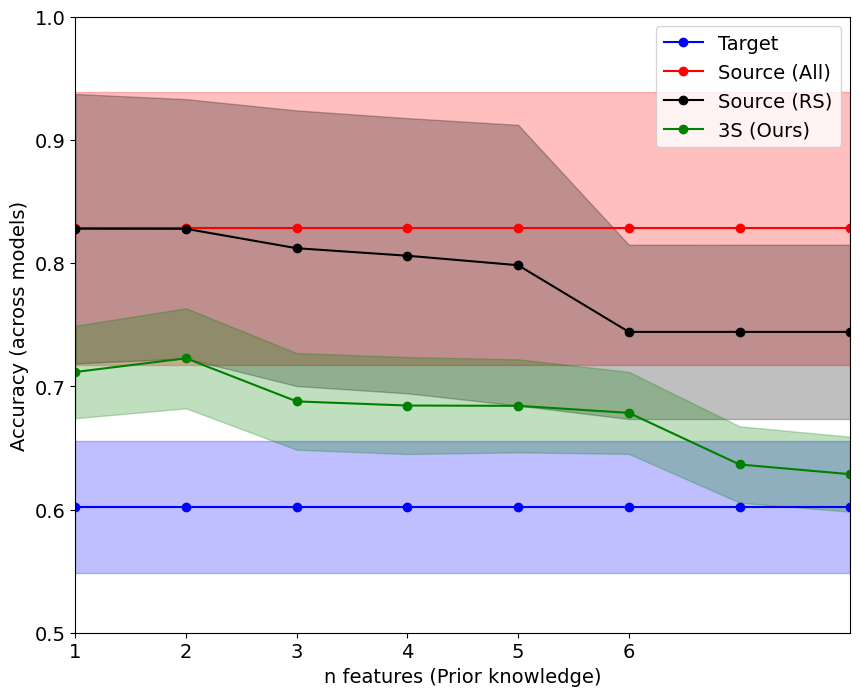}
    \caption{$\cD_{syn}$ is better able to approximate performance in the target domain compared to baselines and that performance
improves as more prior knowledge is incorporated via added features. Points are connected to highlight trends.}
    \label{fig:covid-shift}
\end{figure}

\textbf{Analysis.}
Fig. \ref{fig:covid-shift} shows the average estimated performance of $f$, as a function of the number of features observed from the target dataset. We see that the \threeS estimates are closer to the oracle across the board compared to baselines. Furthermore, for increasing numbers of features (i.e. increasing prior knowledge), we observe that \threeS estimates converge to the oracle. 

\textbf{Raw performance.} We also illustrate raw estimate errors for different downstream models in Table \ref{table: covid-raw}, which illustrates that \threeS outperforms real data alone, rejection sampling, as well as, specialized methods ATC, IM and DOC for OOD prediction with lower downstream performance estimate errors.

\begingroup
\setlength{\tabcolsep}{4pt} 
\renewcommand{\arraystretch}{1.1} 
\begin{table}[h]
\caption{\texttt{3S} has lower performance estimate error in target domain for different downstream models. Rows yellow have access to more information than \texttt{3S}, in the form of unlabeled data from the target domain. $\downarrow$ is better}
\begin{tabular}{lllllllllll}
\toprule
             & mean   & ada    & bag    & gbc     & mlp    & rf     & knn    & lr     & knn    & dt     \\ \hline\hline
3S-Testing   & \bf 0.0446 & \bf 0.0112 & \bf 0.0962 &   \bf  0.0205 &  0.1777 & \bf 0.0278 & \bf 0.0100 & \bf 0.0105 & \bf 0.0100 & \bf 0.0100 \\ \hline
All (Source) & 0.2262 & 0.1029 & 0.3890 & 0.0929  & 0.2276 & 0.3732 & 0.1774 & 0.0521 & 0.1774 & 0.3942 \\ \hline
RS (Source)  & 0.1421 & 0.1842 & 0.1781 & 0.1303  & 0.2954 & 0.1613 & 0.0497 & 0.1138 & 0.0497 &  0.0244 \\ \hline
\rowcolor[HTML]{FFFBC8} 
ATC \cite{gargleveraging}          & 0.1898 & 0.0236 & 0.2948 & 0.0542 & \bf 0.0331 & 0.2264 & 0.3985 & 0.0638 & 0.3985 & 0.4198 \\ \hline
\rowcolor[HTML]{FFFBC8} 
IM \cite{chen2021mandoline}      & 0.1782 & 0.1030 & 0.3638 & 0.0330  & 0.1563 & 0.2828 & 0.0555 & 0.0140 & 0.0555 & 0.4170 \\ \hline
\rowcolor[HTML]{FFFBC8} 
DOC \cite{guillory2021predicting}    & 0.1634 & 0.1019 & 0.2646 & 0.0304  & 0.1557 & 0.2215 & 0.1186 & 0.0010 & 0.1186 & 0.4060 \\
\hline
\end{tabular}
\label{table: covid-raw}
\end{table}
\endgroup

\subsubsection{Sensitivity to shifts on additional features}

\textbf{Motivation.} In the main paper, we have characterized the sensitivity across operating ranges for two features in two datasets (Adult and SEER). Ideally, a practitioner would like to understand the sensitivity to all features in the dataset. We now conduct this assessment on the Adult dataset, producing model sensitivity curves for all features.

\textbf{Analysis.} We include the model sensitivity curves for all features as part of the example model report in Appendix \ref{appx:modelreport}, see Figures \ref{fig:modelreport-sensitivity}, (a)-(t).

\newpage
\subsection{Other definitions of subgroups}
We illustrate how the definition of subgroup allows two other interesting use cases for granular evaluation: (i) subgroups defined using a point of interest (i.e. how would a model perform on patients that look like X), and (ii) subgroups defined in terms of local density (i.e. how would a model perform on unlikely patients). 
\label{appx:other_definitions_subgroups}

\subsubsection{Subgroups relating to points of interest}
\label{sec:method_poi}
\textbf{Motivation.} End-users may also be interested in knowing how a model performs on some point of interest $x^*$. For example, a clinician may have access to a number of models and may need to decide for the specific patient in front of them what the model's potential predictive performance might be for the specific patient. This is relevant because global performance metrics may hide that models underperform for specific samples.

We can often assess model performance on samples similar to the sample of interest, i.e. ``neighborhood performance''. A challenge in reality is that we often only have access to a held-out test dataset  (i.e. $\cD_{test}$), rather than the entire population (i.e. $\cD_{oracle}$). How then can we quantify performance? 

Usually, it is not possible to assess local performance using test data alone, since there may be very few samples that are similar enough---with ``similar'' defined in terms of some distance metric, i.e. $\mathcal{S} = \{x\in\mathcal{X}|d(x,x^*)<\epsilon\}$, with distance metric $d$ and some small distance $\epsilon\geq 0$. As before, we can instead generate synthetic data in the region $\mathcal{S}$ and compute the performance on this set instead. This reduces the dependence on the small number of samples, in turn reducing variance in the estimated performance.

\textbf{Set-up.} We compare two approaches: (i) find nearest-neighbor points in $\cD_{test}$)---which might suffer from limited similar samples---, or (ii) use \threeS and generate synthetic samples $\cD_{syn}$ in some neighborhood of $x^*$. Similarly as before, we assess these two methods by comparing estimates with a pseudo-ground truth that uses nearest neighbors on a much larger hold-out set, $\cD_{oracle}$. Again, we compare (1) \emph{mean absolute performance difference} and (2) \emph{worst-case performance difference} between a specific evaluation set and the oracle dataset.  We average across 10 randomly queried points $x^*$ and use $k=10$ nearest neighbors.

\textbf{Analysis.} The results in Table \ref{sample-interest} show that \threeS has a much lower neighborhood performance gap, both average and worst case for $x^*$ across models, when compared to the assessment using $\cD_{test}$. The rationale is that by using synthetic data, we can generate more examples $\cD_{syn}$ that closely resemble $x^*$, whereas with $\cD_{test}$ we might be limited in the similar samples that can be queried - hence resulting in the higher variance estimates and poorer overall performance.

\begin{table*}[!h]
\vspace{-0cm}
\centering
\caption{\footnotesize{Comparing two types of query methods to evaluate performance on points of interest $x^*$, which illustrates that \threeS closer approximates an oracle both on average and worst case.}}
\vspace{-0mm}
\scalebox{0.9}{
\begin{tabular}{c|cc|cc}
\multicolumn{1}{l}{Model} & \multicolumn{2}{|c|}{Mean performance difference $\downarrow$} & \multicolumn{2}{|c}{Worst-case performance difference $\downarrow$} \\ \hline
\multicolumn{1}{c|}{}     & $\cD_{syn}$ (\threeS)                      & \multicolumn{1}{c|}{$\cD_{test, f}$}       & $\cD_{syn}$ (\threeS)                         & \multicolumn{1}{c}{$\cD_{test, f}$}             \\ \hline
\multicolumn{1}{c|}{MLP}  & \bf 0.083     & \multicolumn{1}{c|}{0.15}    & \bf 0.482         & \multicolumn{1}{c}{0.60}       \\
\multicolumn{1}{c|}{RF}   & \bf 0.086     & \multicolumn{1}{c|}{0.18}    & \bf 0.256         & \multicolumn{1}{c}{0.50}           \\
\multicolumn{1}{c|}{GBDT}  & \bf 0.093     & \multicolumn{1}{c|}{0.18}    & 0.50                             & \multicolumn{1}{c}{0.50}       \\ \hline
\end{tabular}
}
\label{sample-interest}
\vspace{-0mm}
\end{table*}

\textbf{\textcolor{BrickRed}{Takeaway.}}
\threeS's synthetic data can more robustly estimate performance on individual points of interest based on the samples generated in the neighborhood of $x^*$.

\subsubsection{Subgroups as high- and low-density regions}
\label{sec:method_likely_unlikely}
\textbf{Motivation} Models often perform worse on outliers and low-density regions due to the scarcity of data during training. We generate insight into this by defining subgroups in terms of density.

\textbf{Methodology.} 

We would like to partition the points into sets of most likely to least likely w.r.t. density. We use the notion of $\alpha$-support from \citep{alaa2022faithful}, namely:
\begin{equation}
    \Supp^\alpha(p) = \underset{S\subseteq \tilde{\mathcal{X}}}{\arg\min}\ V(S)\ s.t.\  p(\tilde{X}\in S) = \alpha,
\end{equation}
with $V$ some volume measure (e.g. Lebesgue). In other words, $\alpha$-support $\Supp^\alpha(p)$ denotes the smallest possible space to contain $\tilde{X}$ with probability $\alpha$---which can be interpreted as the $\alpha$ most likely points. Subsequently, we can take a sequence of quantiles, $(q_i)_{i=0}^k\in [0,1],$ with $q_i = \frac{i}{k}$ and look at the sequence of support sets, $(\Supp^{q_i}(p))_{i=0}^k$.

The $\alpha$-support itself always contains the regions with the highest density. To actually partition the points into sets from likely to unlikely, we instead look at the difference sets. That is, let us define sets $S_i = \Supp^{q_i}(p)\backslash \Supp^{q_{i-1}}(p)$ for $i=1,..., k$.

In fact, we do not know $p$ exactly and even if we did, it is usually intractable to find an exact expression for the $\alpha-$support. Instead, we compute the $\alpha$-support in the generative model's latent space, and not the original space. The latent distribution is usually chosen as a $d_z$-dimensional standard Gaussian, which has two advantages: (i) the distribution is continuous---cf. the original space, in which there may exist a lower-dimensional manifold on which all data falls; and (ii) the $\alpha$-support is a simple sphere, with the radius given by $\text{CDF}^{-1}_{\chi^2_{d_z}}(\alpha)$.

\begin{wrapfigure}{r}{0.33\textwidth}
\vspace{-0mm}
\captionsetup{font=footnotesize}
  \centering
    \includegraphics[width=0.25\textwidth]{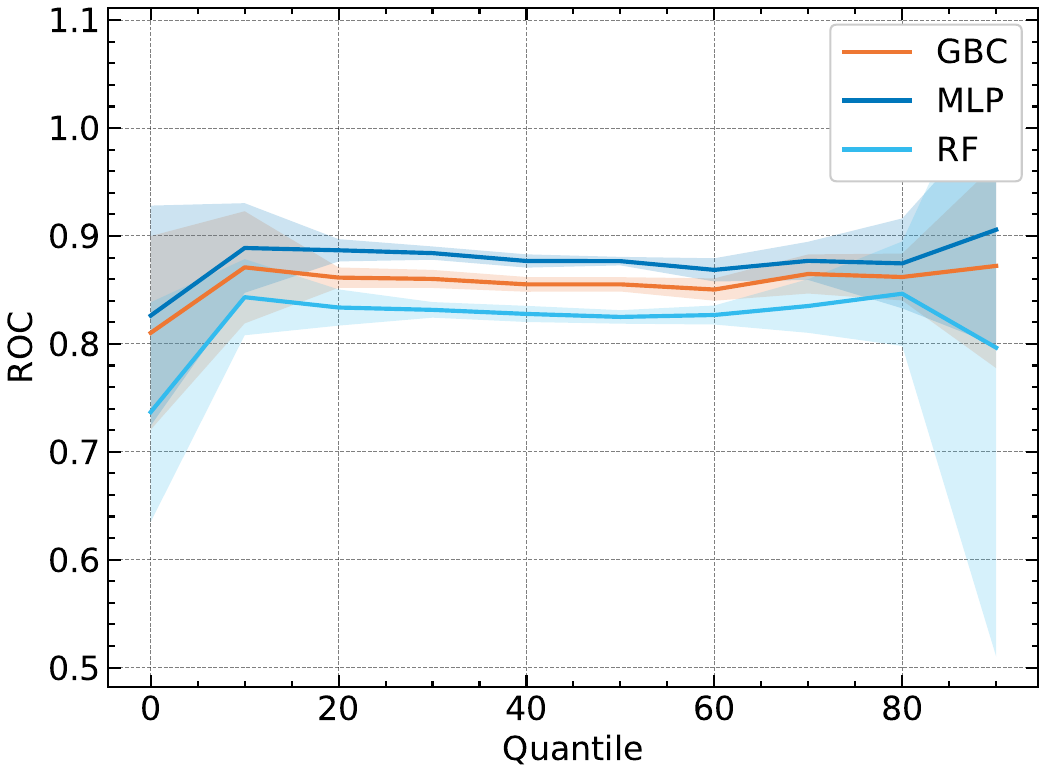}
   \vspace{-0mm}
   \caption{\footnotesize{Performance on likely vs unlikely data, where error bars are over 10 runs}}
    \vspace{-0mm}
    \rule{\linewidth}{.25pt}
   \vspace{-0mm}
    \label{fig:likely}
\end{wrapfigure}

\textbf{Set-up.}  We train a generative model $G:\mathcal{Z}\rightarrow \tilde{\mathcal{X}}$ as before, where the input $\mathbf{z}$ is $d_z$-dimensional Gaussian noise. During generation we save inputs $\{z_j\}_{j=1}^{k N}$ and generate corresponding data $\mathcal{D}_{syn}=\{\tilde{x}_j\}_{j=1}^{kN}$. Sets for likely and unlikely points are defined using $\alpha$-support in latent space $\mathcal{Z}$, for quantiles $\mathbf{q}=(q_i)_{i=1}^k$---see Section \ref{sec:method_likely_unlikely}. Specifically, let the index sets for each quantile set be given by $I_i = \{j| ||z_j||_2 < \text{Quantile}(\{||z_j||_2\}, q_i)\}$, giving synthetic sets $\mathcal{D}_{syn}^{q_i} = \{\tilde{x}_j\in\mathcal{D}_{syn}| j\in I_i\}$ for all $i$. We then evaluate the predictive model $f$ on each $\mathcal{D}_{syn}^{q_i}$ and plot results w.r.t. $\mathbf{q}$. We use the SEER dataset.

\textbf{Analysis.} Fig. \ref{fig:likely} shows that for the tail quantiles (unlikely) data that the model performance is indeed highly variable and prone to possible poor performance. This is in contrast to the likely data, which has much more stable performance. 

\textbf{\textcolor{BrickRed}{Takeaway.}}
\threeS's synthetic data can be used to quantify model performance on unlikely and likely data.

\subsection{Improving $G$ when training data of $f$ is available}\label{appx:subgroup_dtrain}

\textbf{Motivation.} In previous experiments, we have assumed that we have access to only $\cD_{test,f}$ for training $G$. In some scenarios, we have access to $\cD_{train,f}$. For example, the model developer has access to $\cD_{train,f}$. Since the training dataset is often larger than the testing dataset, we could use this data to train $G$. Evidently, this results in some bias: we are now using synthetic data generated using $G$, which is trained on $\cD_{train,f}$, to evaluate a predictive model that is also trained on $\cD_{train,f}$. However, here we show that this bias is outweighed by the improved quality of $G$.

\textbf{Setup.} This experiment evaluates the mean performance difference for (i) \threeS, (ii) \threeS+, and (iii) $\cD_{test,f}$. We follow the same setup as the granular subgroup experiment in Section \ref{exp-d1}, with the only difference being that $\cD_{train,G}=\cD_{train,f}$. We assess the utility of this setup for granular subgroup evaluation.

\textbf{Analysis.} Table \ref{table:dtrain} illustrates the performance similar to that of the main paper. We find that \threeS mostly provides a more accurate evaluation of model performance (i.e. with estimates closer to the oracle)
compared to a conventional hold-out dataset. This is especially true for the smaller subgroups for which synthetic data is indeed necessary. Furthermore, \threeS has a lower worst-case error compared to $\cD_{test,f}$. The results with \threeS+ also illustrate the benefit of augmenting real data with synthetic data, leading to lower performance differences.

\begin{table*}[!t]
\centering
\vspace{-0mm}
\caption{Mean absolute performance difference
between predicted performance and performance
evaluated by the oracle, where $G$ is trained on $\cD_{train,f}$. \threeS better approximates true performance on minority subgroups,
compared to test data alone.}
\vspace{-0cm}
\scalebox{0.75}{

\begin{tabular}{c|c|ccc|ccc}
Model                 & Subgroup (\%)                      & \multicolumn{3}{|c|}{Mean performance diff. $\downarrow$}                  & \multicolumn{3}{|c}{Worst-case performance diff. $\downarrow$}        \\ \hline
                      & \multicolumn{1}{c|}{}              &  \threeS        & \threeS+          & $\cD_{test, f}$   &  \threeS        & \threeS+                  & $\cD_{test, f}$   \\ \hline
\multirow{4}{*}{RF}   
& \#1 (86\%)                     & 6.05 $\pm$ 0.72                     & 2.33 $\pm$ 0.42                      & 9.16 $\pm$ 4.48                      & 7.08                     & 2.84                     & 11.65                                        \\ 
 & \#2 (9\%)                     & 4.16 $\pm$ 0.33                     & 4.29 $\pm$ 0.47                      & 5.85 $\pm$ 3.18                      & 4.44                     & 4.82                     & 8.94                                        \\ 
 & \#3 (3\%)                     & 2.15 $\pm$ 1.04                     & 1.88 $\pm$ 0.8                      & 13.59 $\pm$ 5.34                      & 3.41                     & 3.03                     & 19.73                                        \\ 
 & \#4 (1\%)                     & 4.54 $\pm$ 0.33                     & 4.54 $\pm$ 0.34                      & 7.1 $\pm$ 3.93                      & 4.89                     & 4.96                     & 13.23                                        \\ 
 & \#5 (1\%)                     & 3.15 $\pm$ 0.77                     & 3.06 $\pm$ 0.77                      & 8.72 $\pm$ 0.0                      & 4.2                     & 4.12                     & 8.72                                        \\
\hline
\multirow{4}{*}{GBDT} 
& \#1 (86\%)                     & 7.37 $\pm$ 1.09                     & 3.25 $\pm$ 0.32                      & 1.19 $\pm$ 0.81                      & 9.22                     & 3.6                     & 2.42                                        \\ 
 & \#2 (9\%)                     & 3.44 $\pm$ 0.52                     & 3.37 $\pm$ 0.54                      & 2.56 $\pm$ 1.08                      & 3.98                     & 4.02                     & 4.49                                        \\ 
 & \#3 (3\%)                     & 2.6 $\pm$ 1.53                     & 2.46 $\pm$ 1.38                      & 4.53 $\pm$ 1.04                      & 4.29                     & 3.97                     & 5.73                                        \\ 
 & \#4 (1\%)                     & 0.76 $\pm$ 0.58                     & 0.76 $\pm$ 0.57                      & 6.51 $\pm$ 5.07                      & 1.31                     & 1.34                     & 14.53                                        \\ 
 & \#5 (1\%)                     & 2.23 $\pm$ 0.82                     & 2.15 $\pm$ 0.85                      & 7.73 $\pm$ 3.33                      & 3.47                     & 3.45                     & 9.4                                        \\

\hline
\multirow{4}{*}{MLP}  
& \#1 (86\%)                     & 6.72 $\pm$ 1.2                     & 3.0 $\pm$ 0.46                      & 1.01 $\pm$ 0.62                      & 8.55                     & 3.5                     & 1.92                                        \\ 
 & \#2 (9\%)                     & 4.32 $\pm$ 0.46                     & 4.13 $\pm$ 0.54                      & 2.45 $\pm$ 1.46                      & 4.91                     & 4.87                     & 4.41                                        \\ 
 & \#3 (3\%)                     & 1.98 $\pm$ 1.1                     & 1.88 $\pm$ 0.99                      & 4.06 $\pm$ 0.8                      & 3.58                     & 3.33                     & 4.69                                        \\ 
 & \#4 (1\%)                     & 2.57 $\pm$ 0.5                     & 2.49 $\pm$ 0.53                      & 7.44 $\pm$ 3.17                      & 3.37                     & 3.36                     & 12.42                                        \\ 
 & \#5 (1\%)                     & 2.17 $\pm$ 0.62                     & 2.13 $\pm$ 0.64                      & 4.3 $\pm$ 3.66                      & 3.24                     & 3.22                     & 8.72                                        \\

\hline
\multirow{4}{*}{SVM} 
& \#1 (86\%)                     & 6.74 $\pm$ 0.58                     & 3.0 $\pm$ 0.49                      & 1.26 $\pm$ 0.93                      & 7.37                     & 3.6                     & 2.46                                        \\ 
 & \#2 (9\%)                     & 5.62 $\pm$ 0.14                     & 5.31 $\pm$ 0.4                      & 3.63 $\pm$ 1.25                      & 5.83                     & 5.81                     & 5.63                                        \\ 
 & \#3 (3\%)                     & 1.09 $\pm$ 0.73                     & 1.04 $\pm$ 0.73                      & 2.51 $\pm$ 1.09                      & 2.43                     & 2.33                     & 3.88                                        \\ 
 & \#4 (1\%)                     & 1.08 $\pm$ 0.4                     & 0.97 $\pm$ 0.39                      & 11.62 $\pm$ 4.31                      & 1.51                     & 1.44                     & 16.01                                        \\ 
 & \#5 (1\%)                     & 4.22 $\pm$ 0.6                     & 4.17 $\pm$ 0.61                      & 3.1 $\pm$ 2.99                      & 5.2                     & 5.18                     & 6.71                                        \\ 
\hline

\multirow{4}{*}{AdaBoost} 
& \#1 (86\%)                     & 6.89 $\pm$ 0.96                     & 3.06 $\pm$ 0.31                      & 1.11 $\pm$ 0.71                      & 8.61                     & 3.44                     & 2.14                                        \\ 
 & \#2 (9\%)                     & 3.91 $\pm$ 0.3                     & 3.76 $\pm$ 0.44                      & 2.31 $\pm$ 1.86                      & 4.24                     & 4.31                     & 5.15                                        \\ 
 & \#3 (3\%)                     & 2.13 $\pm$ 1.43                     & 2.01 $\pm$ 1.3                      & 4.59 $\pm$ 2.48                      & 4.22                     & 4.0                     & 7.79                                        \\ 
 & \#4 (1\%)                     & 1.24 $\pm$ 0.65                     & 1.16 $\pm$ 0.69                      & 8.66 $\pm$ 6.46                      & 1.99                     & 1.97                     & 19.89                                        \\ 
 & \#5 (1\%)                     & 3.42 $\pm$ 0.71                     & 3.36 $\pm$ 0.73                      & 5.17 $\pm$ 3.55                      & 4.47                     & 4.44                     & 8.05                                        \\ 
\hline

\multirow{4}{*}{Bagging} 
& \#1 (86\%)                     & 5.81 $\pm$ 1.02                     & 2.65 $\pm$ 0.53                      & 10.01 $\pm$ 4.49                      & 7.03                     & 3.61                     & 12.83                                        \\ 
 & \#2 (9\%)                     & 4.89 $\pm$ 0.68                     & 5.06 $\pm$ 0.44                      & 7.65 $\pm$ 3.22                      & 5.97                     & 5.54                     & 10.69                                        \\ 
 & \#3 (3\%)                     & 3.72 $\pm$ 1.65                     & 3.37 $\pm$ 1.64                      & 8.5 $\pm$ 4.77                      & 5.98                     & 5.67                     & 14.55                                        \\ 
 & \#4 (1\%)                     & 7.01 $\pm$ 0.28                     & 7.0 $\pm$ 0.3                      & 6.16 $\pm$ 3.51                      & 7.46                     & 7.48                     & 11.21                                        \\ 
 & \#5 (1\%)                     & 5.06 $\pm$ 1.09                     & 5.0 $\pm$ 1.09                      & 4.41 $\pm$ 4.07                      & 6.59                     & 6.55                     & 9.4                                        \\ 
\hline

\multirow{4}{*}{LR} 
 & \#1 (86\%)                     & 6.51 $\pm$ 0.63                     & 2.97 $\pm$ 0.3                      & 1.09 $\pm$ 0.41                      & 7.57                     & 3.38                     & 1.81                                        \\ 
 & \#2 (9\%)                     & 4.11 $\pm$ 0.32                     & 3.9 $\pm$ 0.51                      & 3.33 $\pm$ 0.98                      & 4.52                     & 4.52                     & 4.57                                        \\ 
 & \#3 (3\%)                     & 1.15 $\pm$ 0.73                     & 1.11 $\pm$ 0.64                      & 4.6 $\pm$ 2.72                      & 2.28                     & 2.08                     & 7.61                                        \\ 
 & \#4 (1\%)                     & 1.69 $\pm$ 0.59                     & 1.62 $\pm$ 0.62                      & 7.16 $\pm$ 4.27                      & 2.57                     & 2.55                     & 13.48                                        \\ 
 & \#5 (1\%)                     & 4.08 $\pm$ 0.76                     & 4.02 $\pm$ 0.78                      & 4.76 $\pm$ 3.21                      & 5.44                     & 5.41                     & 7.38                                        \\
\hline
\end{tabular}

}
\vspace{-0cm}
\label{table:dtrain}
\end{table*}

\clearpage

\clearpage
\section{Example Model Report}
\label{appx:modelreport}
Below we present an example of the type of model report that could be produced when evaluating models using \threeS testing.

\noindent\fbox{%
    \parbox{\textwidth}{%
 \textbf{Dataset}: Adult \citep{uci}.
}}

\noindent\fbox{%
    \parbox{\textwidth}{%
 \textbf{Intersectional model performance matrix}: diagnosing at a granular level.
}}

\begin{figure}[hbt]
\centering
 \includegraphics[width=0.95\textwidth]{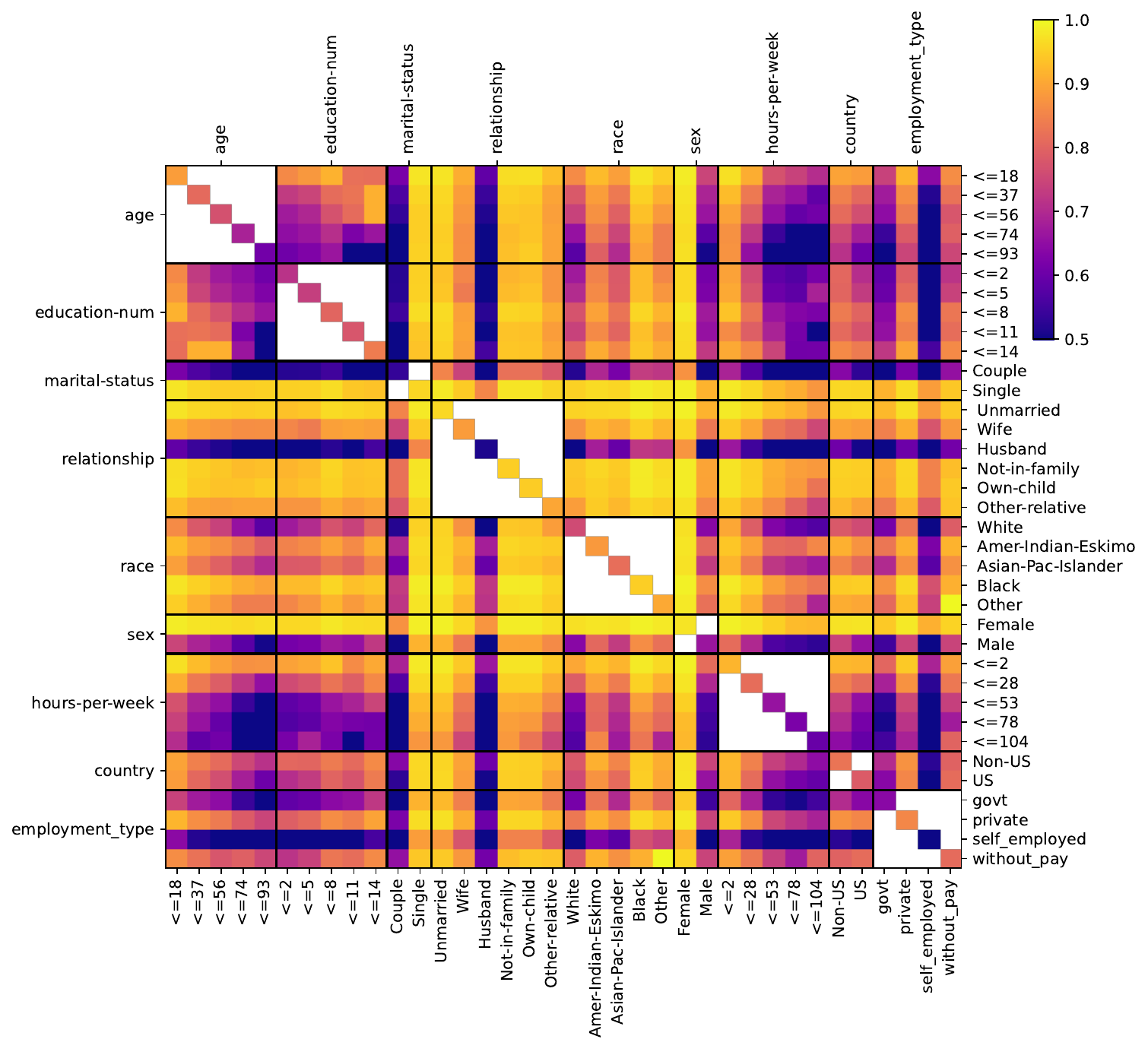}
\captionof{figure}{Intersectional performance matrix for the RF model, which diagnoses underperforming 2-feature subgroups (darker implies underperformance).}\label{model-matrix}
\end{figure}

\vspace{1cm}

\noindent\fbox{%
    \parbox{\textwidth}{%
 \textbf{INSIGHT}: Model underperformance on self-employed
who work more than 80 hours a week. (bottom right arrow)

    }%
}

\newpage

\noindent\fbox{%
    \parbox{\textwidth}{%
 \textbf{Model sensitivity curves: helping to understand performance trends for shifts across the operating range}:

    }%
}

\begin{figure*}[!h]

\centering
    \begin{subfigure}[t]{.45\textwidth}
        \includegraphics[width=\textwidth]{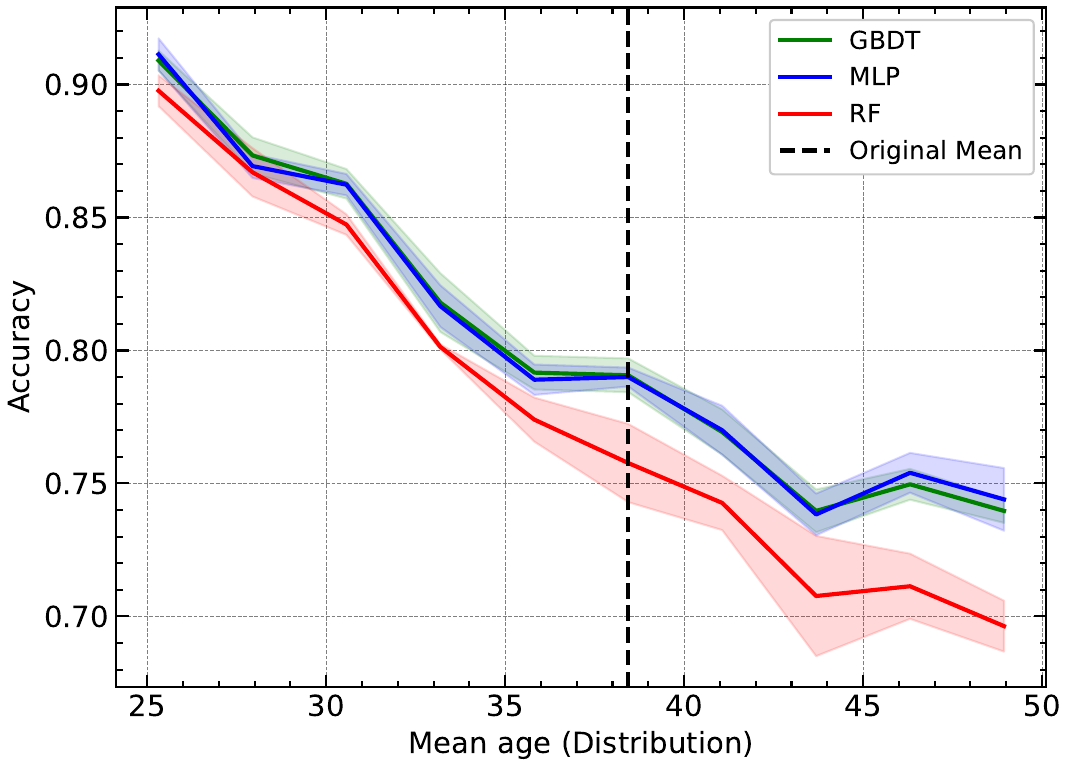}

        \caption{\footnotesize{Age: Performance decreases as mean age increases.}}
    \end{subfigure}%
    \hspace{10mm}
    \begin{subfigure}[t]{.45\textwidth}
        \includegraphics[width=\textwidth]{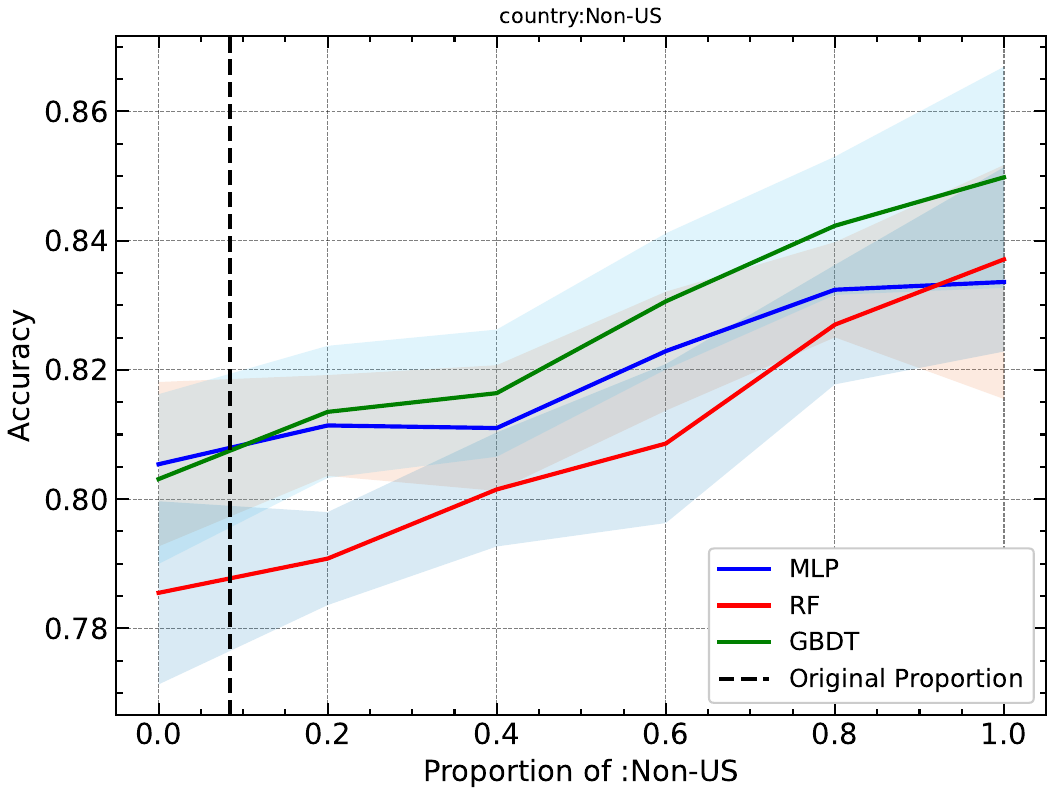}

        \caption{\footnotesize{Country of origin: performance increases as the proportion of non-US individuals increases}}
    \end{subfigure}%
    \\
    
    \begin{subfigure}[t]{.45\textwidth}
        \includegraphics[width=\textwidth]{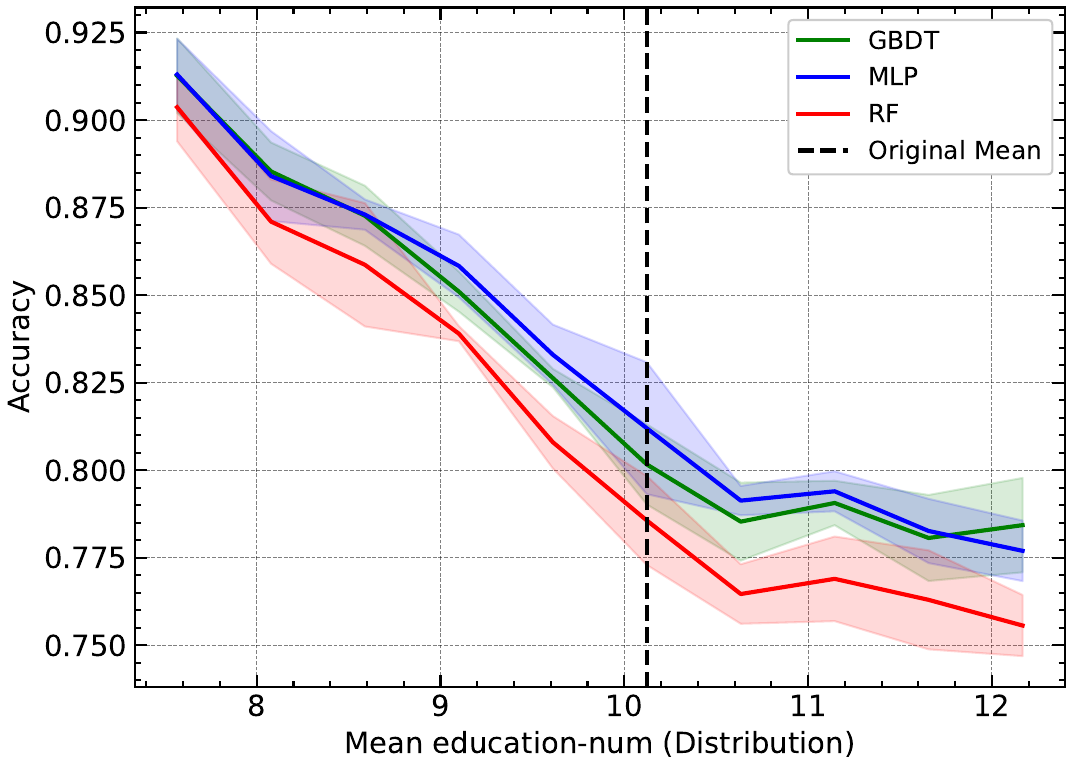}
        \caption{\footnotesize{Number of years of education. Performance decreases as mean number of education years increases.}}
    \end{subfigure}%
            \hspace{10mm}
    \begin{subfigure}[t]{.45\textwidth}
        \includegraphics[width=\textwidth]{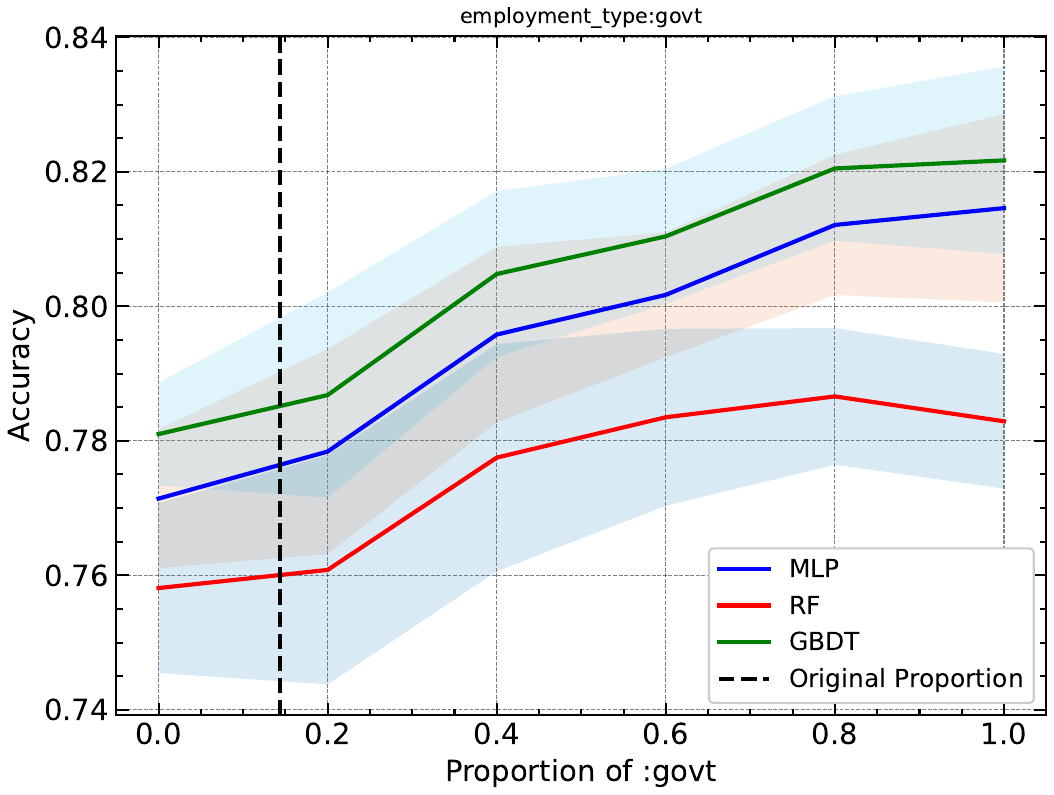}

        \caption{\footnotesize{Employment type (government): performance increases as the proportion of government employed individuals increases}}
    \end{subfigure}%
    \\
    \begin{subfigure}[t]{.45\textwidth}
        \includegraphics[width=\textwidth]{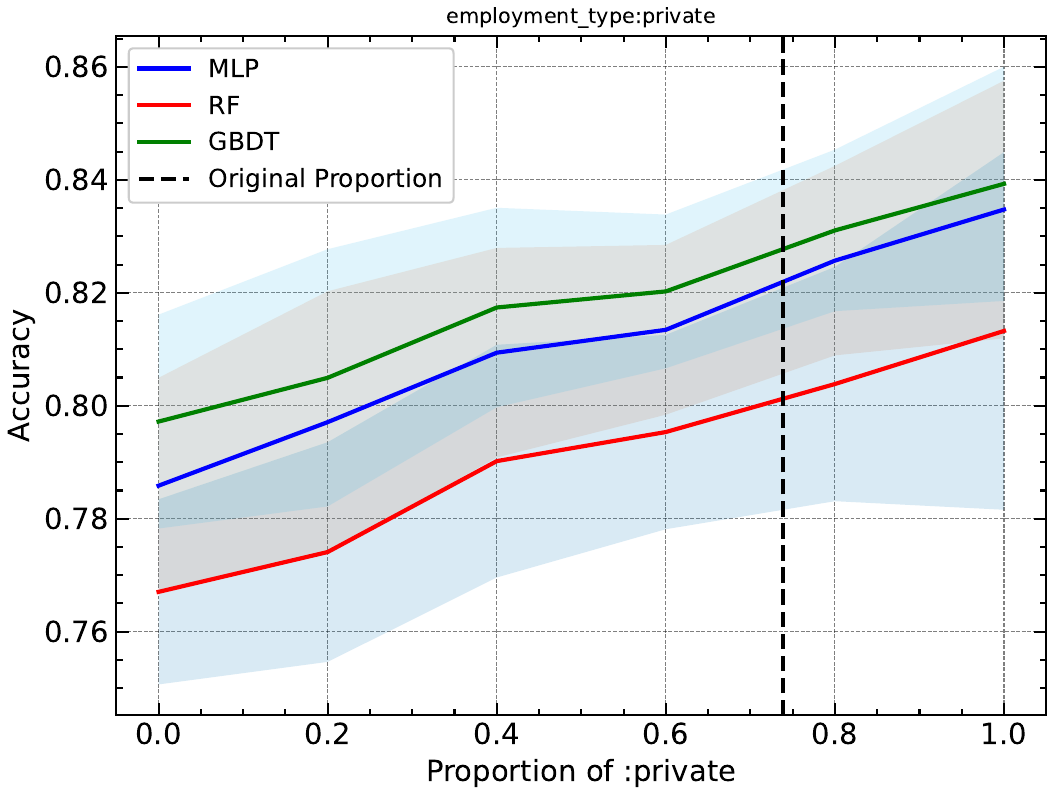}

        \caption{\footnotesize{Employment type (private): performance increases as the proportion of private employed individuals increases}}
    \end{subfigure}%
    \hspace{10mm}
    \begin{subfigure}[t]{.45\textwidth}
        \includegraphics[width=\textwidth]{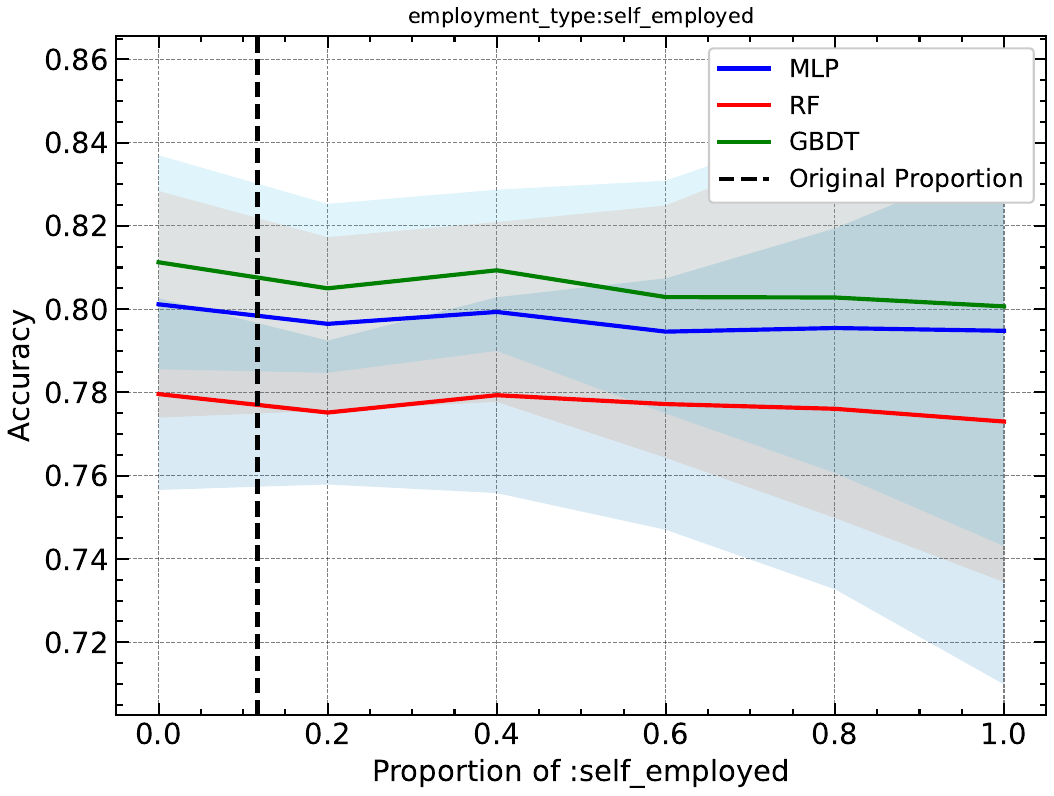}

        \caption{\footnotesize{Employment type (self-employed): performance is consistent even as proportion of self-employed individuals increases}}
    \end{subfigure}%

    \caption{Model sensitivity curves for different features, illustrating the relationships/model performance across the operating range.}
    \vspace{-0mm}
    \rule{\textwidth}{.5pt}
 \vspace{-0mm}
\end{figure*}

\begin{figure*}[!h]\ContinuedFloat

\centering

    \begin{subfigure}[t]{.45\textwidth}
        \includegraphics[width=\textwidth]{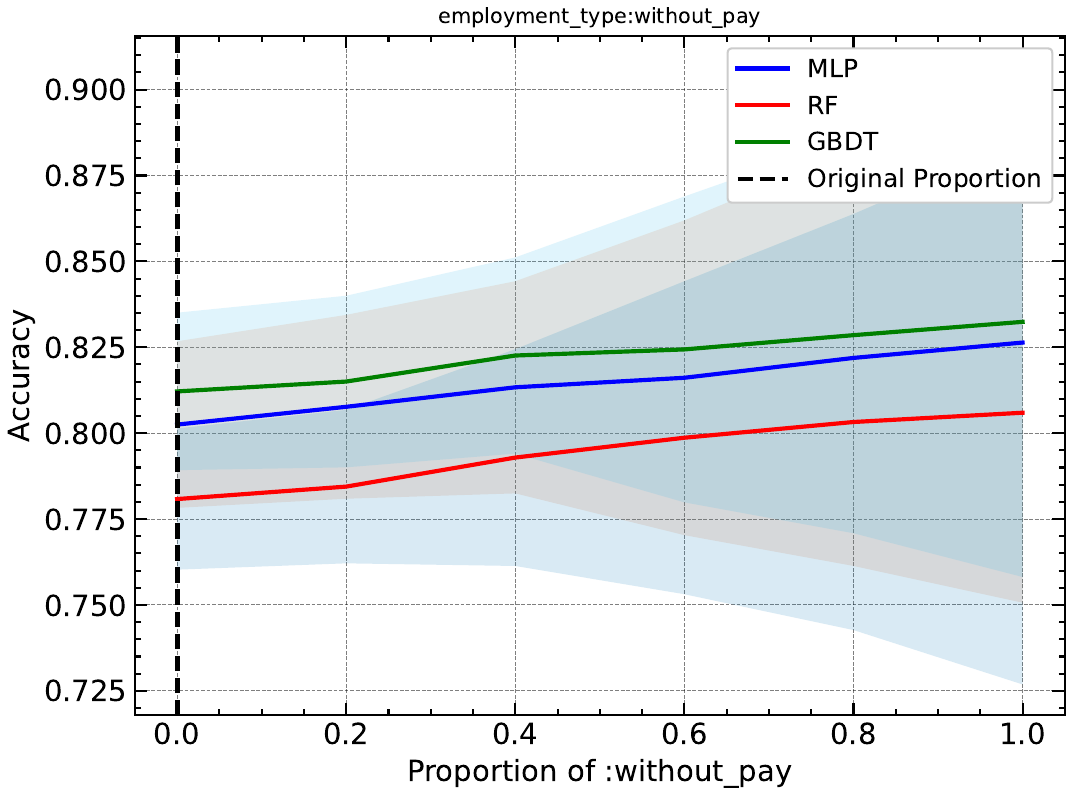}

        \caption{\footnotesize{Employment type (without pay): performance is consistent even as proportion of without-pay individuals increases}}
    \end{subfigure}%
    \hspace{10mm}
    \begin{subfigure}[t]{.45\textwidth}
        \includegraphics[width=\textwidth]{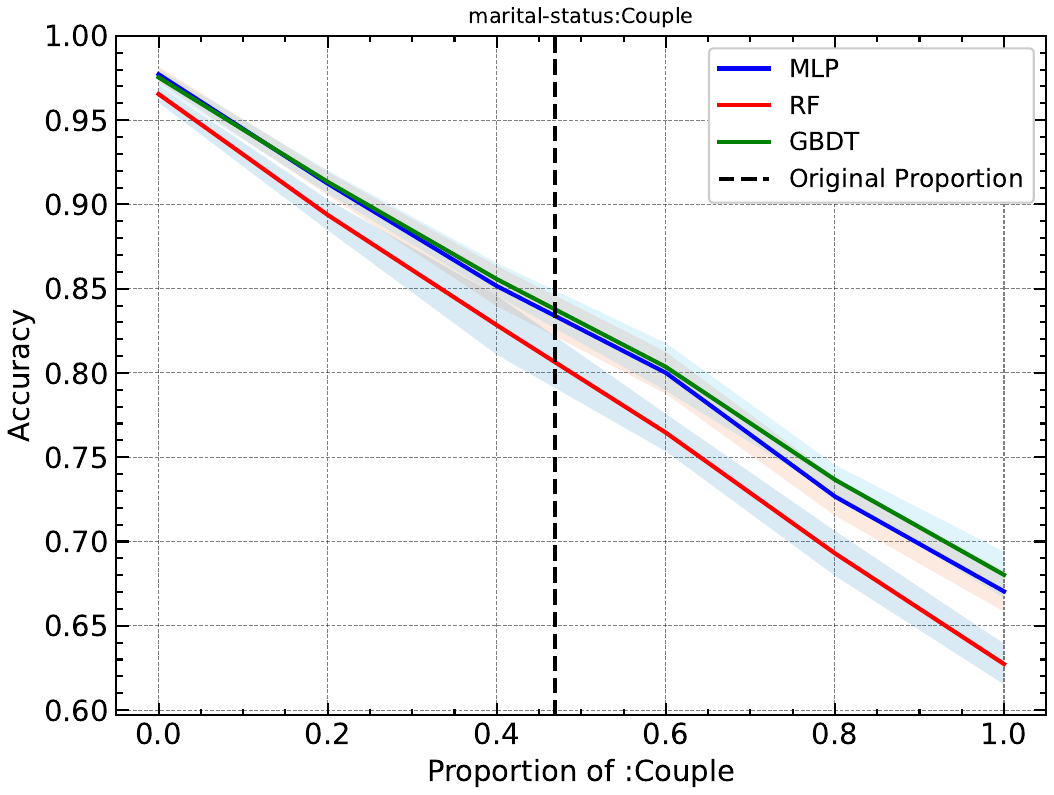}

        \caption{\footnotesize{Marital status (couple): performance decreases as the proportion of married individuals increases}}
    \end{subfigure}%
    \\
    
    \begin{subfigure}[t]{.45\textwidth}
        \includegraphics[width=\textwidth]{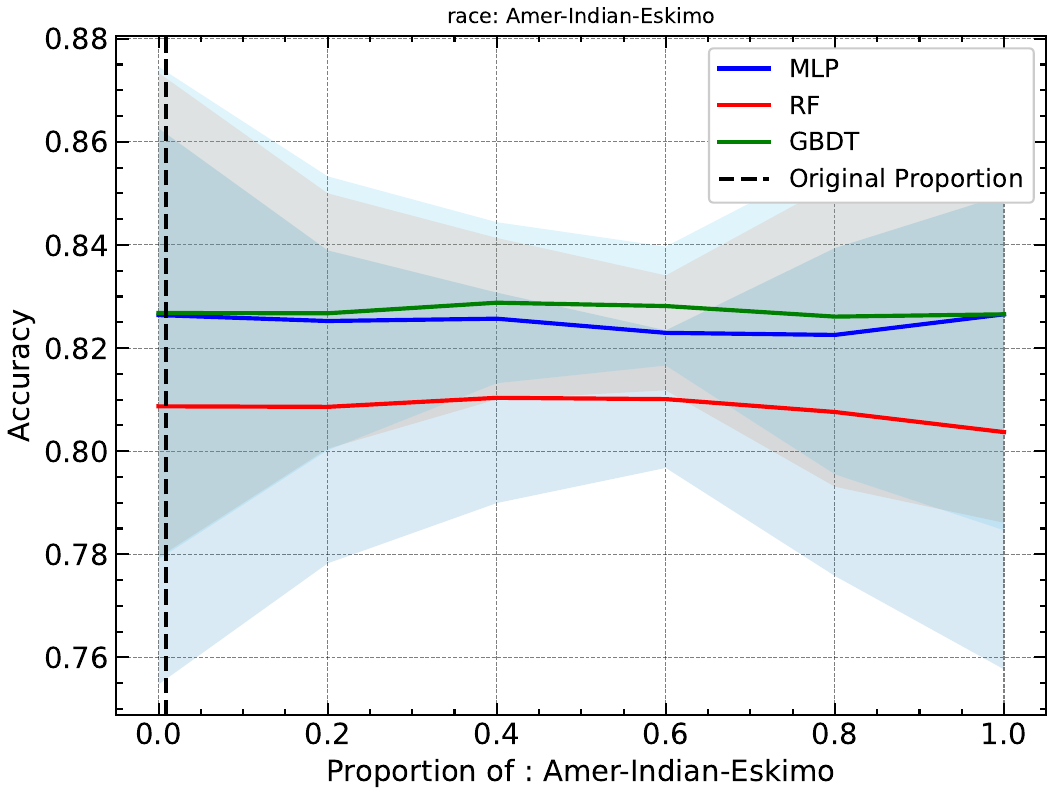}

        \caption{\footnotesize{Race (American-Indian-Eskimo): performance remains consistent even as the proportion of American-Indian-Eskimo individuals increases}}
    \end{subfigure}%
      \hspace{10mm}
    \begin{subfigure}[t]{.45\textwidth}
        \includegraphics[width=\textwidth]{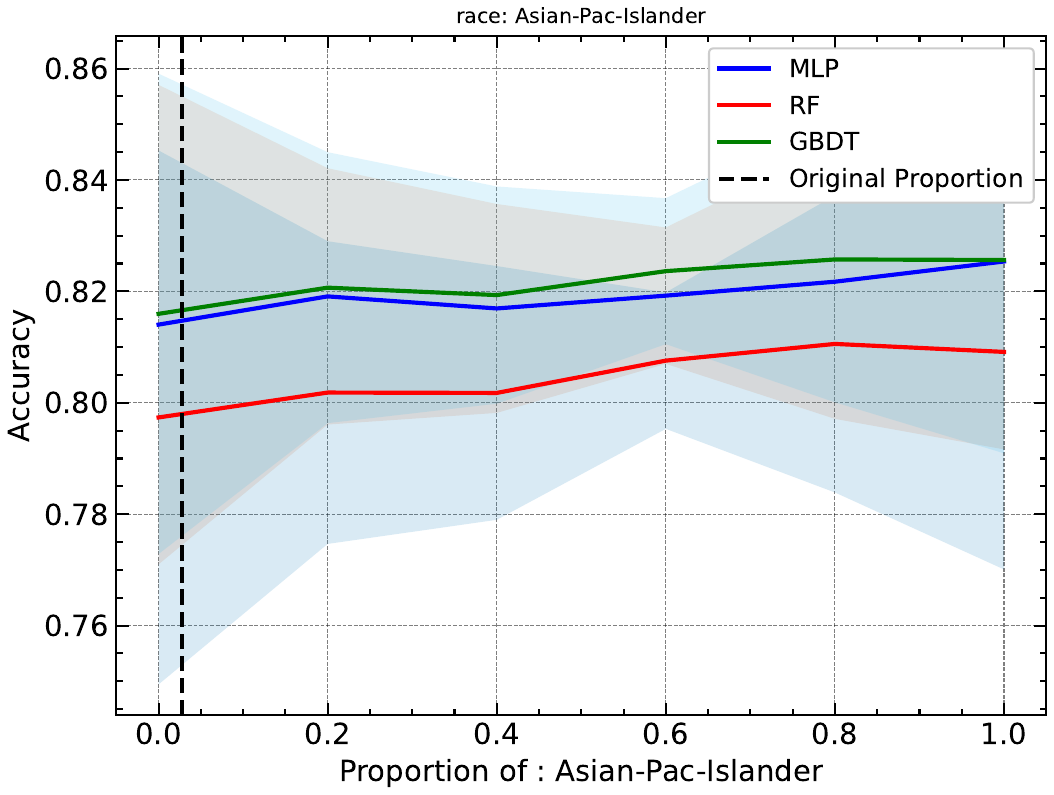}
        \caption{\footnotesize{Race (Asian-Pacific-Islander): performance remains consistent even as the proportion of Asian-Pacific-Islander individuals increases}}
    \end{subfigure}%
     \\
     
    \begin{subfigure}[t]{.45\textwidth}
        \includegraphics[width=\textwidth]{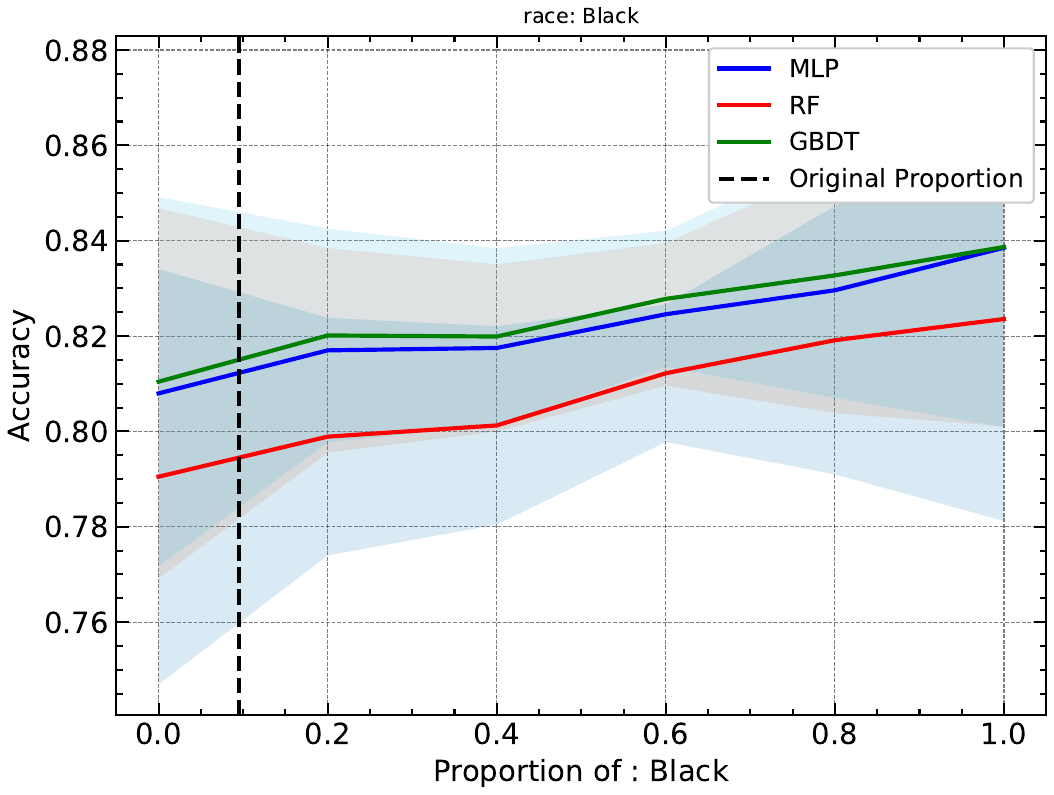}

        \caption{\footnotesize{Race (Black): performance increases as the proportion of black individuals increases}}
    \end{subfigure}%
     \hspace{10mm}
    \begin{subfigure}[t]{.45\textwidth}
        \includegraphics[width=\textwidth]{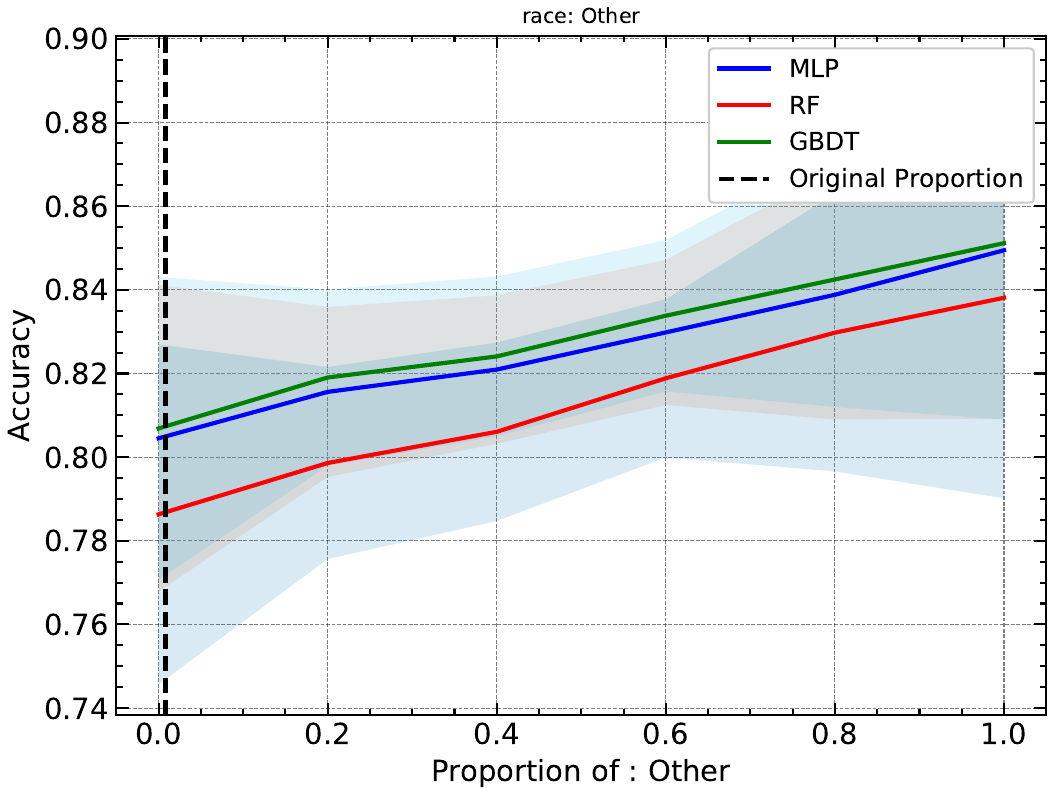}

        \caption{\footnotesize{Race (Other): performance increases as the proportion of other individuals increases}}
    \end{subfigure}%

    \caption{Model sensitivity curves for different features, illustrating the relationships/model performance across the operating range.}
    \vspace{-0mm}
    \rule{\textwidth}{.5pt}
 \vspace{-0mm}
\end{figure*}

\begin{figure*}[!h]\ContinuedFloat

\centering
    
        \begin{subfigure}[t]{.45\textwidth}
        \includegraphics[width=\textwidth]{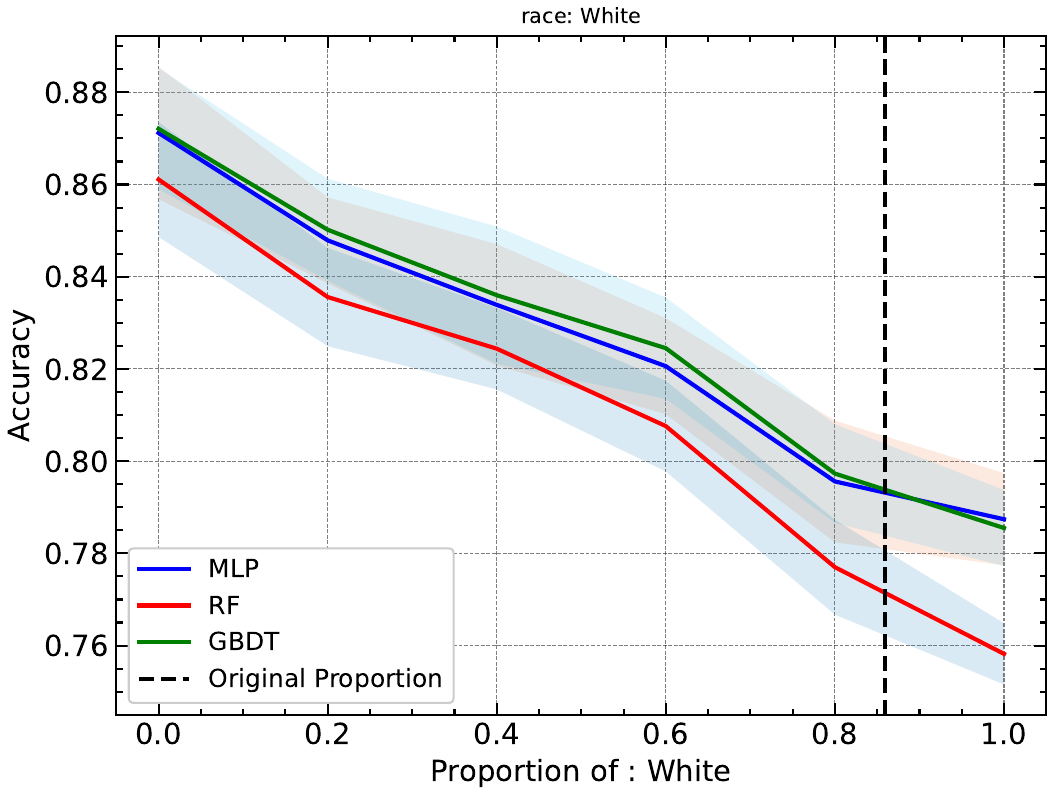}

        \caption{\footnotesize{Race (White): performance decreases as the proportion of white individuals increases}}
    \end{subfigure}%
          \hspace{10mm}
    \begin{subfigure}[t]{.45\textwidth}
        \includegraphics[width=\textwidth]{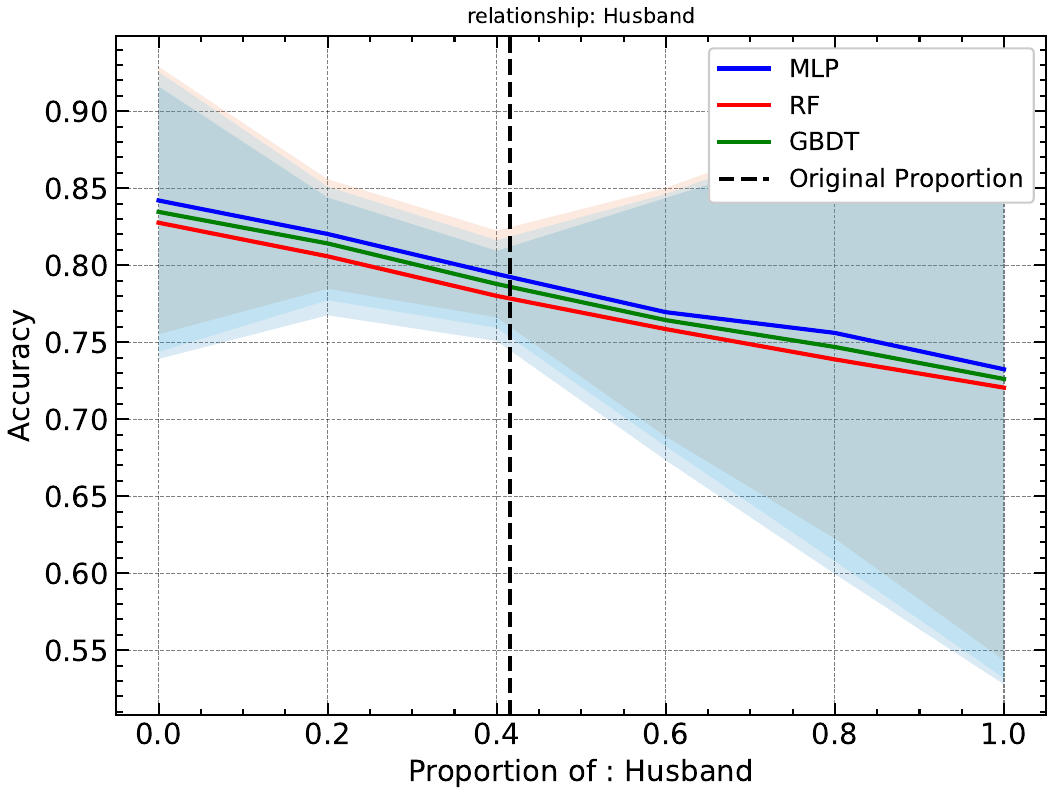}

        \caption{\footnotesize{Relationship (Husband): performance decreases as the proportion of individuals classed as husbands increases}}
    \end{subfigure}%
    \\
    \begin{subfigure}[t]{.45\textwidth}
        \includegraphics[width=\textwidth]{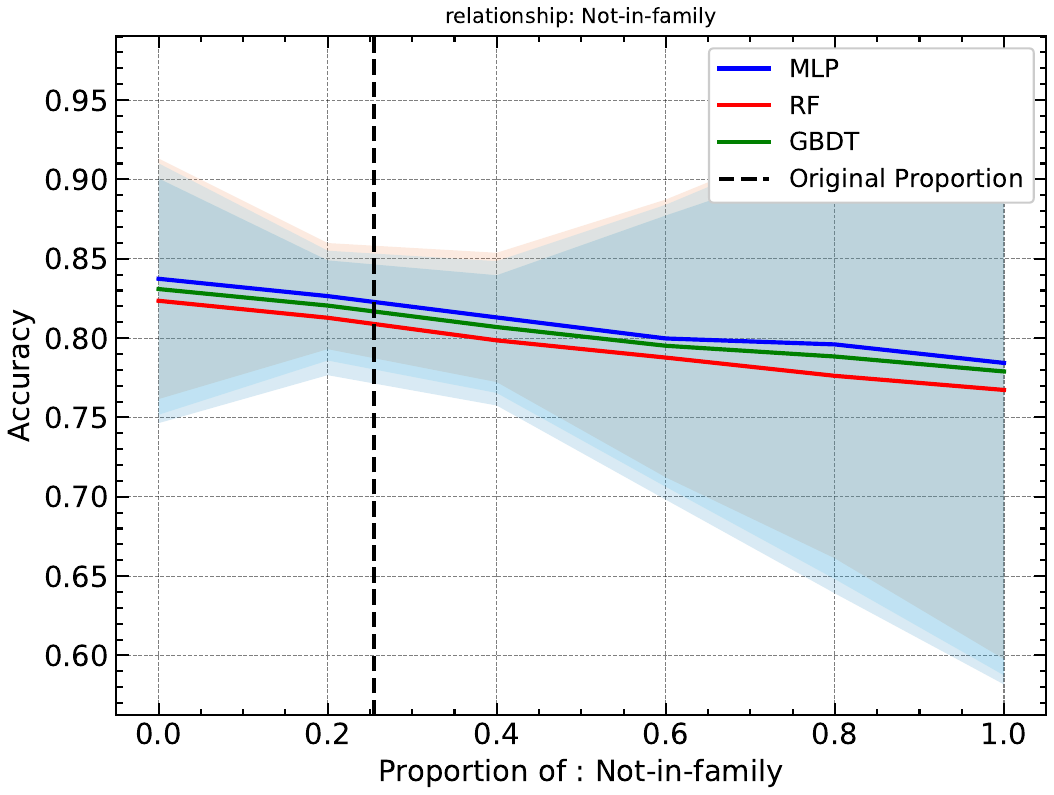}

        \caption{\footnotesize{Relationship (Not-in-family): performance decreases as the proportion of individuals classed as Not-in-family increases}}
    \end{subfigure}%
          \hspace{10mm}
    \begin{subfigure}[t]{.45\textwidth}
        \includegraphics[width=\textwidth]{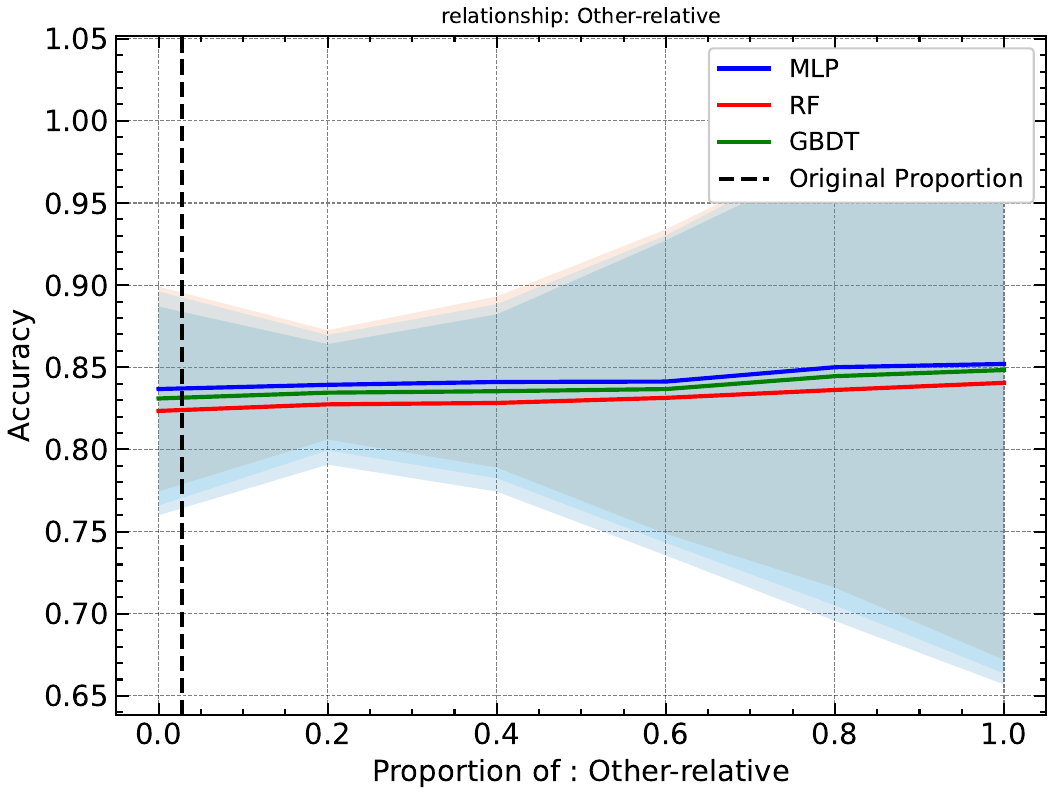}

        \caption{\footnotesize{Relationship (other-relative): performance remains consistent as the proportion of individuals classed as other-relative increases}}
    \end{subfigure}%
      \\
    \begin{subfigure}[t]{.45\textwidth}
        \includegraphics[width=\textwidth]{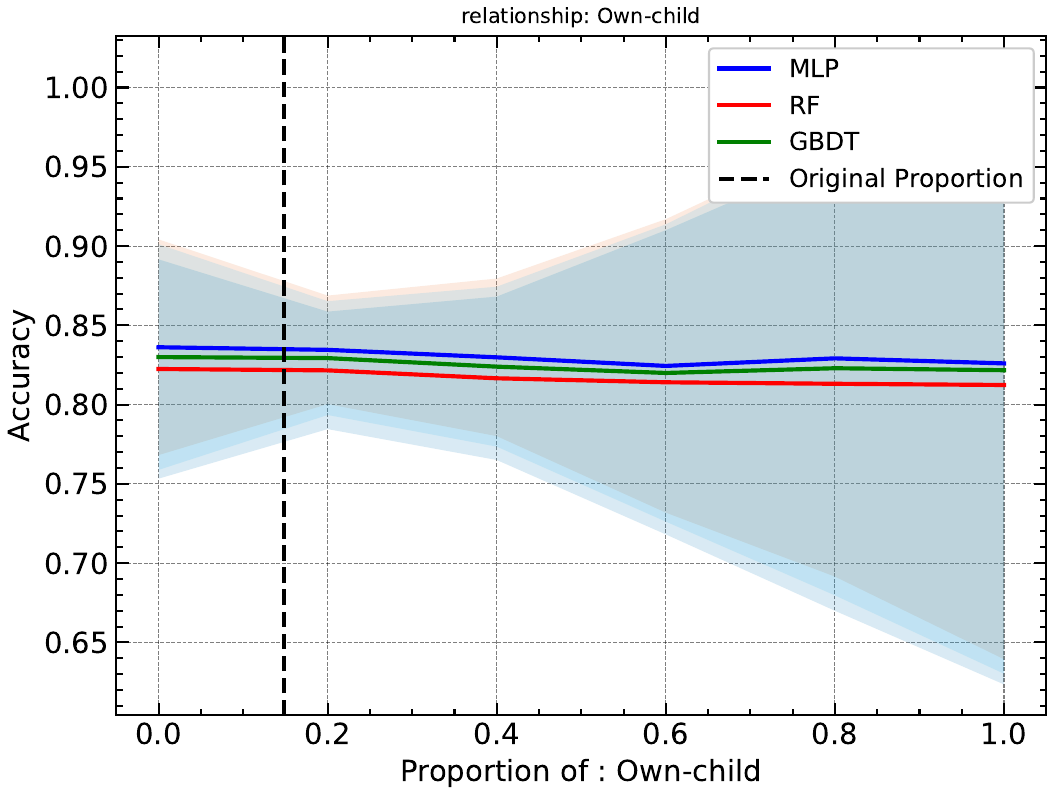}

        \caption{\footnotesize{Relationship (Own-child): performance remains consistent as the proportion of individuals classed as Own-child increases}}
    \end{subfigure}%
          \hspace{10mm}
    \begin{subfigure}[t]{.45\textwidth}
        \includegraphics[width=\textwidth]{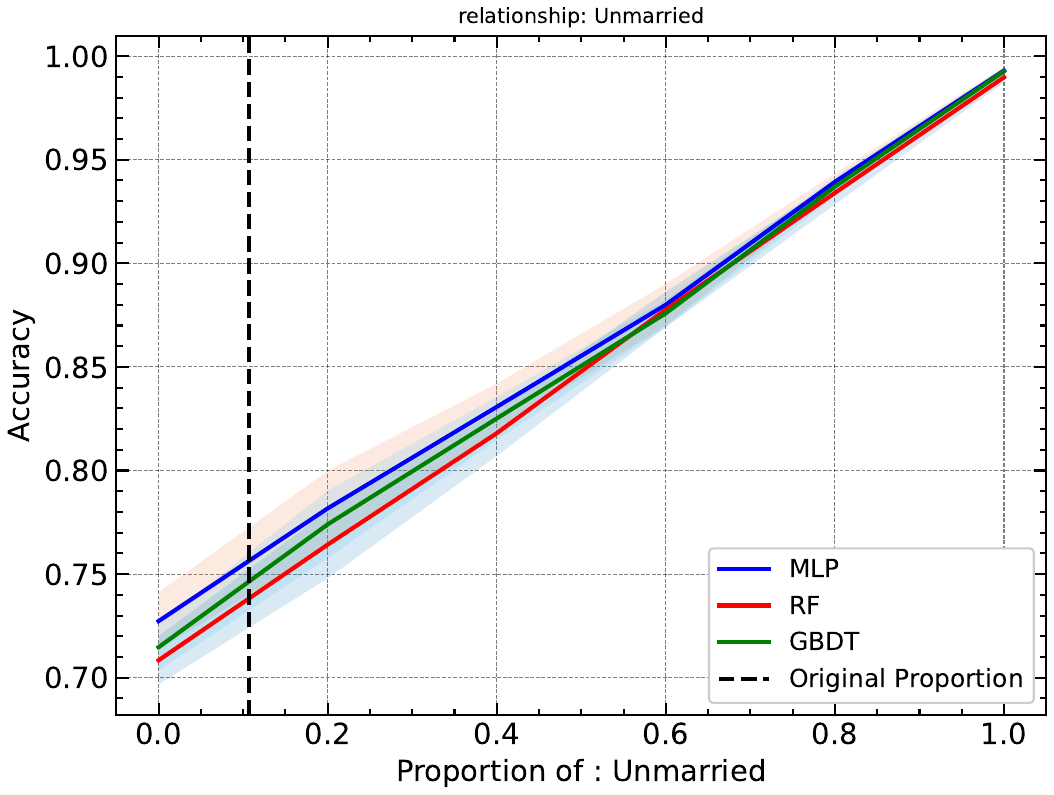}

        \caption{\footnotesize{Relationship (Unmarried): performance increases as the proportion of individuals classed as Unmarried increases}}
    \end{subfigure}%
    \caption{Model sensitivity curves for different features, illustrating the relationships/model performance across the operating range.}
    \vspace{-0mm}
    \rule{\textwidth}{.5pt}
 \vspace{-0mm}
\end{figure*}

\begin{figure*}[!h]\ContinuedFloat

\centering
    \begin{subfigure}[t]{.45\textwidth}
        \includegraphics[width=\textwidth]{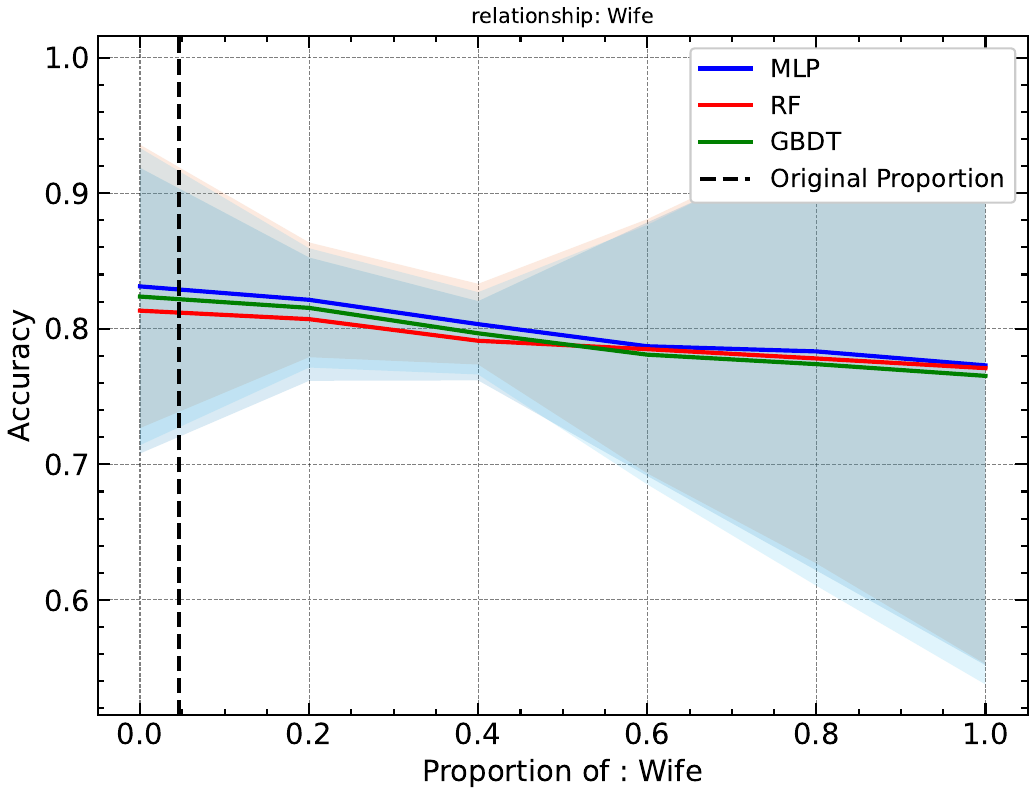}

        \caption{\footnotesize{Relationship (Wife): performance decreases as the proportion of individuals classed as Wife increases}}
    \end{subfigure}%
          \hspace{10mm}
    \begin{subfigure}[t]{.45\textwidth}
        \includegraphics[width=\textwidth]{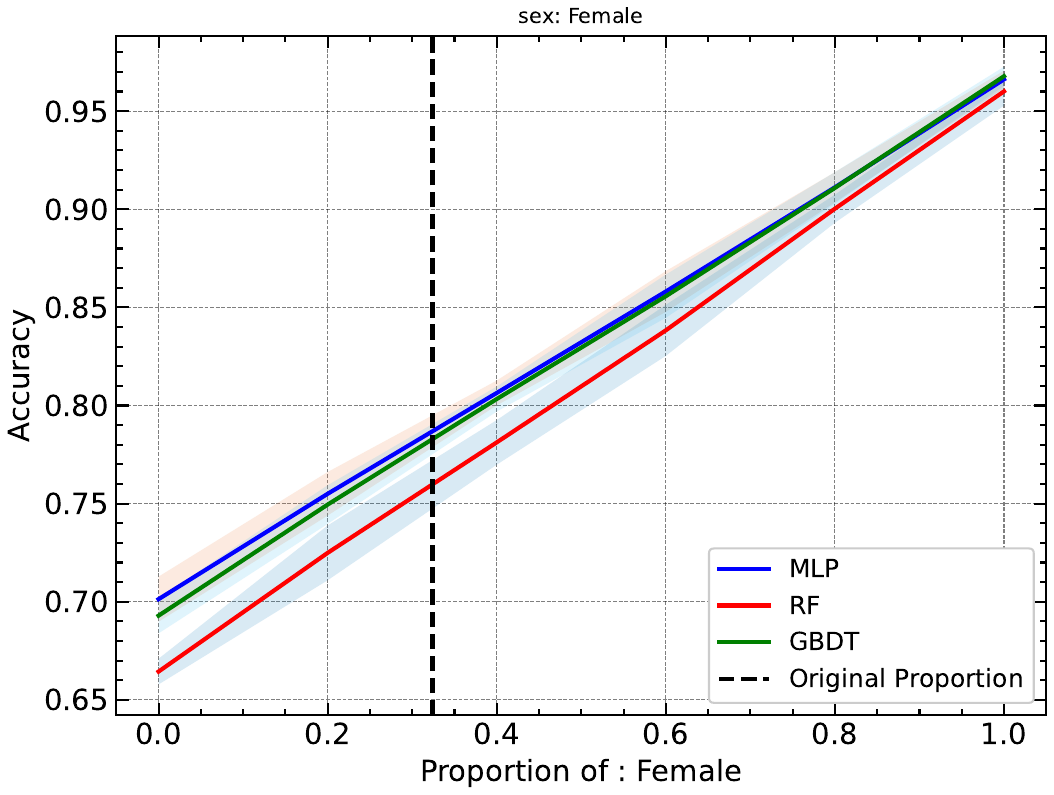}

        \caption{\footnotesize{Sex (Female): performance increases as the proportion of females increases}}
    \end{subfigure}%

    \caption{Model sensitivity curves for different features, illustrating the relationships/model performance across the operating range.}\label{fig:modelreport-sensitivity}
    \vspace{-0mm}
    \rule{\textwidth}{.5pt}
 \vspace{-0mm}
\end{figure*}

\end{document}